\documentclass{article}

\usepackage{microtype}
\usepackage{graphicx}
\usepackage{subfigure}
\usepackage{booktabs} % for professional tables

\usepackage{hyperref}

\usepackage[accepted]{icml2024}

\usepackage{amsmath}
\usepackage{amssymb}
\usepackage{mathtools}
\usepackage{amsthm}
\usepackage{fancybox}
\usepackage{bbding}
\usepackage{lipsum}

\usepackage{algorithm}
\usepackage{amssymb,multirow,paralist,color,amsmath}
\usepackage{subfigure}
\usepackage{epstopdf}
\usepackage{booktabs} % for professional tables
\usepackage{siunitx}
\usepackage{makecell}
\usepackage{hyperref}
\usepackage[normalem]{ulem}
\usepackage{fancyvrb}
\usepackage{listings}
\usepackage{tcolorbox}
\usepackage{enumitem}
\usepackage{wrapfig}

\def \A {\mathcal{A}}

\def \O {\mathcal{O}}

\def \R {\mathbb{R}}

\def \w {\mathbf{w}}
\def \v {\mathbf{v}}
\def \t {\mathbf{t}}
\def \x {\mathbf{x}}

\def \E {\mathrm{E}}

\def \x {\mathbf{x}}
\def \p {\mathbf{p}}

\def \1 {\mathbf{1}}

\def \u {\mathbf{u}}

\def \E {\mathrm{E}}
\def \x {\mathbf{x}}

\def \D {\mathcal{D}}

\def \u {\mathbf{u}}

\def \w {\mathbf{w}}

\def \R {\mathbb{R}}

\def \A {\mathcal{A}}

\def \v {\mathbf{v}}
\def \M {\mathcal{M}}

\def \p {\mathbf{p}}

\def \t {\mathbf{t}}

\def \E {\mathbb{E}}

\def\Tau{{\rm T}}

\definecolor{darkgreen}{RGB}{0, 180, 0}

\DeclareMathOperator*{\argmin}{arg\,min}

\usepackage[capitalize,noabbrev]{cleveref}

\theoremstyle{plain}
\newtheorem{theorem}{Theorem}[section]

\newtheorem{lemma}[theorem]{Lemma}

\theoremstyle{definition}

\theoremstyle{remark}

\usepackage{setspace}
\AtBeginDocument{%
  \addtolength\abovedisplayskip{-0.35\baselineskip}%
  \addtolength\belowdisplayskip{-0.35\baselineskip}%
  \addtolength\abovedisplayshortskip{-0.35\baselineskip}%
  \addtolength\belowdisplayshortskip{-0.35\baselineskip}%
}
\usepackage{titlesec}
\titlespacing*{\section}{0pt}{-0.05\baselineskip}{-0.075\baselineskip}
\titlespacing*{\subsection}{0pt}{-0.075\baselineskip}{-0.05\baselineskip}
\titlespacing*{\subsubsection}{0pt}{-0.075\baselineskip}{-0.05\baselineskip}

\usepackage[textsize=tiny]{todonotes}

\usepackage[T1]{fontenc}
\usepackage{textcomp}
\usepackage[utf8x]{inputenc}

\icmltitlerunning{To Cool or not to Cool? Temperature Network Meets Large Foundation Models via DRO}

\begin{document}

\twocolumn[
%\icmltitle{TempNet: A Small yet Generalizable Temperature Prediction Network for Improving Large Foundation Models}
\icmltitle{To Cool or not to Cool? \\Temperature Network Meets Large Foundation Models via DRO}



\icmlsetsymbol{equal}{*}

\begin{icmlauthorlist}
\icmlauthor{Zi-Hao Qiu}{nju}
\icmlauthor{Siqi Guo}{tamu}
\icmlauthor{Mao Xu}{nju}
\icmlauthor{Tuo Zhao}{gatech}
\icmlauthor{Lijun Zhang}{nju}
\icmlauthor{Tianbao Yang}{tamu}
\end{icmlauthorlist}

\icmlaffiliation{nju}{National Key Laboratory for Novel Software Technology, Nanjing University, Nanjing, China}
\icmlaffiliation{tamu}{Computer Science and Engineering, Texas A\&M University, College Station, USA}
\icmlaffiliation{gatech}{Computer Science and Engineering, Georgia Institute of Technology, Atlanta, USA}

\icmlcorrespondingauthor{Tianbao Yang}{tianbao-yang@tamu.edu}
%\icmlcorrespondingauthor{Firstname2 Lastname2}{first2.last2@www.uk}

\icmlkeywords{Machine Learning, ICML}

\vskip 0.3in
]



\printAffiliationsAndNotice{The work of Z.H. Qiu was conducted when visiting T. Yang's Lab.} % otherwise use the standard text.

\setlength{\textfloatsep}{2pt}% Remove \textfloatsep

\setlength\abovedisplayskip{1pt}
\setlength\belowdisplayskip{1pt}
\begin{abstract}
The temperature parameter plays a profound role  during training and/or inference with large foundation models (LFMs) such as large language models (LLMs) and CLIP models. Particularly, it adjusts the logits in the softmax function in LLMs, which is crucial for next token generation, and it scales the similarities in the contrastive loss for training CLIP models. A significant question remains: ``{\it Is it viable to learn a neural network to predict a personalized temperature of any input data for enhancing LFMs?"}  In this paper, we present {\bf a principled framework} for learning a small yet generalizable temperature prediction network (TempNet) to improve LFMs. Our solution is composed of a novel learning framework with a robust loss underpinned by constrained distributionally robust optimization (DRO), and a properly designed TempNet with theoretical inspiration. TempNet can be trained together with a large foundation model from scratch or learned separately given a pretrained foundation model. It is not only useful for predicting personalized temperature to promote the training of LFMs but also generalizable and transferable to new tasks. Our experiments on LLMs and CLIP models demonstrate that TempNet greatly improves the performance of existing solutions or models, e.g. Table 1. The code to reproduce the experimental results in this paper can be found at \url{https://github.com/zhqiu/TempNet}.

\end{abstract}

\section{Introduction}
\label{sec:introduction}

Originating from statistical mechanics~\cite{jaynes1957information}, temperature scaling is widely applied in softmax-type functions for training and/or inference with LFMs such as LLMs and multi-modal CLIP models\footnote{A contrastive loss is also regarded as a softmax-type function.}. In the deployment of LLMs for generating responses to prompts, a temperature-scaled softmax function models the probability distribution across a vocabulary of tokens. %The crucial role of temperature has been noticed in previous works~\citep{chen2021evaluating,touvron2023llama2}. 
The CLIP model~\citep{radford2021learning} is trained by optimizing a temperature-scaled contrastive loss~\citep{oord2018representation} on immense scale image-text data, which has been used in many downstream tasks.

\iffalse
\begin{table}[t]
\vspace{-5mm}
\caption{Evaluations on AlpacaEval. TT stands for training time measured by GPU hours on A100-80G GPUs. The evaluations were conducted in February, 2024 and the results of GPT4 and Gemini Pro are from Alpaca leaderboard in April, 2024.  Other models are evaluated by us by averaging over three trials. } %The training is conducted on $4$ A100-80GB GPUs.}
\vspace{-2mm}\begin{center}
\begin{small}
\begin{tabular}{l|c|l}
\toprule
Model & TT (h) & Win rate (\%) \\
\midrule
%LLaMA2 Chat 7B (leaderboard) & $\textemdash$  & 4.96\\
%LLaMA2 Chat 13B (leaderboard)& $\textemdash$ & 7.70\\
%LLaMA2 Chat 70B (leaderboard)& $\textemdash$ & 13.87\\
Gemini Pro (leaderboard) & $\textemdash$ & 18.2\\
GPT4 (06/13, leaderboard) & $\textemdash$ & 15.76\\
\midrule
LLaMA2 Chat 7B w/ $\tau=0.7$ & $\textemdash$ &4.78$\pm$0.13 \\
LLaMA2 Chat 7B w/ TempNet & 1.42 &\textbf{5.91}$\pm$0.31 \\
\midrule
LLaMA2 Chat 13B w/ $\tau=0.7$ & $\textemdash$ &7.58$\pm$0.23  \\
LLaMA2 Chat 13B w/ TempNet & 3.68 &\textbf{8.11}$\pm$0.26 \\
\midrule
LLaMA2 Chat 70B w/ $\tau=0.7$ & $\textemdash$ &14.13$\pm$0.37  \\
LLaMA2 Chat 70B w/ TempNet & 11.4 &\textbf{15.81}$\pm$0.56\\
\midrule
Yi Chat 34B w/ $\tau=0.3$ & $\textemdash$ & 28.84$\pm$0.63 \\
Yi Chat 34B w/ TempNet &21.0 &\textbf{29.25}$\pm$0.75\\
\bottomrule
\end{tabular}
\end{small}
\end{center}
\label{tab:alpaca_eval}
\end{table}
\fi

\begin{table}[t]
\vspace{-5mm}
\caption{Evaluations on Instruction Following Capabilities using AlpacaEval\footnotemark[2]. TT stands for training time measured by GPU hours on A100-80G GPUs.   The evaluations of the TempNet-enabled models were conducted in April, 2024 and the results of other models are from Alpaca leaderboard in April, 2024. The win rate measures the fraction of time the model's output is preferred over the GPT4\_turbo's outputs. Length controlled (LC) win-rates are a debiased version of the win-rates that control for the length of the outputs.  More details for our training can be found in Section~\ref{alpaca}. }\label{tab:aplaca}
\vspace{-0.5mm}\begin{center}
 \resizebox{0.48\textwidth}{!}{
\begin{tabular}{l|c|c|c}
\toprule
Model & TT (h) & LC Win Rate & Win Rate\\
\midrule
LLaMA2 7B (leaderboard) & $\textemdash$ & 5.35\% & 4.96\% \\
LLaMA2 7B w/ TempNet & 1.42 & \textbf{5.74}\% & \textbf{5.43}\% \\
\midrule
LLaMA2 13B (leaderboard) & $\textemdash$ & 8.43\% & 7.70\%  \\
LLaMA2 13B w/ TempNet & 3.68 &\textbf{8.58}\% & \textbf{7.73}\%\\
\midrule
LLaMA2 70B (leaderboard) & $\textemdash$ & 14.69\% & 13.89\%  \\
LLaMA2 70B w/ TempNet & 11.4 &\textbf{15.83}\% & \textbf{15.05}\%\\
%\midrule
%Gemini Pro (leaderboard) & $\textemdash$ & 24.38\% & 18.17\% \\
%GPT-4 0613 (leaderboard) & $\textemdash$ & 30.18\% & 15.76\% \\
\bottomrule
\end{tabular}}
\end{center}
\label{tab:alpaca_eval}
\end{table}

\footnotetext[2]{\url{https://tatsu-lab.github.io/alpaca_eval/}}

%Temperature scaling in a softmax-type function has an effect to smoothen or roughen the associated probabilities, which has important implications in representation learning
%and generative tasks. Higher temperatures usually lead to more uniform probabilities, while lower temperatures create more concentrated probabilities. %The higher the temperature the closer of the probability distribution to the uniform distribution. The lower the temperature the more concentrated of the associated probability distribution. 
Temperature scaling plays a critical role in softmax-type functions, as increasing the temperature leads to more uniform probabilities, while decreasing it results in more concentrated probabilities. 
This has a notable impact in generative tasks using LLMs as varying the temperature parameter in the softmax function affects the
diversity of the generated texts. For example,  creative questions would benefit from higher temperatures for diverse answers, while factual questions require lower temperatures for more homogeneous responses avoiding hallucinations~\citep{huang2023survey}.  %a creative question should have more diverse answers while a factual question should have more homogeneous answer. 
%Hence, a larger temperature would be beneficial for the former task and a smaller temperature would be better in the latter task. 
In training CLIP models, the temperature parameter in a contrastive loss controls the degree of penalization on negative pairs, which also has a profound impact on the learned representations. %Similarly, in contrastive learning, 
Data with frequent semantics benefit from using high temperatures for increasing tolerance of many false negatives with similar semantics, while data with rare semantics need low temperatures to enhance distinctiveness from other samples~\citep{qiu2023not,kukleva2023temperature}. 

\begin{figure}[t]
\centering\includegraphics[width=0.23\textwidth]{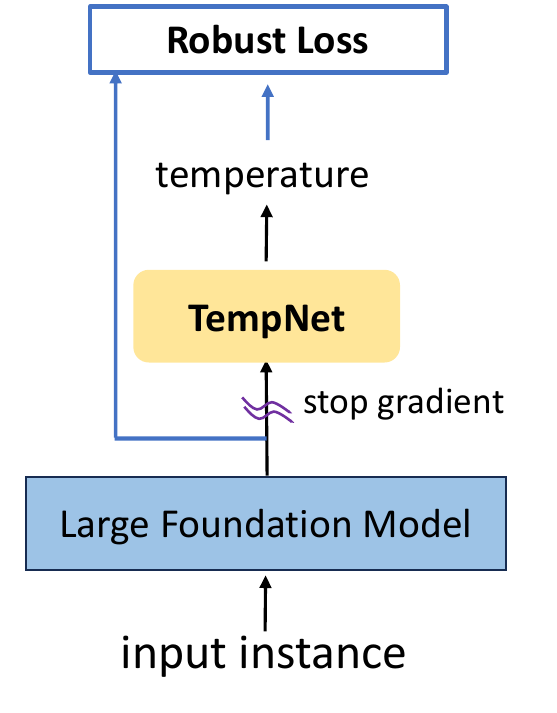}
%\centering\includegraphics[width=0.18\textwidth]{tempnet-structure-2.pdf}
\vspace*{-0.25in}\caption{Framework of Training LFMs with TempNet.}\label{fig:tmp}
%\vspace*{-0.25in}
\end{figure}

Therefore, a significant question is {\it how to set the temperature parameter during the training and/or inference.} The current practice for LLMs during inference is to set it empirically depending on tasks, e.g., instruction following tasks usually use 0.7 as the default temperature~\cite{alpaca_eval} while math reasoning tasks usually use 0.2 as the default value. If the temperature parameter is used in training (e.g., CLIP training), a naive approach is to treat it as a hyperparameter and tune it according to some performance criteria. However, this approach is problematic as: (i) tuning the temperature parameter for training LFMs is unrealistic due to the huge cost of training; (ii) a single temperature ignores the diversity and heterogeneity of data in the real world.   Hence, {\bf an approach to automatically set the temperature based on the context/semantics is highly desirable.} 

Although heuristic approaches have been adopted to tackle these issues~\cite{radford2021learning,kukleva2023temperature,DBLP:journals/corr/abs-1809-04157,hu2017toward,li2023curriculum,wang2020contextual,zhang2021temperature,manna2023dystress}, a more principled approach is needed. A {\it recent advancement}~\citep{qiu2023not} presents a novel method for contrastive learning (CL) with automatic temperature individualization. The authors presented a KL-divergence-constrained distributionally robust optimization (KL-DRO) based robust contrastive loss, incorporating an individualized temperature parameter for each anchor data optimized via Lagrangian duality theory. %The authors introduced a new robust contrastive loss based on KL-divergence-constrained distributionally robust optimization (KL-DRO)~\cite{qi2023stochastic}, and used the Lagrangian duality theory to transform it into a dual form with an individualized temperature parameter for each anchor data, acting as a Lagrangian multiplier to be optimized. 
Although this method has proven to be effective in certain CL tasks, maintaining individualized temperatures presents several challenges. It lacks scalability for large datasets due to linearly increasing memory costs and difficulties in fully optimizing individualized parameters. Additionally, this approach poses a risk of overfitting, thereby sacrificing the method's generalizability and transferability.

While a potential solution is to utilize a neural network for predicting a personalized temperature for each sample, effectively and systematically learning a temperature prediction network for multiple tasks remains a formidable challenge. In this paper, we  propose a principled learning framework to learn a small yet generalizable temperature network (called TempNet) on top of LFMs, as depicted in Figure~\ref{fig:tmp}, which integrates a DRO-based robust loss and a properly designed TempNet. Our main contributions are:

\begin{figure*}[t]
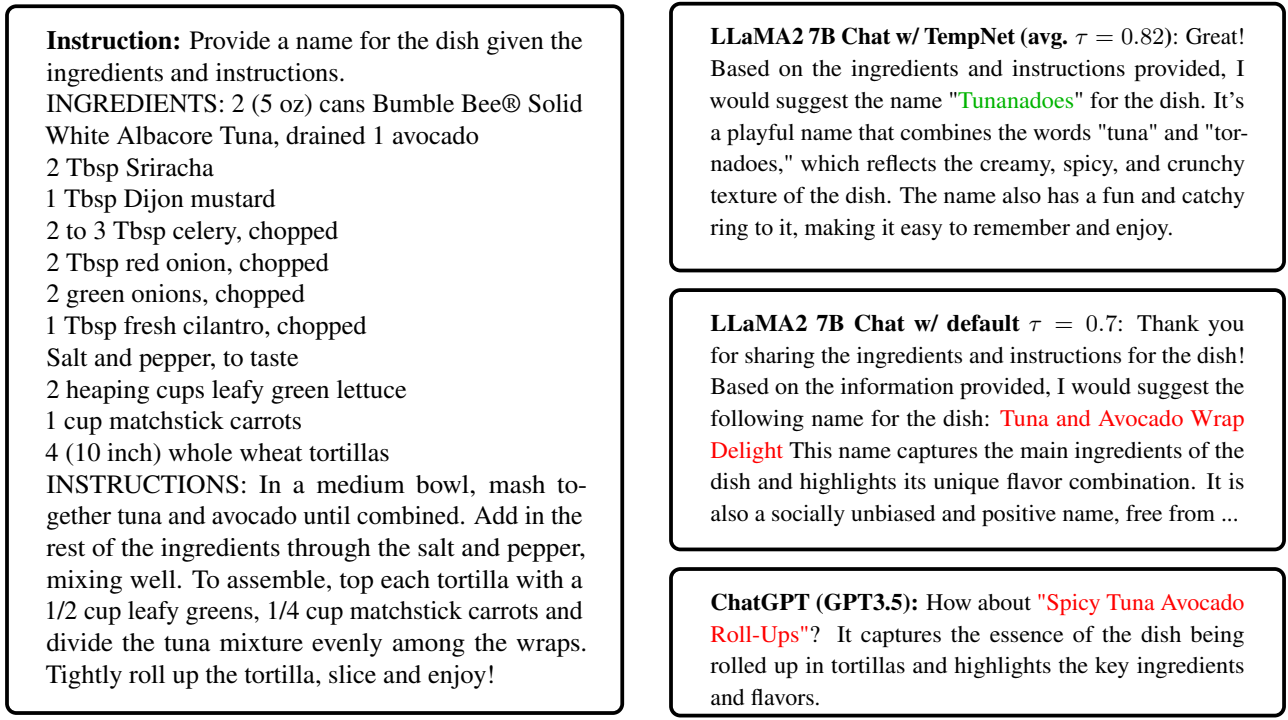

\begin{minipage}{0.48\textwidth}
\begin{tcolorbox}[colback=white,colframe=black]
    {\noindent\bf Instruction:} Provide a name for the dish given the ingredients and instructions.\\
        INGREDIENTS:
2 (5 oz) cans Bumble Bee® Solid White Albacore Tuna, drained
1 avocado\\
2 Tbsp Sriracha\\
1 Tbsp Dijon mustard\\
2 to 3 Tbsp celery, chopped\\
2 Tbsp red onion, chopped\\
2 green onions, chopped\\
1 Tbsp fresh cilantro, chopped\\
Salt and pepper, to taste\\
2 heaping cups leafy green lettuce\\
1 cup matchstick carrots\\
4 (10 inch) whole wheat tortillas\\
INSTRUCTIONS:
In a medium bowl, mash together tuna and avocado until combined. Add in the rest of the ingredients through the salt and pepper, mixing well.
To assemble, top each tortilla with a 1/2 cup leafy greens, 1/4 cup matchstick carrots and divide the tuna mixture evenly among the wraps. Tightly roll up the tortilla, slice and enjoy!
\end{tcolorbox}
\end{minipage}
\hfill
\begin{minipage}{0.48\textwidth}

\begin{tcolorbox}[colback=white,colframe=black, width=8.2cm,height=3.6cm]
{\small\textbf{LLaMA2 7B Chat w/ TempNet (avg. $\tau=0.82$)}:  Great! Based on the ingredients and instructions provided, I would suggest the name "\textcolor{darkgreen}{Tunanadoes}" for the dish. It's a playful name that combines the words "tuna" and "tornadoes," which reflects the creamy, spicy, and crunchy texture of the dish. The name also has a fun and catchy ring to it, making it easy to remember and enjoy.}
\end{tcolorbox}
\begin{tcolorbox}[colback=white,colframe=black,width=8.2cm,height=3.5cm]
{\small\textbf{LLaMA2 7B Chat w/ default $\tau=0.7$}: Thank you for sharing the ingredients and instructions for the dish! Based on the information provided, I would suggest the following name for the dish:
\textcolor{red}{Tuna and Avocado Wrap Delight}
This name captures the main ingredients of the dish and highlights its unique flavor combination. It is also a socially unbiased and positive name, free from ... %any harmful or offensive language. If you have any other questions or need further assistance, please feel free to ask! 
}
\end{tcolorbox}
\begin{tcolorbox}[colback=white,colframe=black,width=8.2cm,height=2cm]
{\small{\noindent\bf ChatGPT (GPT3.5):}
How about \textcolor{red}{"Spicy Tuna Avocado Roll-Ups"}? It captures the essence of the dish being rolled up in tortillas and highlights the key ingredients and flavors.}
\end{tcolorbox}
\end{minipage}
\vspace{-2mm}
\caption{An example of instruction following, which demonstrates that TempNet makes the LLaMA2 more creative when it should be by using personalized temperatures than using the default value of $0.7$.   The left side displays the instruction, and the right side shows responses generated by different models. More results are included in Appendix~\ref{sec:app:alpaca_eval_analysis}. We demonstrate the personalized $\tau$ for TempNet's responses in Figure~\ref{fig:tau_for_each_token}.}
\label{fig:example_alpaca}
\vskip -0.2in
\end{figure*}

\vspace{-2.5mm}
\begin{itemize}[leftmargin=*]
  \vspace{-1.2mm}  \item We propose a new framework of training LFMs by optimizing a KL-DRO based robust loss with a temperature prediction network motivated from variational analysis. 
     \vspace{-1.5mm} \item We design novel TempNets that predict a temperature based on the output of LFMs through several layers, consisting of transformation, projection, a theory inspired parameterized pooling  and an output layer.  %for training and/or inference with CLIP models and LLMs.  
     \vspace{-1.5mm} \item We conduct extensive experiments to demonstrate the effectiveness of TempNets for learning  LLMs and CLIP models in various settings, including training from scratch, finetuning LLMs, and learning TempNet only with a frozen foundation model. %In Table~\ref{tab:alpaca_eval}, we demonstrate the improvements that TempNet brings to the LLaMA2 Chat series of models. 
     \vspace{-1.5mm} \item We conduct deep empirical analysis of the generalizability of TempNets, the characteristics of predicted temperatures for different texts, and ablation studies of network design. %In Figure~\ref{fig:example_alpaca}, we illustrate through an example how TempNet generates personalized temperatures to produce better results. 
\end{itemize}
\vspace*{-0.15in}
The {\bf noticable benefits of our approach} are the following: (i) The TempNet can be efficiently trained for any LLMs with the foundation model fixed (cf. Table~\ref{tab:aplaca}), and it only has marginal overhead when trained together with large foundation models (cf. Table~\ref{tt}); (ii) The predicted temperature by TempNet is context/semantics dependent. Hence, it enables the adoption of personalized temperature during the inference of LLMs such that the model knows when it should be creative by using a relatively large temperature (cf. the example shown in Figure~\ref{fig:example_alpaca}) and knows when it should be more factual by using a relatively low temperature (cf. the example shown in Appendix~\ref{sec:app:alpaca_eval_analysis}).

\section{Related Work}
\label{sec:related_work}

{\noindent\bf Temperature scaling.} Temperature scaling in a softmax-type function plays a significant role, which has been observed and analyzed in multiple learning problems, e.g.,  knowledge distillation~\citep{hinton2015distilling}, model calibration~\cite{guo2017calibration}, reinforcement learning~\citep{ma2017softmax}, contrastive learning~\cite{wang2021understanding}, multi-class classification with noisy labels~\cite{zhu2023label},  language modeling~\cite{wang2020contextual,austin2021program,chen2021evaluating,xu2022systematic,gloeckle2023temperature}. Due to the critical importance of temperature scaling in softmax-type functions, multiple strategies for its modulation have been introduced, as reviewed below.

{\noindent\bf Adjusting a global temperature.} A naive approach for adjusting the temperature parameter is to treat it like a hyper-parameter, whose value is tuned according to some performance criteria~\cite{hinton2015distilling,guo2017calibration,ma2017softmax}. To better adapt the temperature parameter to the dynamic training process, several papers propose some schedules to adjust the temperature value during model training~\cite{hu2017toward,DBLP:journals/corr/abs-1809-04157,kukleva2023temperature,manna2023dystress}, e.g., a cosine schedule. %For example,  \citet{kukleva2023temperature} use a cosine schedule to decrease the temperature value in self-supervised contrastive learning. 

Many studies  use the gradient descent method for updating a global temperature based on some loss function in different learning scenarios, e.g., reinforcement learning~\cite{Kim2019AdaptiveTT,he2018determining}, contrastive learning~\cite{radford2021learning,cherti2023reproducible,goel2022cyclip,li2021align},  knowledge distillation~\cite{liu2022meta}.  However, the loss functions in these works for computing the gradient in terms of the temperature are usually heuristic or ad-hoc. For example, \citet{radford2021learning,li2021align,goel2022cyclip,cherti2023reproducible} simply use a mini-batch based contrastive loss to compute the gradient of $\tau$ for training vision-language models such as CLIP. 

%Although these efforts have achieved improvements by adjusting the temperature, the criteria they are based on are largely heuristic and fail to consider the individualization of the temperature parameters.

%\cite{Kim2019AdaptiveTT} use a heuristic to update the temperature.  \citet{he2018determining} propose both an evaluation function designed to measure the effectiveness of a temperature parameter and an iterative updating rule to determine the optimal temperature value.  a new evaluation function is designed to measure the effectiveness of temperature parameter by considering the information-loss of transformation and the diversity among probability vector elements. Secondly, an iterative updating rule is derived to determine the optimal temperature parameter by calculating the minimum of evaluation function %The significance of the temperature parameter in language models has consistently been a focus of researchers. \citet{hu2017toward} employ a decreasing temperature parameter in text generation model training, finding that it reduces training variance and accelerates convergence. 

% objective和network -- heuristic
{\noindent\bf Temperature Prediction.} 
The idea of using the target network to induce a function for predicting a personalized temperature of each instance has been explored in several works~\cite{li2023curriculum,wang2020contextual,zhang2021temperature,balanya2022adaptive,joy2023sample,manna2023dystress}. %As we enter the era of large models, %\citet{xu2022systematic} demonstrate how different temperatures affect the generation quality of code language models. \citet{austin2021program,chen2021evaluating} perform program synthesis with LLMs and observe that lower temperatures (more greedy decoding) perform better with only a single evaluation allowed, but higher temperature, less greedy strategies begin to solve more tasks with more sample budgets. The researchers of LLaMA 2~\citep{touvron2023llama2} notice an intriguing phenomenon that RLHF (Reinforcement Learning from Human Feedback) dynamically re-scales the temperature contingent upon the context, indicating different contexts should have different temperatures for better modeling. \citet{gloeckle2023temperature} identify temperature scaling (i.e., y dividing the output logits by a fixed $\tau$ before softmax function) as an effective method for model selection when fine-tuning large language models. Although the temperature parameter has been proven to significantly impact the training and inference of language models, there still lacks a systematic method to optimize individualized temperature parameter for each input.
\citet{wang2020contextual} consider language modeling tasks and propose to use a single projection layer that transforms the output representation of a neural language model into a vector of temperatures corresponding to each token in the vocabulary.
 \citet{zhang2021temperature} simply use one coordinate of the embedded vector as the temperature value in contrastive learning, but observe sacrificed  performance on downstream tasks. \citet{li2023curriculum} study knowledge distillation and use an adversarial training approach to learn a network for predicting the temperature parameter based on the predictions of the target network.   % propose a heuristic objective with learnable temperatures for each token in vocabulary, and learn the temperatures using a deep neural network.
%\citet{balanya2022adaptive,joy2023sample} introduce post-hoc methods for model calibration, which involve using a pretrained model to predict the temperature parameters for adjusting the distribution accordingly. %\cite{manna2023dystress} has used a cosine function to dynamically set the temperature value for each pair in the contrastive learning, which depends on the similarity of each data pair.  
In contrast to previous methods, we propose a principled training framework rooted in DRO and variational analysis, with a novel design of TempNet, leading to better improvements on both LLMs and CLIP models.

{\noindent \bf DRO-based robust losses and optimizing individualized temperatures.} To the best of our knowledge, \citet{qiu2023not} is the first work that  employs KL-DRO technique to formulate a robust contrastive loss, which induces individualized temperate parameters from the Lagrangian dual theory. They proposed a tailored algorithm for solving the resulting robust contrastive loss with individualized temperatures for self-supervised learning.  This DRO-based formulation was also adopted in~\cite{wu2023understanding,wu2023bsl} without considering how to optimize individualized temperatures.

\section{Preliminaries}
\label{sec:preliminaries}

Below, we present the softmax loss of LLMs and the global contrastive loss for training CLIP models. 

We first discuss the softmax loss for LLMs. Let $\x=(t_1, \ldots, t_{m})$ denote an example with a sequence of tokens, where $t_j\in\mathcal V=\{v_1, \ldots, v_K\}$ denotes a token from a vocabulary of size $K$.  The probability of $\x$ is modeled by $p(\x) = \prod_{j=1}^{m}p(t_j|t_{1}, \ldots, t_{j-1})$. Modern LLMs are trained by using a softmax function to model the probability of $p(t_j|t_1,\ldots, t_{j-1})$, i.e., 
\begin{align*}
p(t_j|t_1,\ldots, t_{j-1}) = \frac{\exp(h(\w; t_1, \ldots, t_{j-1})^{\top}W_{t_j})}{\sum_{k=1}^K\exp(h(\w; t_1, \ldots, t_{j-1})^{\top}W_{k})},
\end{align*}
where $W_1,\ldots, W_K$ denotes the token embedding vectors of that in $\mathcal V$, $h(\w;  t_1, \ldots, t_{j-1})$ denotes the representation of the input sequence of tokens produced by a transformer network parameterized by $\w$. We abuse the notation $W_{t_j}$ to denote the embedding vector of the token $t_j$.  The parameters $\Theta=(\w, W)$ are learned by minimizing the negative log-likelihood over a set of data $\D=\{\x_1, \ldots, \x_n\}$ with $\x_i = (t^i_1, \ldots, t^i_{m_i})$:
\begin{align}\label{eqn:llm}
\hspace*{-0.1in}\min_{\Theta}-\frac{1}{n}\sum_{i=1}^n[\log p(\x_i):=\sum_{j=1}^{m_i}\log p(t^i_{j}|t^i_{1}, \ldots, t^i_{j-1})].
\end{align}
%To the best of our knowledge, the temperature-scaled softmax loss has not been used in training LLMs. We will introduce our approach in next section.  
\iffalse 
Let $\D=\{S_1,S_2,\cdots,S_T\}$ denote the dataset containing $T$ text sequences. Let $W_{i,j}$ denote the $j$-th token in the $i$-th text sequence. Denote by $\x_{i,j}=W_{i,1},\cdots,W_{i,j-1}$ a random sequence from $S_i$. During training/inference, the LM receives a context sequence (e.g., $\x_i$) as input and outputs logits (denote by $l_i$), which are then passed through a softmax function to obtain a distribution for the next token:
\begin{equation*}
     P(W_{i,j} | \x_{i,j}; \theta) = \frac{\exp(l_i^j/\tau)}{\sum_{k=1}^{K}\exp(l_i^k/\tau)},
\end{equation*}
where $\theta$ denotes the parameters of the LM, and $\tau$ is temperature parameter controlling the smoothness of the distribution. The training objective for LMs is given by:
\begin{equation}
     \mathcal{L}(\theta)=\sum_{i=1}^{T}\sum_{j=1}^{n_i} \ell(\x_{i,j};\theta) := -\log \frac{\exp(l_i^{j^*}/\tau)}{\sum_{k=1}^{K}\exp(l_i^k/\tau)},
\label{eq:lm_ce_loss}
\end{equation}
where $j^*$ represents the ground truth label for the $j$-th token.
\fi
Next we present the contrastive loss for training CLIP models.  It is notable that contrastive losses have been widely used for pretraining encoder networks in a self-supervised learning fashion on various data types. For simplicity, we just restrict our discussion to training CLIP models on image-text data. However, our approach can be adopted for traing many other models based on the contrastive loss. % For CL, we focus on bimodal contrastive learning because it has a much wider range of applications in practice compared to unimodal contrastive learning. 
Let $\D=\{(\x_1, \t_1),\cdots,(\x_n,\t_n) \}$ denote $n$ image-text pairs. % Let $\T=\{\t_j\in\D^{\prime}\}$ be the set of texts, and $\mathcal{I}=\{\x_j\in\D^{\prime}\}$ be the set images.
%Let $\T_i^{-}=\{\t_j\in\D^{\prime},j\neq i\}$ be the set of negative texts for image $\x_i$, and $\mathcal{I}_i^{-}=\{\x_j\in\D^{\prime},j\neq i\}$ be the set of negative texts for image $\t_i$. 
Let $E_I(\cdot)$ and $E_T(\cdot)$ denote the encoder for images and texts parameterized by $\w_1$ and $\w_2$, respectively, which output normalized vector representations. A traditional approach is by minimizing a mini-batch based contrastive loss~\cite{chen2020simple}. However, this approach is {\bf sensitive to batch-size} and is also problematic for us to unify the softmax loss and contrastive loss. To address this issue,   we adopt the two-way temperature-scaled \emph{global contrastive loss (GCL)}~\citep{yuan2022provable}, which is defined below for $(\x_i, \t_i)$:
\begin{align*}
    \ell_i(\w; \D)=& -\tau_1\log\frac{\exp(E_I(\x_i)^{\top}E_T(\t_i)/\tau_1)}{\sum_{\t\in\mathcal{T}_i^{-}} \exp(E_I(\x_i)^{\top}E_T(\t)/\tau_1)} \\
    &-\tau_2\log\frac{\exp(E_I(\x_i)^{\top}E_T(\t_i)/\tau_2)}{\sum_{\x\in\mathcal{I}_i^{-}} \exp(E_I(\x)^{\top}E_T(\t_i)/\tau_2)},
\end{align*}
where $\tau_1, \tau_2$ are two temperature parameters and $\mathcal T_i^{-}$ denotes the set of all negative texts of $\x_i$ and $\mathcal I_i^{-}$  denotes the set of all negative images of $\t_i$. Thus, an objective for CL is given by averaging the above GCL over all image-text pairs. 

\section{A Robust Learning Framework with a Temperature Prediction Network}
\label{sec:functional_opt}
The proposed framework for learning a temperature prediction network builds upon robust contrastive losses with individual temperature optimization underpinned by KL-divergence-constrained DRO~\cite{qi2023stochastic,qiu2023not}. %is greatly inspired by a recent work~\cite{qiu2023not}, which proposes a robust global contrastive loss based on KL-divergence-constrained DRO with individualized temperature parameters. We use their approach to motivate our solution using TempNet.  
To this end, we first provide a unified framework to formulate the robust softmax loss and the robust global contrastive loss.  

\subsection{DRO-based Robust Losses} 
\vspace*{-0.05in}For any instance $z\in\Omega$ (e.g., a sequence of tokens in LLMs and an image or a text in CLIP training), assume we have a positive logit $L_{+}(z)$ and a finite set of contrasting logits $L_{k}(z), k\in\mathcal C$, where $\mathcal C$ is a discrete set and all logits are computed based on a deep neural network parameterized by $\Theta$. %The specific interpretation of logits in different tasks is detailed later.  
The goal of learning is to push the difference $L_{k}(z) - L_{+}(z)$ to be small for $k\in\mathcal C$. A DRO-based robust loss is: 
\begin{align*}
\ell_{\text{DRO}}(z)= \max_{\p\in\Delta} &
\sum_{k\in\mathcal C}p_k (L_{k}(z) - L_{+}(z))   - \tau_0 \text{KL}\left(\mathbf p, \frac{1}{|\mathcal C|}\right)\\
&s.t.\;\text{KL}(\mathbf p, 1/|\mathcal C|)\leq \rho,
\end{align*}
where $\Delta=\{\mathbf p: \sum_{k\in\mathcal C} p_k=1\}$ denotes a simplex,  $\tau_0>0$ and $\rho$ are two hyper-parameters shared by all instances, and $ \text{KL}(\mathbf p, 1/|\mathcal C|) = \sum_k p_k\log(p_k|\mathcal C|)$ is the KL divergence between $\mathbf p$ and the uniform distribution. By using the Lagrangian duality theory, one can derive a dual from of the above maximization problem, yielding a closed-form of the probabilities $p_k = \frac{\exp(L_{k}(z)/\tau)}{\sum_{l\in\mathcal C}\exp(L_{l}(z)/\tau)}$ and an equivalent form of the robust loss (see Appendix~\ref{sec:app:dro_robust}): 
\begin{align}\label{eqn:dro}
&\ell_{\text{DRO}}(z) = \min_{\tau\geq \tau_0} f_{\Theta}(z, \tau)\\
& f_{\Theta}(z, \tau) = \tau \log\bigg(\frac{1}{|\mathcal C|}\sum_{k\in\mathcal C}\exp\left(\frac{L_{k}(z) - L_{+}(z)}{\tau}\right)\bigg)+ \tau \rho\notag 
\vspace*{-0.1in}\end{align}
where  $\tau$ is a (shifted) Lagrangian multiplier of the KL-divergence constraint. As a result, with a set of instances $\mathcal S=\{z_1,\ldots, z_n\}$, the learning problem becomes: \\
\vspace*{-0.1in}
\begin{align}\label{eqn:rl}
%&\min_{\Theta}\frac{1}{n}\sum_{i=1}^n\sum_{j=1}^{m_i}\ell_{\text{DRO}}(t^i_j)=\\
&\min_{\Theta}\frac{1}{n}\sum\nolimits_{i=1}^n\min_{\tau_i\geq \tau_0}f_{\Theta}(z_i, \tau_i).
\end{align}
Below, we discuss two instantiations of the above robust objective for LLMs and contrastive learning of CLIP models.

\noindent$\bullet~${\bf Robust Softmax Loss for LLMs.} Let $z=t_{1:j-1}=(t_1, \ldots, t_{j-1})$ be a sequence of tokens followed by a token $t_j$. The positive logit is the prediction score $L_{+}(t_{1:j-1})=h(\w; t_{1}, \ldots, t_{j-1})^{\top}W_{t_j}$ corresponding to the next token $t_j$, and the contrasting logits are prediction scores $L_{k}(t_{1:j-1})=h(\w; t_{1}, \ldots, t_{j-1})^{\top}W_k, k\in\mathcal V$  of all tokens in the vocabulary $\mathcal V$. Hence, optimizing a robust softmax loss over a set of data $\D$ for training LLMs becomes:
\begin{align}\label{eqn:rllm}
%&\min_{\Theta}\frac{1}{n}\sum_{i=1}^n\sum_{j=1}^{m_i}\ell_{\text{DRO}}(t^i_j)=\\
&\min_{\Theta}\frac{1}{n}\sum_{i=1}^n\sum_{j=1}^{m_i}\min_{\tau_{ij}\geq \tau_0}f_{\Theta}(t^i_{1:j-1}, \tau_{ij}).
%&\text{s.t.}\: f_{\Theta}(\tau_{ij}, t^i_j) =  \tau_{ij} \log\bigg[\frac{1}{|\mathcal C|}\sum_{k\in\mathcal C}\exp(\frac{L_{k}(t^i_j) - L_{+}(t^i_j)}{\tau})\bigg]\notag\\
% & \hspace*{1in}+ \tau_{ij} \rho\notag
\end{align}
Note that if we ignore the minimization over $\tau_{ij}$ by setting it as 1 the objective becomes equivalent to that in~(\ref{eqn:llm}) of existing apporaches for training LLMs~\cite{DBLP:journals/corr/abs-2005-14165}.  

\noindent$\bullet~${\bf Robust Contrastive Loss for CLIP training.} Let $z_i = (\x_i, \t_i)$ be an image-text pair. For an image $\x_i$, we denote its positive logit by $L_+(\x_i) =  E_{I}(\x_i)^{\top}E_{T}(\t_i)$ and all contrasting logits by $L_k(\x_i) = E_{I}(\x_i)^{\top}E_{T}(\t_k), \forall \t_k\in\mathcal T_i^{-}$.  Similarly, for a text $\t_i$, we denote its positive logit as $L_+(\t_i) =  E_{T}(\t_i)^{\top}E_{I}(\x_i)$ and contrasting logits by $L_k(\t_i) = E_{T}(\t_i)^{\top}E_{I}(\x_k), \forall \x_k\in\mathcal I_i^{-}$.  %Then the robust loss $\ell_{\text{DRO}}(\x_i) + \ell_{\text{DRO}}(\t_i)$ is exactly the one proposed in~\cite{qiu2023not}, yielding 
Then, we have the following problem for contrastive learning with individualized temperatures: 
\begin{align}\label{eqn:rcl}
%&\min_{\Theta}\frac{1}{n}\sum_{i=1}^n\sum_{j=1}^{m_i}\ell_{\text{DRO}}(t^i_j)=\\
&\min_{\Theta}\frac{1}{n}\sum_{i=1}^n\min_{\tau_{i,1}, \tau_{i,2}\geq \tau_0}f_{\Theta}(\x_i, \tau_{i,1}) + f_{\Theta}(\t_i, \tau_{i,2}).
%&\text{s.t.}\: f_{\Theta}(\tau_{i}^1, \x^i_j) =  \tau_{ij} \log\bigg[\frac{1}{|\mathcal C|}\sum_{k\in\mathcal C}\exp(\frac{L_{k}(t^i_j) - L_{+}(t^i_j)}{\tau})\bigg]\notag\\
% & \hspace*{1in}+ \tau_{ij} \rho\notag
\end{align}
Different from conventional softmax loss and global contrastive loss, the robust softmax loss in~(\ref{eqn:rllm}) and the robust contrastive loss in~(\ref{eqn:rcl}) introduce a set of new temperature variables, e.g., $\{\tau_{ij}\}_{i=1,\ldots, n, j=1, \ldots, m_i}$ in LLMs, to be optimized. Hence, a tailored optimizer needs to be developed for solving both losses. \citet{qiu2023not} have proposed a stochastic algorithm named iSogCLR to optimize the robust contrastive losses due to the challenge that the number of logits $L_k(\x_i), \forall\t_k\in\mathcal T_i^{-1}$, and $L_k(\t_i), \forall\x_k\in\mathcal I_i^{-1}$ is the same size of dataset. 

{\bf Remark:} For optimizing the robust softmax loss in training LLMs, since the logits $L_k(t^i_{1:j-1}), k\in\mathcal V$ are computed every iteration, the optimization over $\tau_{ij}$ for those sampled tokens can be easliy solved by a traditional solver, e.g., the Newton Method (cf. Appendix~\ref{sec:app:newton}).

\subsection{Robust Learning with a Temperature Network}
There are several limitations of  optimizing individualized temperature parameters in the above robust losses, including lack of scalability, generalizability, and transferablity.  

\begin{itemize}[leftmargin=*]
\vspace*{-0.1in}
    \item The {\bf scalability issue} is profound in solving the robust contrastive losses by maintaining and updating individualized temperature parameters. %, where computing all logits $L_k(\x_i), \forall\t_k\in\T_i^{-}$ and $L_k(\t_i), \forall\x_k\in\mathcal I_i^{-}$ at each iteration is prohibited.  % is not scalable for big datasets.  
This is because the memory cost of maintaining individualized temperatures increases linearly as the size of the dataset, and the individualized temperature parameters may not be fully optimized due to the limits of computing resources. % to update them many times.
 \vspace*{-0.05in}   \item The {\bf lack of generalizability} is caused by potential overfitting. This is because the number of variables to be optimized is of the same size of training data. Moreover, in the real-world, the training data usually contains some noise, e.g.,  a word may be mispelled, and an image caption does not match its content.  %For example, in training LLMs, a word may be mispelled. In training CLIP models, an image may not match the associated text description, which is very common on the internet. Hence,  the positive logit $L_+(z)$ is not truly positive for a noisy instance $z$.  
Thus, minimizing the the difference $L_k(z) - L_+(z)$ for all $k$ would push the wrong positive logit $L_+(z)$ to be large. As a consequence, optimizing the robust loss with individualized temperature parameter would exacerbate such effect and harm the generalization of the learned representations.
 \vspace*{-0.05in}   \item The {\bf lack of transferablity} can be understood in the way that the learned temperatures are only for training instances that cannot be transferred to new data, and  finding the temperature for a new data requires solving an individual optimization problem, which is costly  in contrastive learning and increases inference time for using LLMs.
\end{itemize}
\vspace*{-0.05in}

One potential solution to get around these issues is to express $\tau$ as a function  of the input instance $z$, i.e., $\tau(\cdot)\in \Tau:\Omega \rightarrow \R$, such that the robust learning problem becomes:
\begin{align}\label{eqn:trcl}
%&\min_{\Theta}\frac{1}{n}\sum_{i=1}^n\sum_{j=1}^{m_i}\ell_{\text{DRO}}(t^i_j)=\\
&\min_{\Theta}\inf_{\tau\in\Tau}\frac{1}{n}\sum_{i=1}^nf_{\Theta}(z_i, \tau(z_i)).
%&\text{s.t.}\: f_{\Theta}(\tau_{i}^1, \x^i_j) =  \tau_{ij} \log\bigg[\frac{1}{|\mathcal C|}\sum_{k\in\mathcal C}\exp(\frac{L_{k}(t^i_j) - L_{+}(t^i_j)}{\tau})\bigg]\notag\\
% & \hspace*{1in}+ \tau_{ij} \rho\notag
\end{align}
It is noteworthy that the lemma below from variational analysis justifies the above objective in the infinite-sample case.
%The lemma below from variational analysis motivates  the above objective. %in the {\bf infinite-sample case} when $\Tau$ is a decomposable measurable function space. 

\begin{lemma}[\citet{rockafellar2009variational},Theorem 14.60]

Let $\Tau$ be a space of measurable functions from $\Omega$ to $\R$ that is decomposable relative to a $\sigma$-finite measure $\mu$ on the $\sigma$-algebra $\A$. Let $f:\Omega\times\R\rightarrow\R$ be a normal integrand. Then, as long as $\int_{z\in\Omega} f(z,\tau(z))\mu({\rm d}z)\neq \infty, \forall \tau(\cdot)\in\Tau$, we have
\begin{equation*}
    \inf_{\tau\in\Tau}\int_{z\in\Omega} f(z, \tau(z))\mu({\rm d}z)=\int_{z\in\Omega}\left(\inf_{\tau\in\R}f(z, \tau)\right)\mu({\rm d}z).
\end{equation*}
Moreover, as long as the above infimum is not $-\infty$, we have that $\tau^{\prime}\in\argmin_{\tau\in\Tau} \int_{z\in\Omega} f(z, \tau(z))\mu({\rm d}z)$ if and only if $\tau^{\prime}(z)\in \argmin_{\tau\in\R} f(z, \tau)$ for $\mu$-almost every $z\in\Omega$.
\end{lemma}
\vspace*{-0.1in}
{\bf Remark:} We refer the readers to~\citet{rockafellar2009variational} for more exposition of definitions of decomposable space and normal integrand. In fact, we can map the extended function $\tilde f_\Theta(z, \tau)=f_\Theta(z, \tau) + \delta_{\tau\geq \tau_0}(z, \tau)$  into $f(z, \tau)$  in the above lemma such that the pointwise minimization can be solved by optimization over a functional space. 

%Nevertheless, the condition that $\Tau$ is a decomposable and measurable function space is a strong condition, and learning $\tau(\cdot)$ from an arbitrary decomposable functional space is also not viable. Indeed, it is  not necessary given that (i) we only have a finite number of samples rather than infinite number of samples as in the above lemma; (ii) we would like to use backpropagation to learn a parametric function $\tau(\cdot)$ implemented by a neural network that can be generalizable and transferable. 
Because neural networks have been extensively used as function approximators~\citep{HornikEtAl89}, exhibiting strong capabilities in function fitting and generalization. Therefore, {\bf our proposed solution} is solving the robust losses with a temperature prediction network:
\begin{align}\label{eqn:trcl-1}
&\min_{\Theta}\min_{\w'}\frac{1}{n}\sum_{i=1}^nf_{\Theta}(z_i, \tau_{\w'}(z_i)).
\end{align}
where $\w'$ is the parameter of the temperature prediction network $\tau_{\w'}(\cdot):\Omega\rightarrow\R^{+}$. The  lemma below shows that the above objective is a relaxed upper bound of that in~(\ref{eqn:rl}). 
%\vspace*{-0.05in}
\begin{lemma} If the network is designed such that $\tau_{\w'}(z)\geq \tau_0$, then
$\frac{1}{n}\sum_{i=1}^n\min_{\tau_i\geq \tau_0}f_{\Theta}(z_i, \tau_i)\leq \min_{\w'}\frac{1}{n}\sum_{i=1}^nf_{\Theta}(z_i, \tau_{\w'}(z_i))$.
\label{lemma:upper_bound}
\end{lemma}
%\vspace*{-0.05in}
%The next question to be addressed is how to design the structure of the temperature prediction network. 

\section{TempNet Design}
\label{sec:tempnet_design}
The design of a proper temperature prediction network %is a daunting task as the search space of network structures is large.
presents a significant challenge due to the extensive search space of network structures. %In order to make the training and inference efficient, we adopt the following principles in the design of TempNet.  
To ensure efficiency in both training and inference, certain guiding principles have been incorporated into TempNet's design. First, TempNet shares common layers of underlying LFMs such that the size of TempNet can be greatly reduced. Second, TempNet operates directly on the output of the underlying foundation models, i.e., the output logits of LLMs and the normalized embedding vector of CLIP. The benefit of doing this is that the foundation model can be updated or frozen while learning the TempNets. Third, TempNet is inductive biased so that the generalization can be improved~\citep{mitchell1980need}. %There are many inductive biased techniques that have been used in the design of deep neural networks, e.g., skip connection in ResNet, normalization layers in almost all modern networks. 
Various inductive bias techniques are used in neural network design, such as skip connections in ResNet and normalization layers, which are ubiquitous in contemporary networks.

To begin with, we introduce a lemma that states an implicit form of the optimal $\tau\in\R$ that minimizes $f(z, \tau)$ in~(\ref{eqn:rl}).  
\begin{lemma}
Given all contrasting logits $L_k(z), k\in\mathcal C$ for any $z$, then $\tau_*=\arg\min_{\tau}f_{\Theta
}(z, \tau)$ satisfies:\\ 
\begin{equation}
\tau_* = \frac{1}{\rho} \left[\sum_{k\in\mathcal C}\bigg(\frac{\exp\left(\frac{L_{k}(z) }{\tau_*} \right)}{\sum_{j\in\mathcal C} \exp\left(\frac{L_{j}(z)}{\tau_*}\right)} - \frac{1}{|\mathcal C|}\bigg)L_{k}(z) - b_z\right],
\label{eq:formulation_tempnet}
\end{equation}
with $b_z = \tau_*\log\bigg(\frac{1}{|\mathcal C|}\sum_{k\in\mathcal C}\exp\left(\frac{L_{k}(z)}{\tau_*}\right)\bigg)   - \E_{k\in\mathcal C} L_k(z)$.
\label{lemma:formulation_tempnet}
\end{lemma}
%\vspace*{-0.25in}
\vspace*{-0.15in}{\bf Remark:} The above lemma shows how the optimal temperature without any constraints is pooled from the logits.  The first term in the bracket is a difference between attention-based pooling and average pooling of logits, and $b_z$ is the difference between a softmax-based pooling and average pooling of logits. The proof  is deferred to Appendix~\ref{sec:app:lemma_proof}.

%We will use a neural network to mimic the first term and treat $b_z$ as a global bias term.

%We present the structures of TempNet used in LLMs and contrastive learning in Figure~\ref{fig:tmp}. TempNets for LLMs and contrastive learning share a common structure, consisting of three blocks namely a transformation/projection block, a  pooling layer and an output layer.  

\begin{figure}[t]
\centering\includegraphics[width=0.2\textwidth]{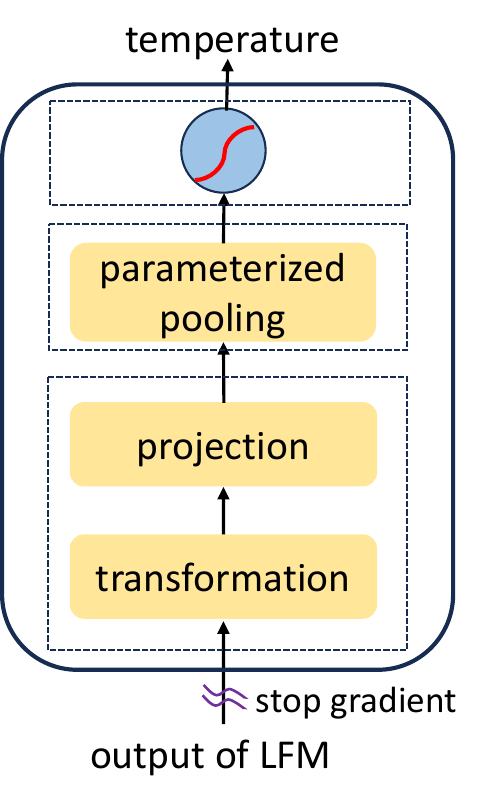}
\vspace*{-0.18in}\caption{Structure of TempNet.}\label{fig:tmp2}
%\vspace*{-0.25in}
\end{figure}

Then, we design TempNet by multiple layers to mimic the computation in~(\ref{eq:formulation_tempnet}). The general architecture of TempNet %for LLMs and contrastive learning are 
is depicted in Figure~\ref{fig:tmp2}. One can observe that there are three primary components, which will be elaborated below. %TempNets operate directly on the output of the underlying foundation models. The benefit of doing this is that the foundation model can be updated or frozen while learning the TempNets.  There are three primary components: a transformation/projection block, a parameterized pooling layer, and an output layer. The transformation/projection block has one transformation layer (a standard feed-forward layer) and a projection layer (that simply transforms the input vector by a matrix-vector product). 

{\bf The transformation-projection block.} The block contains a transformation layer (a standard feed-forward layer) and a projection layer (transforms the input vector by a matrix-vector product). This block serves two purposes: (i) reducing dimensionality from outputs of LFMs; (ii) extracting semantic information useful for temperature prediction. %The reason we need this block at the beginning is that using the outputs of LLMs and CLIP models directly for computation is problematic, as elaborated below.

\begin{itemize}[leftmargin=*]
\vspace*{-0.1in}
    \item {\bf For CL:} Due to that it is prohibitive to compute all logits  for each image and text at each iteration,  our idea is to use the transformation-projection block to compute prototypical logits with a much smaller size for images and texts separately such that a similar pooling operation as~(\ref{eq:formulation_tempnet}) can be applied. In the following, we present the discussion for processing images, noting that texts can be processed similarly.  To this end, we first transform the input normalized embedding $E_I(\x_i)\in\R^{d_0}$ into another vector $\v = \sigma(W'_1E_I(\x_i)+b_1)\in\R^{d_1}$ by a feed-forward layer with parameters $W_1'\in\R^{d_1\times d_0}, b_1\in\R^{d_1}$. Then we apply a projection layer to compute normalized prototypical logits $\u = \bar W_2'^{\top} \v\in\R^{d_2}$, where $\bar W_2'=(\bar\w'_{2,1}, \ldots, \bar\w'_{2,d_2})$ is a normalized parameter matrix of $W'_2=(\w'_{2,1}, \ldots, \w'_{2,d_2})\in\R^{d_1\times d_2}$ such that $\bar\w'_{2,k}=\w'_{2,k}/\|\bar\w'_{2,k}\|_2$. We interpret the column vectors in $W'_2$ as prototypes such that $\u=(u_1,\ldots, u_{d_2})$ represents prototypical logits. The size of each layer's output will obey $1\leq d_2\leq d_1\leq |\mathcal C|$. %We set $d_2\leq d_1\leq d$. 
    \item {\bf For LLMs:} The difference between TempNet for LLMs and that for CL lies at how to compute prototypical logits $\u$. %The parameterized pooling layer and the output layer are exactly the same. 
    In LLMs, the logits $L_k(z), k\in\mathcal V$ are computed for each instance at every iteration. However, the logit vector $L(z)=(L_1(z), \ldots, L_{d_0}(z))\in\R^{d_0}$ could be high-dimensional in the same size of vocabulary  $d_0=|\mathcal V|$ (e.g., 32000 in LLaMA2). Hence, we first apply a transformation to a normalized logit vector to generate a lower dimensional vector $\bar L(z)$ by $\v = \sigma(W_1'\bar L(z) + b_1)\in\R^{d_1}$, where $\bar L(z) = L(z)/\|L(z)\|_2$. Then, we apply a projection to generate $\u = W_2' \v\in\R^{d_2}$. Similarly, we expect the size of each layer's output to obey $d_0\geq d_1\geq d_2\geq 1$.  
\end{itemize}

{\bf The parameterized pooling layer.} Subsequently, we employ a parameterized pooling method to convert $\u$ to $s$ mimicking (\ref{eq:formulation_tempnet}): 
\begin{align*}
s =  \frac{1}{\rho} \left[\sum_{k=1}^{d_2}\left(\frac{\exp\left(\frac{u_k}{\phi} \right)}{\sum_{k=1}^{d_2} \exp\left(\frac{u_k}{\phi}\right)} - \frac{1}{|d_2|}\right)w'_{3,k}u_k - b\right]
\end{align*}
where $\w'_3=(w'_{3,1}, \ldots, w'_{3,d_2})$ and $b, \phi$ are learnable parameters, respectively. It is notable that the above pooling operation is \emph{inductive-biased} similar to~(\ref{eq:formulation_tempnet}) expect that the scaled logits $L_k(z)/\tau_*,\forall k$ are replaced by scaled prototypical logits $u_k/\phi, k=1,\ldots, d_2$.
Our ablation studies demonstrate the effectiveness of this layer.
%with a learnable parameter $\phi$. In our experiments, we find that using the inductive-biased weight before $w'_{3,k}$ $\frac{\exp\left(\frac{u_k}{\phi} \right)}{\sum_{k=1}^{d_2} \exp\left(\frac{u_k}{\phi}\right)} - \frac{1}{|d_2|}$ 
%is helpful for improving the generalization. We expect the size of each layer's output to obey $d_0\geq d_1\geq d_2\geq 1$.  

{\bf The output layer.} At the end of the TempNet, the output layer converts $s$ into a proper range $[\tau_0, \tau_{\max}]$ by $\tau = (\tau_{\max} - \tau_0)\ \sigma(s) + \tau_0$, where $\sigma(\cdot)$ is a sigmoid function.

Finally, we introduce the training algorithm of our proposed framework. The TempNet is learned by backpropagation. We apply stop-gradient to the input of the TempNet such that it does not affect the training dynamics of the foundation models except for the impact of the temperature parameter. The TempNet can be trained in three different ways: (i) joint training of TempNet and the foundation model from scratch; (ii) joint training of TempNet and the foundation model from a pretrained foundation model; (iii)  training TempNet separately with a given foundation model. In all scenarios, the AdamW optimizer~\citep{loshchilov2017decoupled} with a consine learning rate schedule is used to update the involved network parameters. The initialization of TempNet parameters is discussed in Appendix~\ref{subsec:app:experiments_implementation}. 

%{\bf Training Algorithm.} The TempNet is learned by backpropagation. We apply stop-gradient to the input of the TempNet such that it does not affect the training dynamics of the foundation models except for the impact of the temperature parameter. The TempNet can be trained in three different ways: (i) joint training of TempNet and the foundation model from scratch; (ii) joint training TempNet and the foundation model from a pretrained foundation model; (iii)  training TempNet separately with a given foundation model. In all scenarios, the AdamW optimizer~\citep{loshchilov2017decoupled} with a consine learning rate schedule is used to update the involved network parameters. The initialization of TempNet parameters is discussed in Appendix~\ref{subsec:app:experiments_implementation}. 

%p{5cm}|p{0.9cm}p{1.4cm}p{1.0cm}p{1.0cm}p{1.1cm}|p{1.2cm}p{1.2cm}

\begin{table*}[t]
\vspace{-4mm}
\caption{Results of training LLMs in various settings, including training from scratch, finetuning a pretrained LLM model,
and learning TempNet only with a frozen LLM
model. %For the first two settings, we use the GPT-2 model, while for the last setting, we employ the LLaMA2 model. We employ lm-evaluation-harness library~\citep{eval-harness} for evaluation. 
More results are in Appendix~\ref{sec:app:experiments}.}
\vspace{-2mm}
\begin{center}
 \resizebox{\textwidth}{!}{
\begin{tabular}{lllllll|lc}%{l|p{1.0cm}p{1.0cm}p{1.0cm}p{1.0cm}p{1.0cm}p{0.9cm}|p{1.3cm}p{0.7cm}}
\toprule
\multirow{2}{*}{\thead{Setting}} &
\multicolumn{6}{c}{\thead{Common Sense Reasoning (acc(\%)$\uparrow$)}} & 
\multicolumn{2}{c}{\thead{Language Modeling (ppl $\downarrow$)}} \\
\cmidrule(lr){2-7}
\cmidrule(lr){8-9}
& PIQA & HellaSwag & ARC-e & ARC-c & OBQA & Average & Lambada & \multicolumn{1}{r}{Wikitext} \\
\midrule
Training GPT-2  w/ $\tau=1.0$ & 60.9$\pm$1.1  & \textbf{26.7}$\pm$0.4  & 39.2$\pm$1.0  & 16.5$\pm$1.1  & 13.9$\pm$1.6 & 31.44 & 62.49$\pm$2.70 & 49.86 \\
Training GPT-2 w/ TempNet  & \textbf{61.1}$\pm$1.1  &   26.5$\pm$0.4 & \textbf{40.3}$\pm$1.0  & \textbf{18.1}$\pm$1.1  & \textbf{15.2}$\pm$1.6  & \textbf{32.24} & \textbf{60.13}$\pm$2.43 & \textbf{47.32}\\
\midrule
Finetuning GPT-2 w/ $\tau=1.0$ & 64.9$\pm$1.1  & 31.8$\pm$0.4  & 47.8$\pm$1.0   & 18.3$\pm$1.2  & 14.3$\pm$1.6 & 35.42 & 35.37$\pm$0.42 & 28.12 \\
Finetuning GPT-2 w/ TempNet  & \textbf{65.4}$\pm$1.1  &  \textbf{32.1}$\pm$0.4  & \textbf{49.4}$\pm$1.0  & \textbf{20.1}$\pm$1.2  & \textbf{15.4}$\pm$1.6 & \textbf{36.48} & \textbf{34.16}$\pm$0.42 & \textbf{27.06} \\
\midrule
Fixing LLaMA2 7B w/ $\tau=1.0$  & 78.0$\pm$1.0 & 57.0$\pm$0.5 & 75.5$\pm$0.9 & 42.7$\pm$1.4 & 32.8$\pm$2.1 & 57.20 & 4.01$\pm$0.08 & \textbf{11.4} \\
Fixing LLaMA2 7B w/ TempNet  & \textbf{78.8}$\pm$1.0 & \textbf{61.7}$\pm$0.5  & \textbf{76.9}$\pm$0.9 & \textbf{45.2}$\pm$1.4 & \textbf{34.6}$\pm$2.1  & \textbf{59.44} & \textbf{3.21}$\pm$0.07 & 12.2 \\
\midrule
Fixing LLaMA2 13B w/ $\tau=1.0$  &  79.5$\pm$1.0  &  60.2$\pm$0.5  &  78.7$\pm$0.8  & 47.0$\pm$1.5  &  34.6$\pm$2.1 & 60.00 &  3.62$\pm$0.07  & \textbf{10.0}  \\
Fixing LLaMA2 13B w/ TempNet  &  \textbf{79.7}$\pm$0.9  &  \textbf{62.7}$\pm$0.5   &  \textbf{79.5}$\pm$0.8  &  \textbf{48.1}$\pm$1.5  &  \textbf{36.2}$\pm$2.2 & \textbf{61.24} &  \textbf{2.97}$\pm$0.07  &  11.1 \\
\bottomrule
\end{tabular}}
\end{center}
\label{tab:language_model_results_main}
\vspace{-4mm}
\end{table*}

\begin{table*}[t]
\vspace{-4mm}
\caption{Results on contrastive learning. For image-text retrieval on Flickr30K and MSCOCO, we compute IR@$1$ and TR@$1$ for the Recall@$1$ on image-retrieval (IR) and text-retrieval (TR). For classification tasks, we compute top-$1$ accuracy (\%). We report the average of scores and standard deviation over 3 runs with different random seeds. More results are in Tables~\ref{tab:bimodal_zs_retrieval_full_flickr},~\ref{tab:bimodal_zs_retrieval_full_coco}, and~\ref{tab:bimodal_zs_classification_full} in Appendix~\ref{sec:app:experiments}.}
\vspace{-4mm}
\label{tab:bimodal_results_main}
\vskip 0.2in
\begin{center}
\begin{small}
\begin{sc}
\renewcommand{\arraystretch}{0.6}
\begin{tabular}{p{1.4cm}p{1.7cm}p{1.7cm}p{1.7cm}p{1.7cm}|p{1.7cm}p{1.7cm}p{1.7cm}}
\toprule
\multirow{1}{*}{\thead{Method}} &
\multicolumn{2}{c}{\thead{Flickr30K Retrieval}} &
\multicolumn{2}{c}{\thead{MSCOCO Retrieval}} &
\multicolumn{3}{c}{\thead{Zero-shot Classification top-1 Acc}} \\
\cmidrule(lr){2-3}
\cmidrule(lr){4-5}
\cmidrule(lr){6-8}
& IR@1 & TR@1 & IR@1 & TR@1 & CIFAR10 & CIFAR100 & ImageNet1K  \\
\midrule
CLIP & 40.98$\pm$0.22  & 50.90$\pm$0.17  & 21.32$\pm$0.12  & 26.98$\pm$0.21 & 60.63$\pm$0.19 & 30.70$\pm$0.11 & 36.27$\pm$0.17 \\
CyCLIP & 42.46$\pm$0.13 & 51.70$\pm$0.23 & 21.58$\pm$0.19 & 26.18$\pm$0.24 & 57.19$\pm$0.20 & 33.11$\pm$0.14 & 36.75$\pm$0.21 \\
SogCLR & 43.32$\pm$0.18 & 57.18$\pm$0.20 & 22.43$\pm$0.13 & 30.08$\pm$0.22 & 61.09$\pm$0.24 & 33.26$\pm$0.12 & 37.46$\pm$0.19 \\
iSogCLR & 44.36$\pm$0.12 & 60.20$\pm$0.26 & 23.27$\pm$0.18 & 32.72$\pm$0.13 & 58.91$\pm$0.15 & 33.81$\pm$0.18 & 40.72$\pm$0.23 \\
\midrule
\makecell[l]{TempNet} & \textbf{46.17}$\pm$0.14 &  \textbf{62.51}$\pm$0.19  &  \textbf{24.83}$\pm$0.16  &  \textbf{34.50}$\pm$0.16 &  \textbf{61.77}$\pm$0.18 &  \textbf{34.69}$\pm$0.17 &  \textbf{42.28}$\pm$0.19  \\
\bottomrule
\end{tabular}
\end{sc}
\end{small}
\end{center}
\vskip -0.2in
\end{table*}

%{\bf Initialization.} For the parameters of the transformation layer in TempNet, we employ the widely used kaiming uniform initialization~\citep{he2015delving} in neural networks. The projection layer in TempNet is designed to generate prototypical logits, hence in the context of contrastive learning tasks, we utilize the initial representations of randmoly sampled images/texts to initialize the text/image TempNet's projection layer. For the TempNet associated with LLMs, the projection layer is initialized using the kaiming uniform method. The parameter $\w_3^{\prime}$ in the parameterized pooling, intended to autonomously weigh different components in the prototypical logits, is initialized as an all-ones vector. We set $\phi$ to the temperature values commonly used in specific tasks, such as 1.0 for LLMs and 0.01 for CLIP models. Additionally, we initialize $b$ to 0.

\section{Experiments}
We conduct experiments on LLMs and bimodal CL to illustrate the effectiveness and provide deep analysis of the proposed approach. The details of datasets and training setup are deferred to Appendix~\ref{subsec:app:data_details} and~\ref{subsec:app:training_details}, respectively. %by comparing our method with existing methods evaluated on multiple benchmark tasks. We also verify the generalization capabilities of TempNet in both contrastive learning and language tasks by comparing with approaches using individualized temperatures. Moreover, we study the characteristics of predicted temperatures in LLMs, and evaluate the ability of instruction following of LLMs enabled by TempNet.   Finally, we provide 

\subsection{Experiments on LLMs}\label{sec:exp-llm}
{\noindent\bf Setup.} We consider three experimental settings: (i) training a LLM from scratch with TempNet; (ii) finetuning a LLM with a TempNet; (iii) learning a TempNet with a fixed LLM.  For all settings, we use the OpenWebText2~\citep{gao2020pile} dataset for training, which contains about 17M documents of 66G from Reddit. It is notable that this dataset is relatively small compared with those used for training state-of-the-art LLMs. However, we emphasize that the major goal of the experiments is a proof of concept and obtaining state-of-the-art performance by using enormous data is out of our scope because of limited available computing resources. Indeed, it is good enough for learning a good TempNet given a fixed LLM compared with using a larger dataset (cf. Table~\ref{tab:dataset_tempnet}).  %Most experiments are conducted on a cluster with low-end Nvidia GPUs, including 4090 (24GB memory), 3090 (24GB memory) and A6000 GPUs (48GB memory). %, except that TempNet for LLaMA-2 70B is trained on A100 with 80GB memory. 

For setting (i),  we train a GPT-2 model from scratch using the GPT-NeoX library~\citep{gpt-neox-library} with an initial learning rate 6e-4 and a total of 320k iterations. For setting (ii), we finetune a pretrained GPT-2 model from Pythia~\citep{biderman2023pythia} with an initial learning rate of 1e-5 and a total of 50k iterations. For setting (iii), we experiment with several LLaMA models~\citep{touvron2023llama1,touvron2023llama2}, including LLaMA1-7B,  LLaMA2-7B and LLaMA2-13B. The widths of the transformation and projection layers of all TempNets are set to $d_1=d_2=256$. This means that the TempNet only has 8.2M parameters. The values for $\tau_0$ and $\tau_{\max}$ are set to $0.001$ and $2.0$. The parameter $\rho$ in our robust loss and the learning rate for TempNet are tuned within the ranges of $\{9.0, 9.5, 10.0, 10.5\}$ and $\{$5e-4, 1e-4, 5e-5, 1e-5$\}$, respectively. The criteria for tuning these two parameters is to simply ensure the final averaged temperature value fall in $[0.7, 1.0]$, a typical range  for $\tau$ in the LLMs generative tasks~\citep{chen2021evaluating,touvron2023llama2}. The impact of $\rho$ on the performance is shown in Table~\ref{tab:influence_rho}, which indicates that $\rho=10$ is a good value yielding an average temperature value about 0.81.    

%While training of the GPT-2 model, we employed the GPT-NeoX library~\citep{gpt-neox-library} and adopted two training scenarios: training from scratch and fine-tuning on a pretrained model from Pythia~\citep{biderman2023pythia}. The respective iteration numbers and initial learning rates for these scenarios are set at 320k with 6e-4 and 50k with 1e-5. We utilize the Adam optimizer along with a cosine learning rate decay. In training TempNet based on fixed LLaMA models, the initial learning rate was set to 2e-4 with 10k iterations. For training TempNet on fixed LLaMA-chat models, the initial learning rate is set to 3e-4, and the model is trained for 10 epochs on the Alpaca 52k dataset. The parameter $\rho$ in the objective and the learning rate of the TempNet are tuned within the ranges of $\{9.0, 9.5, 10.0, 10.5\}$ and $\{5e-4, 1e-4, 5e-5, 1e-5\}$, respectively. The default values for $\tau_0$ and $\tau_{\text max}$ are set to 0.001 and 2.0.

%{\noindent\bf Data.} In the training of the GPT-2 (125M) model~\citep{radford2019language} and TempNet based on fixed LLaMA models~\citep{touvron2023llama1,touvron2023llama2}, we utilized the OpenWebText2~\citep{gao2020pile} dataset, which contains about 17M documents from Reddit. During the testing phase, we directly utilized the lm-evaluation-harness library~\citep{eval-harness}, which provides the datasets for our study. More details are presented in the subsequent sections.

{\noindent\bf Evaluation \& Results.} We evaluate the learned TempNet-enabled LLMs (T-LLMs) on   common-sense reasoning tasks and language modeling tasks. In common sense reasoning tasks, we use multiple multi-choice Q\&A datasets (cf. Appendix~\ref{subsec:app:data_details}) and report accuracy.  In language modeling tasks, we use two commonly used datasets Lambda and Wikitext and report perplexity. For TempNet-enabled LLMs, we compute the perplexity using the temperature-scaled probabilities using the temperature predicted by the learned TempNet.   For all evaluations, we utilize the lm-evaluation-harness library~\citep{eval-harness}. 

\begin{figure}[t]
\begin{minipage}[c]{0.235\textwidth}
\centering\includegraphics[width=1\textwidth]{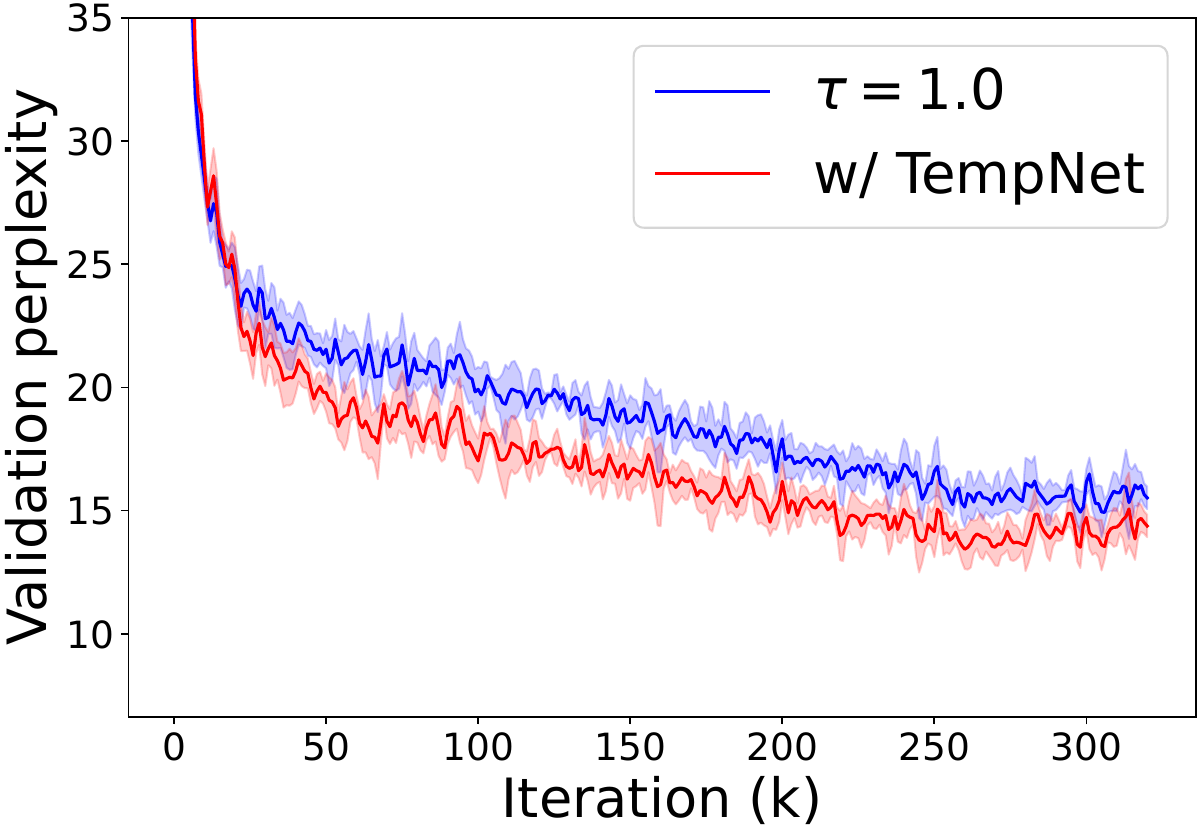}
\end{minipage}
\begin{minipage}[c]{0.235\textwidth}
\centering\includegraphics[width=1\textwidth]{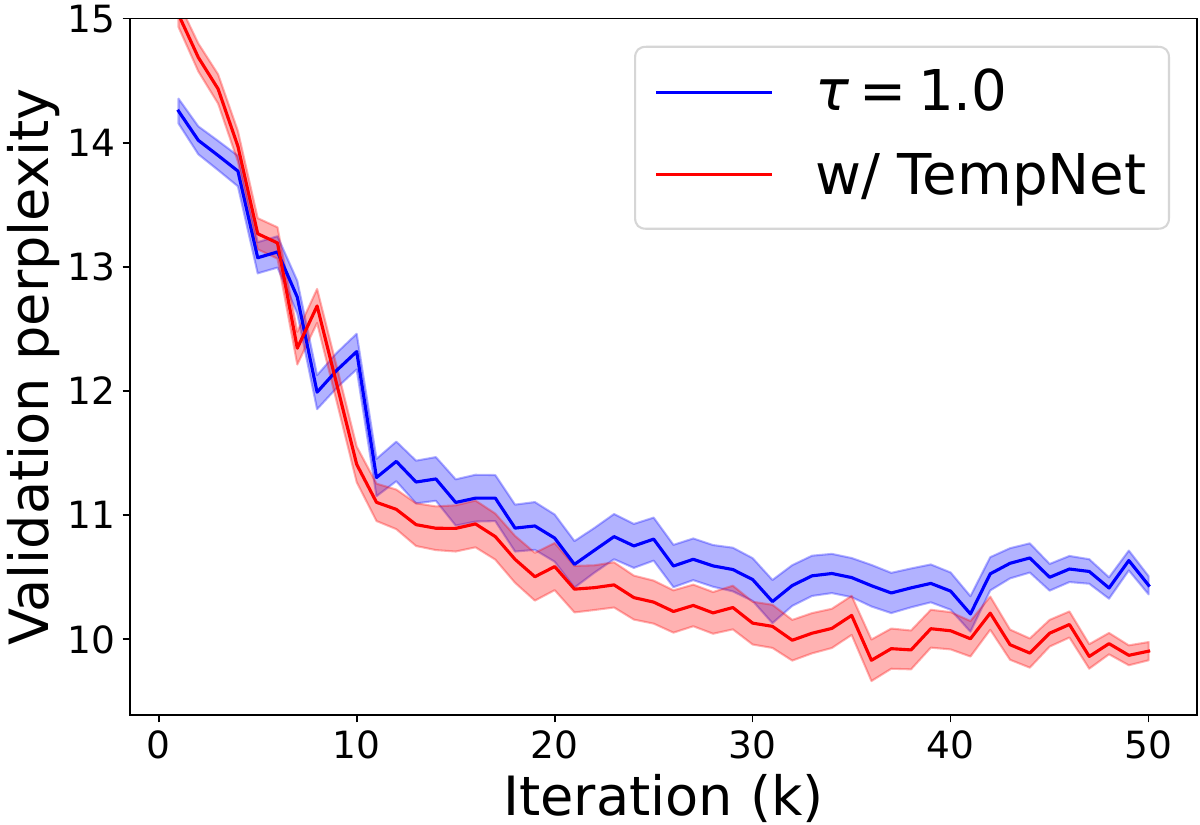}
\end{minipage}
\vspace{-0.3cm}
\caption{Left: training GPT-2. Right: fine-tuning GPT-2.}
\label{fig:training_curves_gpt2_ppl}
\vspace{-0.0cm}
\end{figure}

We present the results in Table~\ref{tab:language_model_results_main}, %and full results in Table~\ref{tab:gpt2_results_full} and~\ref{tab:llama_results_full} in Appendix~\ref{subsec:app:additional_results}, respectively. 
and have the following observations: (1) in all three settings, LLMs with TempNet have considerable improvements on almost all tasks over their variants with a fixed $\tau=1$, except for two cases.  For example, the TempNet-enabled LLaMA2-7B model  achieves $4.7\%$ and $1.5\%$  improvements on HellaSwag and ARC-c datasets over the  baseline, respectively;  
%comparing with $\tau=1.0$, we observe that (1) TempNet achieves measurable improvements in the models' performance on common sense reasoning tasks. Specifically, the LLaMA2 7B model with TempNet achieves 8.24\% and 5.85\%  improvements on HellaSwag and ARC-c datasets over baseline, respectively, confirming the effectiveness of TempNet; and 
(2) TempNet-enabled LLaMA2-7B is better than the original LLaMA2-13B model on HellaSwag and Lambda and is competitive on other tasks. This is a dramatic improvement considering that TempNet has only 8.2M parameters and its training cost with fixed LLaMA2-7B  is about 3 GPU hours on Nvidia 4090. % (costs about \$1.3).  %TempNet is also beneficial in language modeling tasks, where a decrease in perplexity indicates a model's improved proficiency in predicting the next word in a sequence. 
In Fig.~\ref{fig:training_curves_gpt2_ppl}, we visualize the training curves regarding perplexity for setting (i) and (ii). One can observe that TempNet consistently leads to more efficient and effective model optimization process. More training curves are in Fig.~\ref{fig:training_curves_gpt2} in Appendix~\ref{subsec:app:additional_results}.

\begin{table*}[t]
\vspace{-4mm}
\caption{%We demonstrate the performance of two optimization approaches in training the GPT-2 model from scratch: TempNet and Cross Entropy with optimized temperatures (w/ Opt $\tau$), which are obtained using the Newton method. 
 %Evaluations on trained GPT-2 from scratch with TempNet vs with Individually Optimized $\tau$.
The relative importance of TempNet and DRO-based robust loss for training GPT-2.}
\vspace{-0mm}
\begin{center}
 \resizebox{\textwidth}{!}{
\begin{tabular}{lllllll|lc}%{l|p{1.0cm}p{1.0cm}p{1.0cm}p{1.0cm}p{1.0cm}p{0.9cm}|p{1.3cm}p{1.5cm}}
\toprule
\multirow{2}{*}{\thead{Setting}} & 
\multicolumn{6}{c}{\thead{Common Sense Reasoning (acc(\%)$\uparrow$)}} &
\multicolumn{2}{|c}{\thead{Language Modeling (ppl$\downarrow$)}} \\
\cmidrule(lr){2-7}
\cmidrule(lr){8-9}
& PIQA & Hellaswag & ARC-e & ARC-c & OBQA & Average & Lambada & Wikitext \\
\midrule
Baseline1 (No TempNet, No DRO loss) & 60.9$\pm$1.1 &	\textbf{26.7}$\pm$0.4 & 39.2$\pm$1.0 & 16.5$\pm$1.1 & 13.9$\pm$1.6 & 31.44 & 62.49$\pm$2.70 & 49.86 \\
\cmidrule(lr){1-9}
Baseline2 (No TempNet, DRO loss)  & 60.5$\pm$1.1 & 26.3$\pm$0.4 & 39.8$\pm$1.0 & 17.7$\pm$1.1 & 14.6$\pm$0.3 & 31.78 & 62.17$\pm$2.65 & 49.13 \\
\cmidrule(lr){1-9}
Baseline3 (TempNet, No DRO loss) & 60.8$\pm$1.1 &	\textbf{26.7}$\pm$0.4 & 39.5$\pm$1.0 & 17.5$\pm$1.1 & 14.3$\pm$1.6 & 31.76 & 61.48$\pm$2.56 & 48.61 \\
\cmidrule(lr){1-9}
Ours (TempNet, DRO loss) & \textbf{61.1}$\pm$1.1 & 26.5$\pm$0.4 & \textbf{40.3}$\pm$1.0 & \textbf{18.1}$\pm$1.1 & \textbf{15.2}$\pm$1.6 & \textbf{32.24} & \textbf{60.13}$\pm$2.43 & \textbf{47.32} \\
\bottomrule
\end{tabular}}
\end{center}
\label{tab:gpt2_ce_with_optimized_tau_part}
\vspace{-4mm}
\end{table*}

\subsection{Experiments on Bimodal Contrastive Learning}
{\noindent\bf Setup.} 
We conduct experiments on  CC3M data~\citep{sharma2018conceptual} for bimodal contrastive learning.  Following recent studies~\citep{li2021align,dou2022coarse,qiu2023not}, our model incorporates the use of ResNet-50 and a transformer as image and text encoders respectively, initialized with weights from unimodal pretraining as done in~\cite{qiu2023not}. The training process entails a batch size of $512$ over $30$ epochs. %We utilize the Adam-W optimizer with a cosine decay learning rate schedule. 
 The values for $\tau_0$ and $\tau_{max}$ are $0.001$ and $0.05$. The widths of the transformation and projection layers of TempNet are set to $d_1=d_2=256$, which means that the TempNet only has 0.13M parameters. The TempNet is learned in the optimization framework of SogCLR~\cite{yuan2022provable}. 
 The parameter $\rho$ and the learning rate of the TempNet are tuned within the ranges of $\{7.0, 8.0, 9.0, 10.0\}$ and $\{$1e-4, 5e-5, 1e-5, 5e-6$\}$, respectively. As for tuning these two parameters, we simply ensure that the final averaged temperature value in a range of $[0.01,0.02]$, which is commonly used by previous methods~\citep{radford2019language,li2021align}.

 %Model evaluation is conducted on two downstream tasks: cross-modal retrieval and image classification in the zero-shot setting, adhering to established evaluation protocols~\citep{radford2021learning,goel2022cyclip}.

%{\noindent\bf Data.} We use the Conceptual Captions 3M (CC3M)~\citep{sharma2018conceptual} and CC12M~\citep{changpinyo2021conceptual} datasets, which are extensively employed in the domain of bimodal representation learning~\citep{li2021supervision,mu2022slip,goel2022cyclip}. During evaluation, we employ two well-established bimodal datasets: Flickr30K~\citep{plummer2015flickr30k} and MSCOCO~\citep{lin2014microsoft}, and three standard image datasets namely CIFAR10, CIFAR100, and ImageNet1K.

{\noindent\bf Evaluation \& Results.} Evaluation is conducted on two downstream tasks: cross-modal retrieval and zero-shot image classification, adhering to established evaluation protocols~\citep{radford2021learning,goel2022cyclip}. For retrieval, we use two datasets Flickr30K~\citep{plummer2015flickr30k} and MSCOCO~\citep{lin2014microsoft}, and for zero-shot classification we use three standard image datasets namely CIFAR10, CIFAR100, and ImageNet1K. We compare with several baselines, including CLIP, CyCLIP~\citep{goel2022cyclip}, SogCLR~\citep{yuan2022provable} and iSogCLR~\citep{qiu2023not}, which employ different strategies for adjusting the temperature. CLIP and CyCLIP use a heuristic apporach to optimize a temperature parameter on the fly, SogCLR uses a tuned global temperature $\tau=0.01$, and iSogCLR optimizes individualized temperature variables on the fly. 

We present partial results in Table~\ref{tab:bimodal_results_main} and full results in Tables~\ref{tab:bimodal_zs_retrieval_full_flickr},~\ref{tab:bimodal_zs_retrieval_full_coco}, and~\ref{tab:bimodal_zs_classification_full} in Appendix~\ref{subsec:app:additional_results}. Compared with baselines,
our algorithm achieves significant improvements on both downstream tasks. Specifically, the improvement of TempNet over iSogCLR could be as high as  $2.3\%$. Note that TempNet and iSogCLR only differ on how the temperatures are predicted. Hence, the improvement over iSogCLR directly proves the effectiveness of using TempNet. 

%In the subsequent section, we will demonstrate through experiments that the improvements stem from TempNet's superior generalization capabilities, particularly when the data contains noise.

\begin{figure}[t]
\vspace{-0.4cm}
\begin{minipage}[c]{0.235\textwidth}
\centering\includegraphics[width=1\textwidth]{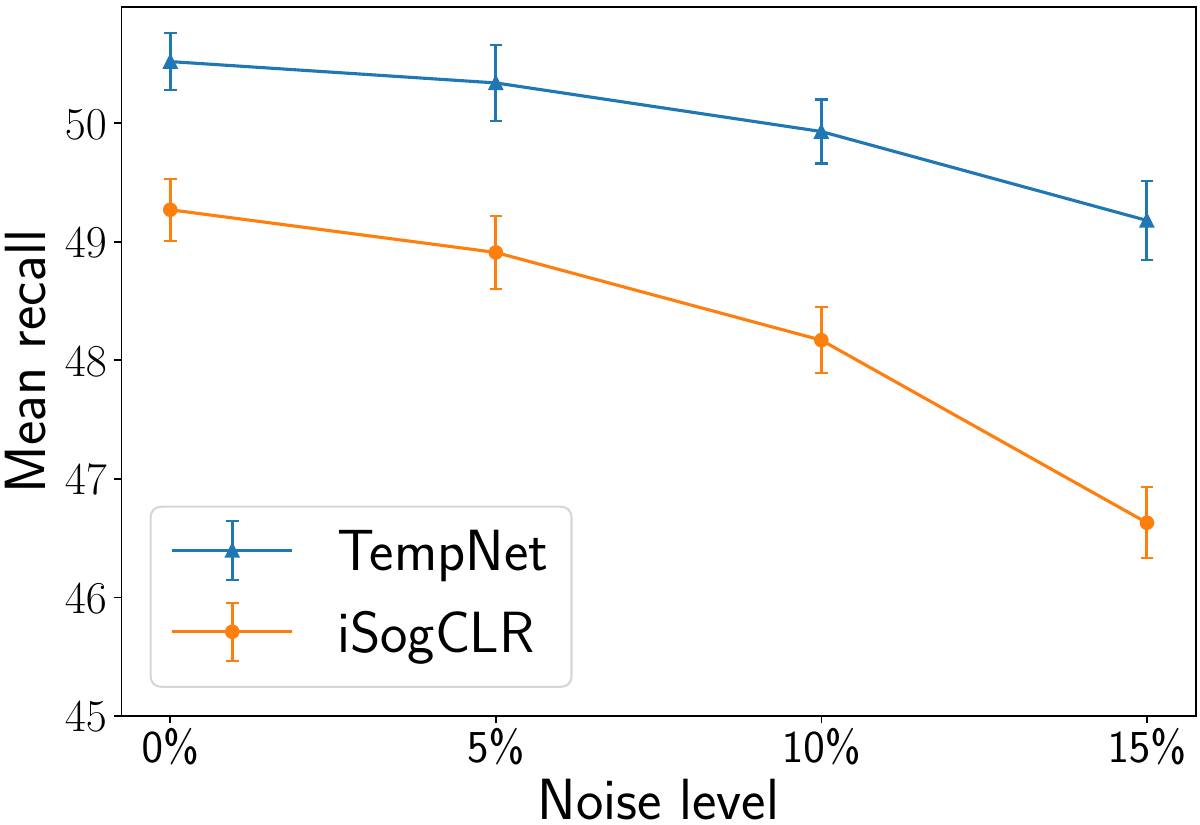}
\end{minipage}
\begin{minipage}[c]{0.235\textwidth}
\centering\includegraphics[width=1\textwidth]{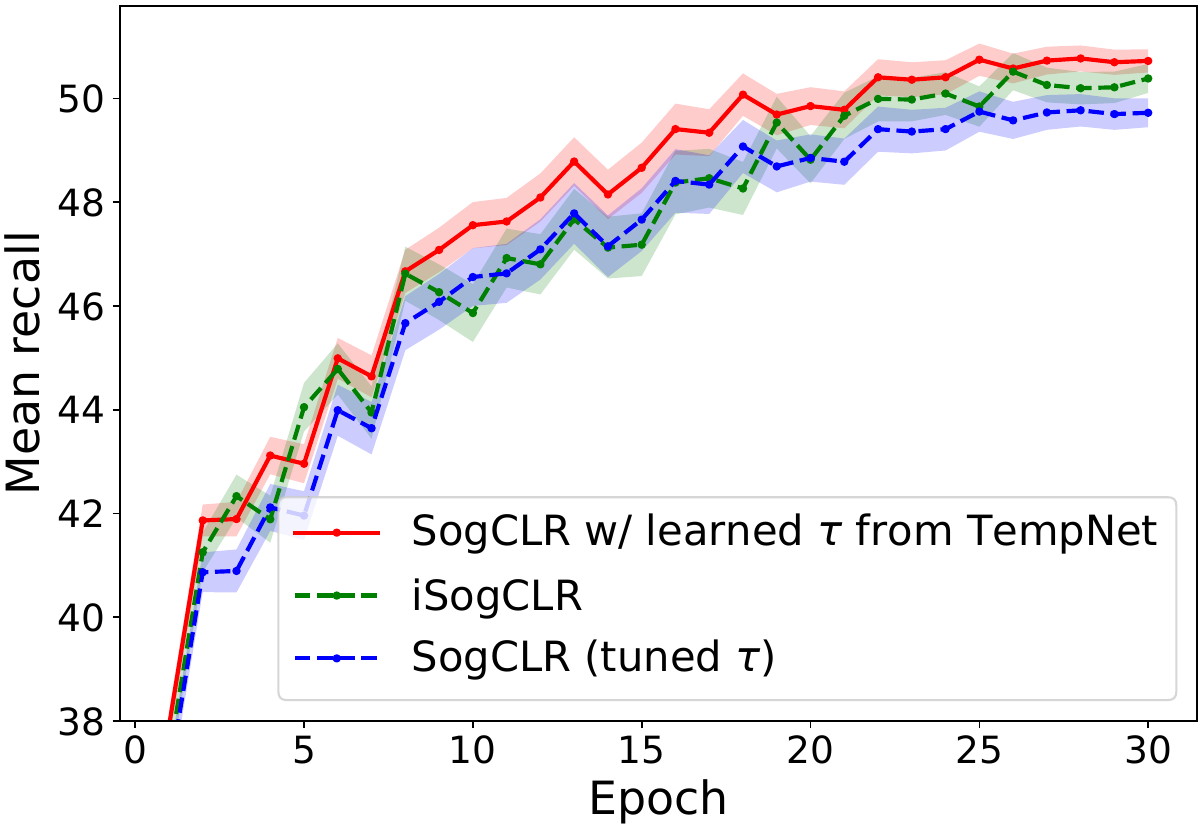}
\end{minipage}
\vspace{-0.3cm}
\caption{Left: the robustness to noise in captions of iSogCLR and TempNet. Right: the tranferability of TempNet. (cf. the texts).}
\label{fig:generalization_tempnet}
%\vspace{-0.2in}
\end{figure}

\subsection{Generalization Abilities of TempNet}
We further demonstrate the generalization abilities of TempNet for learning LLMs and contrastive learning by comparing with the individual temperature optimization under the same robust learning framework. \\{\noindent\bf LLMs with TempNet vs Individually Optimized $\tau$}.
%In this section, we study the generalization performance of TempNet across two domain-specific tasks. First, in language model training, we a novel baseline method.
As mentioned at the end of section 4.1, the individual temperature in the robust softmax loss can be directly optimized by using the Newton method. 
%Noting that all logits are accessible in the softmax loss, we propose to use Newton method to optimize the optimal temperature parameters at each iteration based on the input logits. 
%The details iteration process can be referenced in Appendix~\ref{sec:app:newton}. 
In Table~\ref{tab:gpt2_ce_with_optimized_tau_part}, we refer to the method of optimizing personalized temperatures in this manner as Baseline2, and we compare it with our approach using TempNet.
%We test this baseline for training GPT-2 from scratch and compare it with our method of training GPT-2 with TempNet as described in previous subsection. The results are shown in Table~\ref{tab:gpt2_ce_with_optimized_tau_part}. 
One can observe that employing TempNet yields better results compared to optimizing individualized temperatures, demonstrating TempNet has a better generalization capability. Additionally, in Table~\ref{tab:gpt2_ce_with_optimized_tau_part}, we further verify the impact of DRO-based robust loss. Specifically, we train TempNet along with a standard temperature-scaled cross-entropy loss referred to as Baseline3. %By comparing the results of Baseline3 with Ours, we observe that DRO-based loss also plays a crucial role.
We can observe that solely using DRO loss or TempNet does not improve the standard training method (Baseline1) much. It is the combination of the DRO loss and TempNet that improves the the standard training method by a noticable margin. 

\iffalse
{\noindent\bf Transferability of TempNet.} %In contrastive learning tasks, to validate the generalizability of TempNet, we conduct extensive experiments. 
We conduct an experiment in a transfer learning setting to verify the generalizability of TempNet for contrastive learning.  First, we fix a pretrained CLIP model and only learn a TempNet on the CC12M dataset~\citep{changpinyo2021conceptual}. The TempNet is trained for 10 epochs with an initial learning rate of 1e-4. Subsequently, the learned TempNet is employed to predict the temperatures of all samples in the CC3M dataset. These temperatures are then used in the SogCLR algorithm for model training on CC3M. We compare the resulting model against those trained by iSogCLR and SogCLR (with a tuned temperature 0.01) on the same dataset, and report the mean recall results on MSCOCO data in Fig.~\ref{fig:generalization_tempnet} (left). The mean recall is computed by averaging the recalls of image retrieval and text retrieval at top 1, 5, 10 positions. As seen, using the temperatures predicted by a TempNet learned on CC12M exhibited superior performance on the CC3M dataset, indicating generalization capabilities of TempNet.
\fi

{\noindent\bf Increased Robustness to Noisy Captions for CLIP training.} To further demonstrate the robustness of TempNet, we consider training a CLIP model in a setting where the text caption of images are noisy, %.  Wrong captions of images are common on the internet.  
which is very common in real world. To this end, we use the CC3M dataset and manually add noise to the captions. In particular, we randomly choose $p\%$ of captions and randomly delete one word or substitute one word with `random'. We vary the noise level $p\%$ among $\{5\%, 10\%, 15\%\}$. 
%by  s added with varying levels of noise (such as random deletion or substitution of words in the text), 
We report the mean recall on MSCOCO data of models learned by our method and iSogCLR  under different noise levels in Fig.~\ref{fig:generalization_tempnet} (left). We can see that TempNet exhibits greater robustness to noise, hence indicating a more pronounced generalizability. Additionally, we visualize the distributions of temperature parameters generated by both algorithms under different noise levels in Fig.~\ref{fig:impact_of_noise_to_tau_dists} in Appendix~\ref{subsec:app:additional_results}. TempNet has an effect of correct the wrong temperatures of noisy images with frequent semantics predicted by iSogCLR.   %One can see that TempNet is more robust to noise compared to iSogCLR.

Lastly, we show a {transferability} result of TempNet in CL. A TempNet is learned with a fixed CLIP model on CC12M data~\citep{changpinyo2021conceptual}. Then we use it to predict temperatures on CC3M data for contrastive learning on CC3M using SogCLR. %It gives better performance than iSogCLR on CC3M data and SogCLR with a tuned global temperature as  shown in 
The red curve (mean recall on MSCOCO) in Fig.~\ref{fig:generalization_tempnet} (right) demonstrates the effectiveness of this approach as compared with iSogCLR and SogCLR with a tuned temperature on CC3M data.  Details are in Appendix~\ref{subsec:app:additional_results}. 

\begin{figure}[t]
\vspace{-0.3cm}
\begin{minipage}[c]{0.235\textwidth}
\centering\includegraphics[width=1\textwidth]{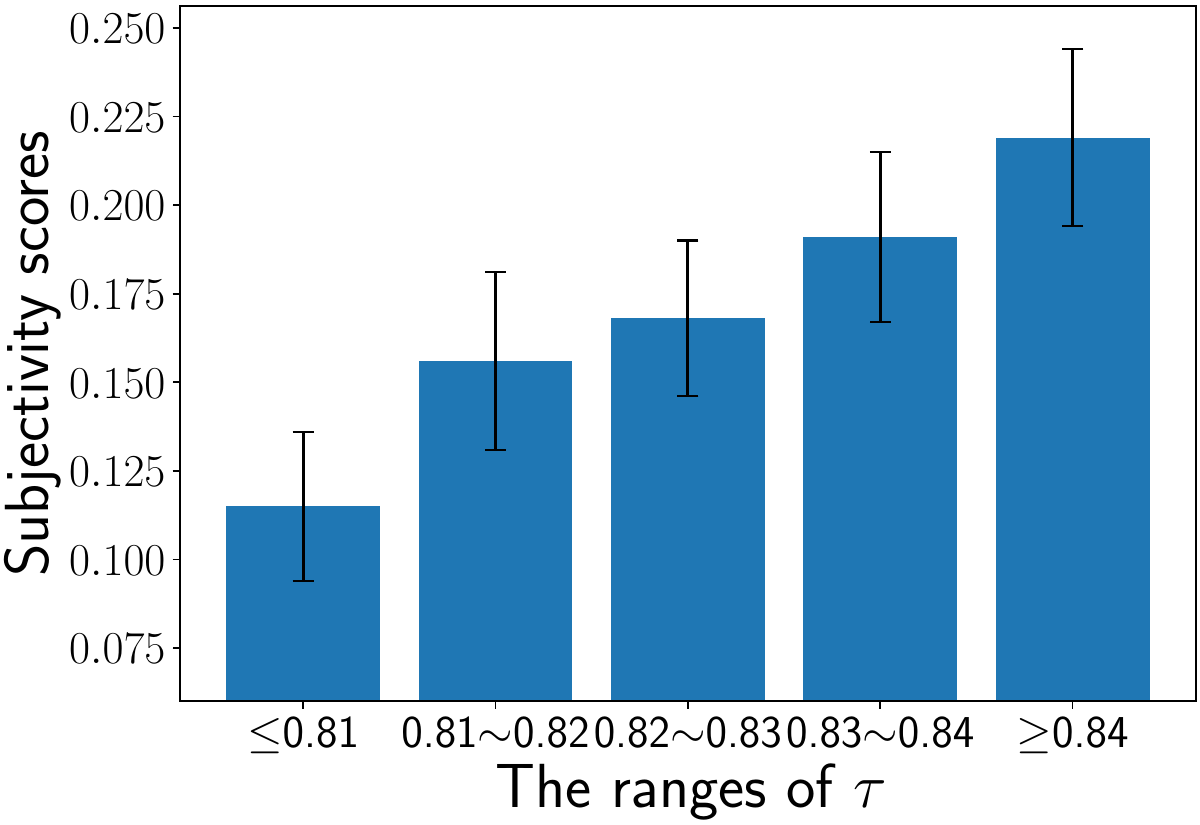}
\end{minipage}
\begin{minipage}[c]{0.235\textwidth}
\centering\includegraphics[width=1\textwidth]{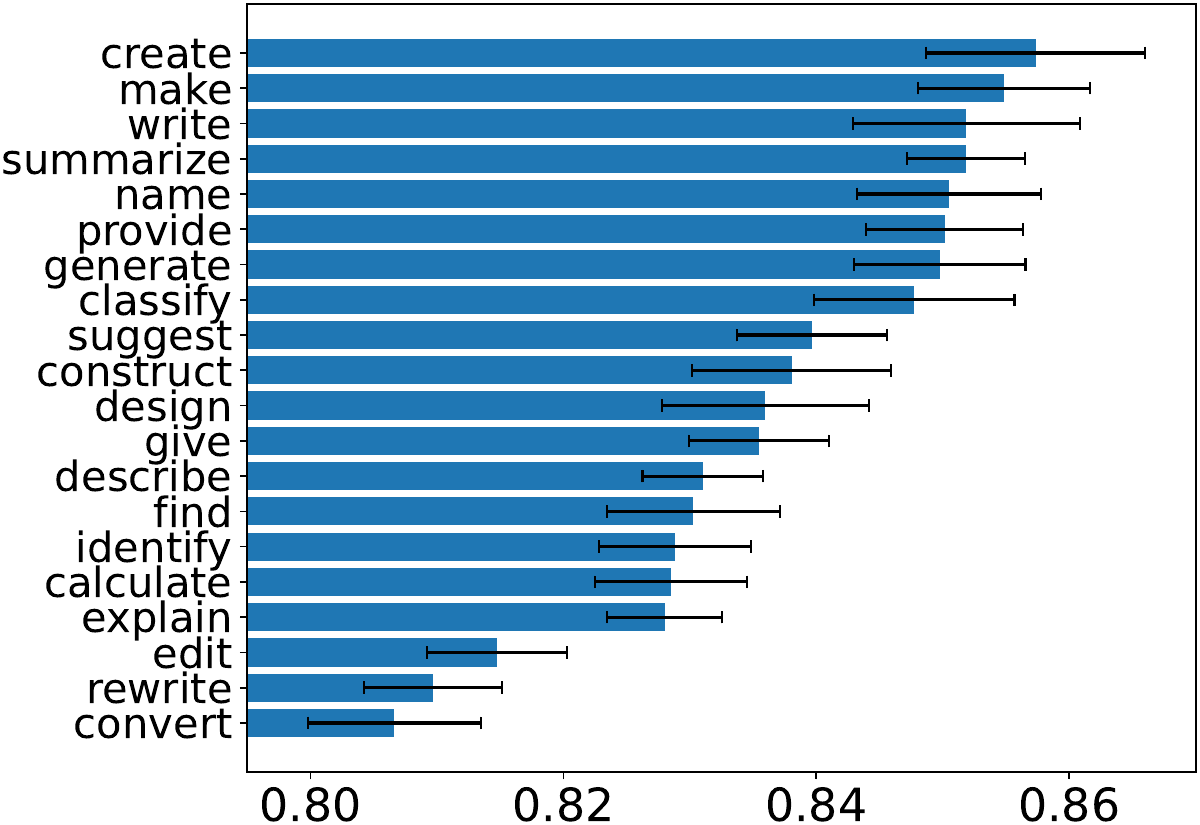}
\end{minipage}
\vspace{-0.3cm}
\caption{The relationship between input text characteristics (left: subjectivity, right: instruction type) and its temperature values.}
\label{fig:text_characteristics_and_tau}
%\vskip -0.15in\
%\vspace{0.05in}
\end{figure}

%\begin{table}[t]
%\vspace{-5mm}
%\caption{Evaluations on AlpacaEval. TT stands for training time.} %The training is conducted on $4$ A100-80GB GPUs.}
%\vspace{-2mm}\begin{center}
%\begin{small}
%\begin{tabular}{l|c|l}
%\toprule
%Model & TT (h) & Win rate (\%) \\
%\midrule
%LLaMA2 Chat 7B (leaderboard) & $\textemdash$  & 4.96\\
%LLaMA2 Chat 13B (leaderboard)& $\textemdash$ & 7.70\\
%LLaMA2 Chat 70B (leaderboard)& $\textemdash$ & 13.87\\
%GPT4 (06/13, leaderboard) & $\textemdash$ & 15.76\\
%Gemini Pro (leaderboard) & $\textemdash$ & 16.85\\
%\midrule
%LLaMA2 Chat 7B w/ $\tau=0.7$ & $\textemdash$ &4.78$\pm$0.13 \\
%LLaMA2 Chat 7B w/ TempNet & 1.42 &\textbf{5.91}$\pm$0.31 \\
%\midrule
%LLaMA2 Chat 13B w/ $\tau=0.7$ & $\textemdash$ &7.58$\pm$0.23  \\
%LLaMA2 Chat 13B w/ TempNet & 3.68 &\textbf{8.11}$\pm$0.26 \\
%\midrule
%LLaMA2 Chat 70B w/ $\tau=0.7$ & $\textemdash$ &14.13$\pm$0.37  \\
%LLaMA2 Chat 70B w/ TempNet & 11.4 &\textbf{15.81}$\pm$0.56\\
%\midrule
%Yi Chat 34B w/ $\tau=0.3$ & $\textemdash$ & 28.84$\pm$0.63 \\
%Yi Chat 34B w/ TempNet &21.0 &\textbf{29.25}$\pm$0.75\\
%\bottomrule
%\end{tabular}
%\end{small}
%\end{center}
%\label{tab:alpaca_eval}
%\end{table}

\subsection{Evaluation on Instruction Following}\label{alpaca}
\vspace*{-0.01in}
% For training TempNet on the LLaMA-chat models~\citep{touvron2023llama2} for text generation, we employed a dataset of 52K instruction-following data used for training Alpaca~\citep{alpaca}. Additionally, we employ the SubjQA~\citep{bjerva2020subjqa} dataset in our in-depth analyses.
 
%The experiments above reveal that different types of context yield distinct temperature parameters, enabling the use of TempNet to potentially enhance text generation quality. 
%In this subsection, 
We evaluate the instruction following ability of TempNet-enabled LLMs by using the evaluator of AlpacaEval 2.0~\cite{alpaca_eval}.  %Therefore, we conduct a series of experiments to validate the efficacy of TempNet in text generation.
In order to compare models on the leaderboard of AlpacaEval, we use the LLaMA2 Chat models, including the 7B, 13B, and 70B variants, which are fine-tuned versions of LLaMA2 tailored for dialogue scenarios. We train TempNets on OpenWebText2 data with fixed LLaMA2 Chat models using hyperparameter settings similar to that in Section~\ref{sec:exp-llm}. Specifically, $\tau_0=0.001, \tau_{\max}=2.0, \rho=10$. However, as noted before the resulting TempNet gives an average temperature values around 0.81, which is larger than the default value of $0.7$ for instruction following. To correct this, we use a simple trick to rescale the temperature values during the inference time by scaling $\tau_{\max}$ such that the averaged predicted temperature value is around 0.7. We refer the readers to Figure~\ref{fig:tau_max_eval} (right) for  the impact of $\tau_{\max}$ at the inference time.       %Given an instruction, we employ TempNet-enabled LLaMA2 models to generate texts. 
The results are reported %of TempNet-enabled LLaMA2 models and other models from the leaderboard 
in Table~\ref{tab:alpaca_eval}. %, where we also evaluate the base models with default $\tau=0.7$ as in~\cite{alpaca_eval}. %We use the same method to experiment with the Yi 34B Chat~\cite{young2024yi} model, where we evaluate the base model with default $\tau=0.3$ as in~\cite{alpaca_eval}.
One can see that TempNet  achieves significant improvements in generative tasks. We have also conducted more detailed case studies to analyze where the improvements come from in Appendix~\ref{sec:app:alpaca_eval_analysis}. The results show that TempNet can output a higher temperature to enhance the flexibility of answers for open-ended questions, and a lower temperature to improve the accuracy of answers for factual questions.
%We can see that xxx. We also %pick 
%show some examples to demonstrate the difference of generated texts from the original LLaMA2 chat model and the TempNet-enabled LLaMA2 chat model in Appendix~\ref{sec:app:alpaca_eval_analysis}.  %The quality of our generated text was assessed using the widely used AlpacaEval~\citep{alpaca_eval} evaluator.

\iffalse
\subsection{Comparing TempNet with Baseline: A Case Study} Here, we reveal why TempNet enhances performance by comparing the performances of LLaMA2 7B Chat (with the default $\tau=0.7$) and LLaMA2 7B Chat + TempNet on the AlpacaEval dataset. In Figure~\ref{fig:example_alpaca}, We select a representative example for which the AlpacaEval annotator, GPT-4, deems the response from LLaMA2 + TempNet to be not only superior to that of LLaMA but also better than the baseline response generated by GPT-4. It's clear that the model using the default temperature generated a relatively ordinary answer, while TempNet produced a creative response. Further analysis on this instruction is provided in the Appendix, showcasing results from different global $\tau$ values and the personalized $\tau$ generated for each token by our method. We conclude that the personalized temperatures produced by TempNet are what enhance the generation quality.
\fi

\subsection{Characteristics of Predicted Temperatures}\vspace*{-0.01in}

Here, we analyze the characteristics of predicted temperatures by TempNet of LLMs for language data. %the relationship between input texts and their corresponding temperature parameters predicted by TempNet. 
To this end, we learn a TempNet with a fixed LLaMA1-7B model on OpenWebText2 following the setup described in section 6.1.

{\noindent\bf Texts of High Subjectivity have High Temperatures.} We first conduct an experiment on the SubjQA dataset~\citep{bjerva2020subjqa}. It contains about 10k questions and answers, and they are assigned a subjectivity score by annotators. %We employ the LLaMA model with the corresponding TempNet to calculate the temperature of each token within a given question-answer sequence. 
We compute the mean values of predicted temperatures by TempNet on a text prompt consisting of a question and the answer, and use it to represent the overall temperature of each question-answer pair. In Fig.~\ref{fig:text_characteristics_and_tau} (left), we categorize all question-answer pairs into five distinct groups based on their temperature values and calculate the average subjectivity score for each group. The results indicate a high positive correlation between subjectivity scores and predicted temperature values of question-answer pairs. %that question-answer pairs with greater subjectivity have higher predicted temperatures, whereas those with low  subjectivity have lower predicted temperatures. %This is consistent with that high temperatures promote more diverse texts. 

{\noindent\bf Creative instructions have High Temperatures.} We further conduct experiments on Alpaca 52k instruction following datasets~\citep{alpaca}. We calculate the average temperature values for each instruction sample using a similar method to our previous experiment. Following the procedure outlined by~\citet{selfinstruct}, we categorize all instruction samples based their root verbs into 20 major categories. Subsequently, we compute the mean value of the temperatures for each category, and present the results in Fig.~\ref{fig:text_characteristics_and_tau} (right). One can observe that the top five instruction types with the highest temperatures are relatively \textbf{subjective}, including \emph{create}, \emph{make}, \emph{write}, \emph{summarize}, and \emph{name}. Conversely, the five instruction types with the lowest temperatures correspond to more \textbf{objective} tasks, such as \emph{calculate}, \emph{explain}, \emph{edit}, \emph{rewrite}, and \emph{convert}.

\subsection{More Studies}

Due to the limited space, more studies are provided in Appendix~\ref{sec:app:experiments}. Specifically, we display the training curves for different methods on GPT-2 and CLIP models in Appendix~\ref{subsec:app:additional_results}, along with the performance of our method on other generative tasks such as GSM8K~\cite{cobbe2021gsm8k} and MT-bench~\cite{zheng2024judging}, and an analysis of time and space complexity for our method. In Appendix~\ref{subsec:app:hyper_param_analysis}, we present an analysis of hyperparameters of our method, including $\rho$ and $\tau_{\text{max}}$. In Appendix~\ref{subsec:app:more_ablation_studies}, we conduct numerous ablation studies to investigate the role of TempNet and DRO loss in our proposed framework, the impact of various components within TempNet, the effect of TempNet's size, the influence of training data. Finally, in Appendix~\ref{sec:app:alpaca_eval_analysis}, we offer several specific case studies on AlpacaEval to gain a deeper understanding of why TempNet enhances the generative performance of LLMs. %Ablation studies about the  %the transformation layer, projection layer, and parameterized pooling layer in  TempNet design are in Appendix~\ref{subsec:app:more_ablation_studies}. % due to limit of space. 

\section{Conclusion}

In this paper, we have proposed a principled solution to learn a properly designed temperature prediction network (TempNet) in a robust learning framework for improving large foundation models. We have conducted extensive experiments and ablations studies to demonstrate the effectiveness of our solution and our design of TempNet for large language models and contrastive learning. As a future work, we will consider adopting our framework and TempNet to train large language models on larger datasets.  %We have designed  By integrating a transformation/projection block and a parameterized inductive-biased pooling layer, TempNet adeptly generate individualized temperature for each input. By optimizing a robust loss based on DRO, TempNet enhances the performance of existing methods/models across a variety of tasks.

\section*{Impact Statements}
This paper presents work whose goal is to advance the field
of Machine Learning. There are many potential societal
consequences of our work, none of which we feel must be
specifically highlighted here.

%\section*{Acknowledgements}

\bibliography{tempnet}
\bibliographystyle{icml2024}

%%%%%%%%%%%%%%%%%%%%%%%%%%%%%%%%%%%%%%%%%%%%%%%%%%%%%%%%%%%%%%%%%%%%%%%%%%%%%%%
%%%%%%%%%%%%%%%%%%%%%%%%%%%%%%%%%%%%%%%%%%%%%%%%%%%%%%%%%%%%%%%%%%%%%%%%%%%%%%%
% APPENDIX
%%%%%%%%%%%%%%%%%%%%%%%%%%%%%%%%%%%%%%%%%%%%%%%%%%%%%%%%%%%%%%%%%%%%%%%%%%%%%%%
%%%%%%%%%%%%%%%%%%%%%%%%%%%%%%%%%%%%%%%%%%%%%%%%%%%%%%%%%%%%%%%%%%%%%%%%%%%%%%%
\newpage
\appendix
\onecolumn

\section{The Derivation of the Equivalent Form of the DRO-based Robust Loss}
\label{sec:app:dro_robust}
Let us start form the DRO-based robust loss
\begin{align*}
\ell_{\text{DRO}}(z)&= \max_{\p\in\Delta} 
\sum_{k\in\mathcal C}p_k (L_{k}(z) - L_{+}(z))   - \tau_0 \text{KL}\left(\mathbf p, \frac{1}{|\mathcal C|}\right)\\
s.t. &\quad \text{KL}(\mathbf p, 1/|\mathcal C|)\leq \rho.
\end{align*}

We follow~\citet{qi2023stochastic} and adopt the Lagrangian duality theory to convert $\ell_{\text{DRO}}(z)$ into its dual form. First, we have
\begin{equation*}
\max_{\mathbf{p} \in \Delta} \min_{\lambda \geq 0} \sum_{k\in\mathcal C}p_k(L_{k}(z) - L_{+}(z)) -\tau_0\text{KL}(\mathbf{p},1/|\mathcal C|)-\lambda(\text{KL}(\mathbf{p},1/|\mathcal C|)-\rho).
\end{equation*}

We apply Sion's minimax theorem~\citep{sion1958general} and obtain
\begin{equation*}
\min_{\lambda \geq 0}\max_{\mathbf{p} \in \Delta} \sum_{k\in\mathcal C}p_k(L_{k}(z) - L_{+}(z)) -\tau_0\text{KL}(\mathbf{p},1/|\mathcal C|)-\lambda(\text{KL}(\mathbf{p},1/|\mathcal C|)-\rho),
\end{equation*}
which is equivalent to
\begin{equation*}
\min_{\lambda \geq 0}\max_{\mathbf{p} \in \Delta} \sum_{k\in\mathcal C}p_k(L_{k}(z) - L_{+}(z)) -(\tau_0+\lambda)(\text{KL}(\mathbf{p},1/|\mathcal C|)-\rho)-\tau_0\rho.
\end{equation*}

Let $\tau=\tau_0+\lambda$, then we have
\begin{equation*}
\min_{\tau \geq \tau_0}\max_{\mathbf{p} \in \Delta} \sum_{k\in\mathcal C}p_k(L_{k}(z) - L_{+}(z)) -\tau(\text{KL}(\mathbf{p},1/|\mathcal C|)-\rho)-\tau_0\rho.
\end{equation*}

Then, the original problem is equivalent to the following problem:
\begin{equation*}
\min_{\Theta} \min_{\tau \geq \tau_0}\max_{\mathbf{p} \in \Delta} \sum_{k\in\mathcal C}p_k(L_{k}(z) - L_{+}(z)) -\tau(\text{KL}(\mathbf{p},1/|\mathcal C|)-\rho)-\tau_0\rho.
\end{equation*}

Next, we fix $\x=(\Theta^{\top},\tau)^{\top}$, derive the optimal solution $\p^{x}(\x)$ that depends on $\x$, and solve the inner maximization problem. To this end, we consider the following problem
\begin{equation*}
    \min_{\mathbf{p} \in \Delta} \sum_{k\in\mathcal C}-p_k(L_{k}(z) - L_{+}(z)) +\tau\text{KL}(\mathbf{p},1/|\mathcal C|),
\end{equation*}
which has the same optimal solution as our original problem. here are actually three constraints to handle, i.e., $p_k\geq 0,\forall k$, $p_k\leq 1,\forall k$ and $\sum_{k\in\mathcal C}p_k=1$. Note that the constraint $p_k\geq 0,\forall k$ is enforced by the term $p_k\log(p_k)$, otherwise the above objective will be infinity. Besides, the constraint $p_k\leq 1$ is automatically satisfied due to $\sum_{k\in\mathcal C}p_k=1$ and $p_k\geq 0,\forall k$. Hence, we only need to tackle the constraint $\sum_{k\in\mathcal C}p_k=1$. To this end, we define the following Lagrangian function:
\begin{equation*}
    L_{\x}(\p,\mu)=\sum_{k\in\mathcal C}-p_k(L_{k}(z) - L_{+}(z))  + \tau \left( \log|\mathcal C| + \sum_{i\in\mathcal C}p_i\log(p_i) \right) + \mu\left(\sum_{i\in\mathcal C}p_i - 1 \right),
\end{equation*}
where $\text{KL}(\mathbf{p},1/|\mathcal C|) =\log|\mathcal C| + \sum_{i\in\mathcal C}p_i\log(p_i)$, and $\mu$ is the Lagrangian multiplier for the constraint $\sum_{i\in\mathcal C}p_i = 1$. The optimal solution satisfy the KKT conditions:
\begin{equation*}
    -(L_{k}(z) - L_{+}(z)) + \tau(\log(p_k^{*}(\x))+1)+\mu=0\quad \text{and}\quad \sum_{i\in\mathcal C}p_i^{*}=1.
\end{equation*}

From the first equation, we can derive $p^*_k(\x)\propto \exp((L_{k}(z) - L_{+}(z))/\tau)$. Due to the second equation, we conclude that $p_k = \frac{\exp(L_{k}(z)/\tau)}{\sum_{l\in\mathcal C}\exp(L_{l}(z)/\tau)}$. Plugging this optimal $\p^{*}$ into the inner maximization problem over $\p$, we have
\begin{equation*}
    \sum_{k\in\mathcal C} p^*_k(\x) (L_{k}(z) - L_{+}(z)) - \tau \left(\log\mathcal C + \sum_{i\in\mathcal C}p_i^*\log(p_i^*)  \right) =\tau\log\left(\frac{1}{\mathcal C}\sum_{k\in\mathcal C}\exp\left(\frac{L_{k}(z) - L_{+}(z)}{\tau}\right) \right),
\end{equation*}
Therefore, we get the following equivalent problem:
\begin{equation}
    \min_{\tau\geq\tau_0} \tau\log\left(\frac{1}{\mathcal C}\sum_{k\in\mathcal C}\exp\left(\frac{L_{k}(z) - L_{+}(z)}{\tau}\right) \right) + (\tau-\tau_0)\rho,
\end{equation}
which is actually (\ref{eqn:dro}) because $\tau_0\rho$ is a constant.

\section{Newton Method for Solving the Optimal Temperature for DRO-based robust loss}
\label{sec:app:newton}

First, let us recall the objective $f_{\Theta}(z,\tau)$:
\begin{align*}
f_{\Theta}(z, \tau) = \tau \log\bigg(\frac{1}{|\mathcal C|}\sum_{k\in\mathcal C}\exp\left(\frac{L_{k}(z) - L_{+}(z)}{\tau}\right)\bigg)+ \tau \rho.
\end{align*}

Then the update rules for Newton method are
\begin{align*}
    \tau_i^0 &= 1.0\ (\text{the initial guess}), \\
     \tau_i^{n+1} &= \tau_i^{n} - \frac{\nabla_{\tau}f_{\Theta}(z, \tau_i^{n})}{\nabla_{\tau\tau}f_{\Theta}(z, \tau_i^{n})}, \ n=0,1,\cdots
\end{align*}
where $\nabla_{\tau}f_{\Theta}(z, \tau_i^{n})$ and $\nabla_{\tau\tau}f_{\Theta}(z, \tau_i^{n})$ can be computed quickly according to their definitions. Specifically, we have
\begin{equation}
    \nabla_{\tau}f_{\Theta}(z, \tau) = \log\bigg(\frac{1}{|\mathcal C|}\sum_{k\in\mathcal C}\exp\left(\frac{L_{k}(z)}{\tau}\right)\bigg) - \sum_{k\in\mathcal C}\frac{\exp\left(\frac{L_{k}(z)}{\tau}\right)}{\sum_{j\in\mathcal C}\exp\left(\frac{L_{j}(z)}{\tau}\right)}\frac{L_{k}(z)}{\tau} + \rho,
\label{eq:newton_grad}
\end{equation}
and 
\begin{align*}
    \nabla_{\tau\tau}f_{\Theta}(z, \tau) &=  \sum_{k\in\mathcal C}\frac{\exp\left(\frac{L_{k}(z)}{\tau}\right)}{\sum_{j\in\mathcal C}\exp\left(\frac{L_{j}(z)}{\tau}\right)}\left(-\frac{L_{k}(z)}{\tau^2}\right) \\
    & - \sum_{k\in\mathcal C} \left\{ \nabla_{\tau}\left( \frac{\exp\left(\frac{L_{k}(z)}{\tau}\right)}{\sum_{j\in\mathcal C}\exp\left(\frac{L_{j}(z)}{\tau}\right)} \right)\frac{L_{k}(z)}{\tau} + \frac{\exp\left(\frac{L_{k}(z)}{\tau}\right)}{\sum_{j\in\mathcal C}\exp\left(\frac{L_{j}(z)}{\tau}\right)}\left(-\frac{L_{k}(z)}{\tau^2}\right) \right\}\\
    &=- \sum_{k\in\mathcal C}\nabla_{\tau}\left( \frac{\exp\left(\frac{L_{k}(z)}{\tau}\right)}{\sum_{j\in\mathcal C}\exp\left(\frac{L_{j}(z)}{\tau}\right)} \right)\frac{L_{k}(z)}{\tau} \\
    &= - \sum_{k\in\mathcal C}  \frac{\exp\left(\frac{L_{k}(z)}{\tau}\right)}{\sum_{j\in\mathcal C}\exp\left(\frac{L_{j}(z)}{\tau}\right)}\left[\sum_{j\in\mathcal C} \frac{\exp\left(\frac{L_{j}(z)}{\tau}\right)}{\sum_{i\in\mathcal C}\exp\left(\frac{L_{i}(z)}{\tau}\right)} \frac{L_{j}(z)-L_{k}(z)}{\tau^2}\right]\frac{L_{k}(z)}{\tau} \\
    &=-\frac{1}{\tau} \sum_{k\in\mathcal C}  \frac{\exp\left(\frac{L_{k}(z)}{\tau}\right)}{\sum_{j\in\mathcal C}\exp\left(\frac{L_{j}(z)}{\tau}\right)}\left[\sum_{j\in\mathcal C} \frac{\exp\left(\frac{L_{j}(z)}{\tau}\right)}{\sum_{i\in\mathcal C}\exp\left(\frac{L_{i}(z)}{\tau}\right)} \frac{L_{j}(z)-L_{k}(z)}{\tau}\right]\frac{L_{k}(z)}{\tau}.
\end{align*}

Let us denote
\begin{equation*}
    A =\sum_{k\in\mathcal C}  \frac{\exp\left(\frac{L_{k}(z)}{\tau}\right)}{\sum_{j\in\mathcal C}\exp\left(\frac{L_{j}(z)}{\tau}\right)}\frac{L_{k}(z)}{\tau}.
\end{equation*}

Thus, we obtain
\begin{align}
\begin{split}
       \nabla_{\tau\tau}f_{\Theta}(z, \tau) &= -\frac{1}{\tau} \sum_{k\in\mathcal C}  \frac{\exp\left(\frac{L_{k}(z)}{\tau}\right)}{\sum_{j\in\mathcal C}\exp\left(\frac{L_{j}(z)}{\tau}\right)}\left[A - \frac{L_{k}(z)}{\tau} \right]\frac{L_{k}(z)}{\tau} \\
    &= -\frac{1}{\tau} \left[ A\cdot\sum_{k\in\mathcal C}  \frac{\exp\left(\frac{L_{k}(z)}{\tau}\right)}{\sum_{j\in\mathcal C}\exp\left(\frac{L_{j}(z)}{\tau}\right)}\frac{L_{k}(z)}{\tau} - \sum_{k\in\mathcal C}  \frac{\exp\left(\frac{L_{k}(z)}{\tau}\right)}{\sum_{j\in\mathcal C}\exp\left(\frac{L_{j}(z)}{\tau}\right)}\left(\frac{L_{k}(z)}{\tau}\right)^2  \right] \\
    &= \frac{1}{\tau} \left[ \sum_{k\in\mathcal C}  \frac{\exp\left(\frac{L_{k}(z)}{\tau}\right)}{\sum_{j\in\mathcal C}\exp\left(\frac{L_{j}(z)}{\tau}\right)}\left(\frac{L_{k}(z)}{\tau}\right)^2 - \left(\sum_{k\in\mathcal C}  \frac{\exp\left(\frac{L_{k}(z)}{\tau}\right)}{\sum_{j\in\mathcal C}\exp\left(\frac{L_{j}(z)}{\tau}\right)}\frac{L_{k}(z)}{\tau}\right)^2  \right]. 
\end{split}
\label{eq:newton_hessian}
\end{align}

Therefore, by plugging $\tau_i^n$ in (\ref{eq:newton_grad}) and (\ref{eq:newton_hessian}), we achieve the gradient and Hessian for Newton method.

\newpage

\section{Proof of Lemmas}
\label{sec:app:lemma_proof}

We first prove Lemma~\ref{lemma:upper_bound}.

\begin{proof}
Let $\tau_i^*$ denote the minimizer of $f_{\Theta}(z_i,\tau_i)$, i.e., $\tau_i^*=\argmin_{\tau_i}f_{\Theta}(z_i,\tau_i)$, which gives us $f_{\Theta}(z_i,\tau_i^*)\leq\min_{\w^{\prime}}f_{\Theta}(z_i,\tau_{\w^{\prime}}(z_i))$. Thus, we have
\begin{equation*}
    \frac{1}{n}\sum_{i=1}^{n}\min_{\tau_i\geq\tau_0}f_{\Theta}(z_i,\tau_i)=\frac{1}{n}\sum_{i=1}^{n}f_{\Theta}(z_i,\tau_i^{*})\leq\frac{1}{n}\sum_{i=1}^{n}\min_{\w^{\prime}}f_{\Theta}(z_i,\tau_{\w^{\prime}}(z_i))=\min_{\w^{\prime}}\frac{1}{n}\sum_{i=1}^{n}f_{\Theta}(z_i,\tau_{\w^{\prime}}(z_i)).
\end{equation*}
\end{proof}

Then we aim to prove Lemma~\ref{lemma:formulation_tempnet}.

\begin{proof}
Recall that
\begin{equation*}
    f_{\Theta}(z, \tau) = \tau \log\bigg(\frac{1}{|\mathcal C|}\sum_{k\in\mathcal C}\exp\left(\frac{L_{k}(z) - L_{+}(z)}{\tau}\right)\bigg)+ \tau \rho
\end{equation*}
As $\tau_*=\argmin_{\tau} f_{\Theta}(z,\tau)$, we have $\nabla_{\tau} f_{\Theta}(z,\tau_*)=0$, i.e.,
\begin{equation*}
    \log\bigg(\frac{1}{|\mathcal C|}\sum_{k\in\mathcal C}\exp\left(\frac{L_{k}(z) - L_{+}(z)}{\tau_*}\right)\bigg) - \sum_{k\in\mathcal C}\frac{\exp\left(\frac{L_{k}(z) - L_{+}(z)}{\tau_*}\right)}{\sum_{j\in\mathcal C}\exp\left(\frac{L_{j}(z) - L_{+}(z)}{\tau_*}\right)}\frac{L_{k}(z) - L_{+}(z)}{\tau_*} + \rho = 0,
\end{equation*}
which can be simplified into 
\begin{equation*}
    \log\bigg(\frac{1}{|\mathcal C|}\sum_{k\in\mathcal C}\exp\left(\frac{L_{k}(z)}{\tau_*}\right)\bigg) - \sum_{k\in\mathcal C}\frac{\exp\left(\frac{L_{k}(z)}{\tau_*}\right)}{\sum_{j\in\mathcal C}\exp\left(\frac{L_{j}(z)}{\tau_*}\right)}\frac{L_{k}(z)}{\tau_*} + \rho = 0,
\end{equation*}
and we have
\begin{equation*}
   \tau_* \log\bigg(\frac{1}{|\mathcal C|}\sum_{k\in\mathcal C}\exp\left(\frac{L_{k}(z)}{\tau_*}\right)\bigg) - \sum_{k\in\mathcal C}\frac{\exp\left(\frac{L_{k}(z)}{\tau_*}\right)}{\sum_{j\in\mathcal C}\exp\left(\frac{L_{j}(z)}{\tau_*}\right)}L_{k}(z) + \rho\tau_* = 0.
\end{equation*}

Let $b_z=\tau_* \log\bigg(\frac{1}{|\mathcal C|}\sum_{k\in\mathcal C}\exp\left(\frac{L_{k}(z)}{\tau_*}\right)\bigg) - \E_{k\in\mathcal C}L_k(z)$, and we obtain
\begin{equation*}
   b_z + \E_{k\in\mathcal C}L_k(z) - \sum_{k\in\mathcal C}\frac{\exp\left(\frac{L_{k}(z)}{\tau_*}\right)}{\sum_{j\in\mathcal C}\exp\left(\frac{L_{j}(z)}{\tau_*}\right)}L_{k}(z) + \rho\tau_* = 0.
\end{equation*}

By rearranging the terms in the above equation, we have
\begin{equation*}
    \tau_* = \frac{1}{\rho}\left[\sum_{k\in\mathcal C}\frac{\exp\left(\frac{L_{k}(z)}{\tau_*}\right)}{\sum_{j\in\mathcal C}\exp\left(\frac{L_{j}(z)}{\tau_*}\right)}L_{k}(z) - \E_{k\in\mathcal C}L_k(z) - b_z \right].
\end{equation*}

Moreover, we can bound $b_z$ as follows. By using the properties of \emph{logsumexp} function, we have
\begin{align*}
    b_z&=\tau_* \log\bigg(\frac{1}{|\mathcal C|}\sum_{k\in\mathcal C}\exp\left(\frac{L_{k}(z)}{\tau_*}\right)\bigg) - \E_{k\in\mathcal C}L_k(z) = \tau_*\log\left(\sum_{k\in\mathcal C}\exp\left(\frac{L_{k}(z)}{\tau_*}\right)\right) - \log|\mathcal C|- \E_{k\in\mathcal C}L_k(z) \\
    &\leq \tau_*\log \left( \sum_{k\in\mathcal C} \exp\left( \frac{\max_{k\in\mathcal C} \{L_{k}(z)\}}{\tau_*}\right)\right)- \log|\mathcal C| - \E_{k\in\mathcal C}L_k(z) 
    =\max_{k\in\mathcal C} \{L_{k}(z)\}- \E_{k\in\mathcal C}L_k(z),
\end{align*}
and due to $e^x$ is convex w.r.t. $x$ and $\E[e^x]\geq e^{\E[x]}$, we have
\begin{align*}
    b_z&=\tau_* \log\bigg(\frac{1}{|\mathcal C|}\sum_{k\in\mathcal C}\exp\left(\frac{L_{k}(z)}{\tau_*}\right)\bigg) - \E_{k\in\mathcal C}L_k(z) \geq \tau_*\log\exp\left(\frac{1}{|\mathcal C|}\sum_{k\in\mathcal C}  \frac{L_{k}(z)}{\tau_*}\right) - \E_{k\in\mathcal C}L_k(z) \\
    &=\tau_*\frac{1}{|\mathcal C|}\sum_{k\in\mathcal C}  \frac{L_{k}(z)}{\tau_*}- \E_{k\in\mathcal C}L_k(z)=0.
\end{align*}

\end{proof}

\section{Experiments}
\label{sec:app:experiments}   

\subsection{Details of Implementation}
\label{subsec:app:experiments_implementation}
{\bf Initialization.} For the parameters of the transformation layer in TempNet, we employ the widely used kaiming uniform initialization~\citep{he2015delving} in neural networks. The projection layer in TempNet is designed to generate prototypical logits, hence in the context of contrastive learning tasks, we utilize the initial representations of randomly sampled images/texts to initialize the text/image TempNet's projection layer. For the TempNet associated with LLMs, the projection layer is initialized using the kaiming uniform method. The parameter $\w_3^{\prime}$ in the parameterized pooling, intended to autonomously weigh different components in the prototypical logits, is initialized as an all-ones vector. We set $\phi$ to the temperature values commonly used in specific tasks, such as 1.0 for LLMs and 0.01 for CLIP models. Additionally, we initialize $b$ to 0.

{\bf Codebase.} In the experiments of LLMs, we use multiple code frameworks. For the GPT-2 experiments, we employed the GPT-NeoX framework~\citep{gpt-neox-library}, which leverages good features  as the popular Megatron-DeepSpeed~\citep{rasley2020deepspeed} library but with substantially increased usability, e.g., model evaluation. In the fine-tuning experiments of GPT-2, we used pretrained weights from Pythia~\citep{biderman2023pythia}. For the LLaMA experiments, we utilized the Megatron framework~\citep{shoeybi2019megatron}, and for the LoRA experiments, we used the alpaca-lora framework\footnote{https://github.com/tloen/alpaca-lora}. When testing model performance, we primarily used the lm-evaluation-harness library~\citep{eval-harness}, which provides numerous common datasets and related tasks. For bimodal contrastive learning experiments, we adopt the code base from~\citep{qiu2023not}. We also adopt ResNet-50 as the image encoder and DistilBert~\citep{sanh2019distilbert} as the text encoder.

\subsection{Details of Datasets}
\label{subsec:app:data_details}

{\bf LLMs Experiments.} We mainly use OpenWebText2 dataset~\citep{gao2020pile}, which is part of EleutherAi/The Pile dataset and is an enhanced version of the original OpenWebTextCorpus covering all Reddit submissions from 2005 up until April 2020. During evaluation, we utilized two datasets Lambada and Wikitext-2 for assessing Language Modeling performance, and five question-answer datasets, PIQA, HellaSwag, ARC-e, ARC-c, OBQA, which are widely used for testing Large Language Models (LLMs)~\cite{brown2020language,touvron2023llama2,biderman2023pythia}.

{\bf Contrastive Learning Experiments.} We mainly use the Conceptual Captions 3M (CC3M) dataset~\citep{sharma2018conceptual}, which contains about 2.9 million image-caption pairs crawled from the Internet. Note that as time goes by, some images are not available. Thus the number of image-caption pairs we use in our experiments is smaller than that in the original papers. We also use the CC12M dataset~\citep{changpinyo2021conceptual}, which is larger and covers a much more diverse set of visual concepts than CC3M. Each image in MSCOCO and Flickr30K datasets has about 5 captions. MSCOCO dataset~\citep{lin2014microsoft} contains 113K images and 567K captions, and Flickr30K dataset~\citep{plummer2015flickr30k} has 32K images and 158K captions. We employ the well-known Karpathy split~\citep{karpathy2015deep} for these two datasets.

\begin{table}[t]
%\vspace{-5mm}
\caption{Hyper-parameters for each LLMs experiment group. ${}^*$We use 20k iterations to train LLaMA2 Chat 70B.}
\begin{center}
\begin{small}
\begin{tabular}{l|c|c|c|c|c}
\toprule
Hyper-parameters & Training GPT-2  & Finetuning GPT-2 & Fixing LLaMA1 & Fixing LLaMA2  & Fixing LLaMA2 Chat \\
\midrule
Warmup Steps Percentage & 0.01 & 0.01 & 0.01 & 0.01 &  0.01\\
Init Learning Rate &  6e-4  & 1e-4 & 2e-4 & 1e-4 &  1e-4 \\
Batch Size & 32 & 32 & 16  & 16 & 16 \\
Weight Decay & 0.1 & 0.1 &  0.1 & 0.01 & 0.01 \\
Training Iterations & 320k & 50k & 10k & 30k & 30k* \\
Learning Rate Decay & Cosine & Cosine  & Cosine  & Cosine & Cosine \\
Adam $\epsilon$ & 1e-8 & 1e-8  &  1e-8 & 1e-8 & 1e-8\\
Adam $\beta_1$ & 0.9 & 0.9  & 0.9 & 0.9 & 0.9\\
Adam $\beta_2$ & 0.95 &  0.95 & 0.95 & 0.999 & 0.999\\
\bottomrule
\end{tabular}
\end{small}
\end{center}
\label{tab:llms-hyper-params}
\end{table}

\begin{table}[t]
%\vspace{-5mm}
\caption{Hyper-parameters for each contrastive learning experiment group.}
\begin{center}
\begin{small}
\begin{tabular}{l|c|c|c}
\toprule
Hyper-parameters & CLIP Model Training   & Transferability of TempNet & Robustness to Noisy Captions \\
\midrule
Warmup Steps & 1000 & 1000 & 1000 \\
Init Learning Rate & 2e-4 & 2e-4 & 2e-4 \\
Batch Size &  512 & 512 & 512 \\
Weight Decay & 0.02 & 0.02 & 0.02 \\
Training Epochs & 30 & 10 & 30  \\
Learning Rate Decay &  Cosine & Cosine & Cosine \\
Adam $\epsilon$ & 1e-8 & 1e-8 & 1e-8 \\
Adam $\beta_1$ & 0.9 & 0.9 & 0.9 \\
Adam $\beta_2$ & 0.999 & 0.999 & 0.999 \\
\bottomrule
\end{tabular}
\end{small}
\end{center}
\label{tab:cl-hyper-params}
\end{table}

\subsection{Training Setup Details}
\label{subsec:app:training_details}

{\bf LLMs Experiments.} In our experiments, we employed the GPT-2 model (125M)~\citep{radford2019language} along with a series of LLaMA models, including LLaMA1 7B~\citep{touvron2023llama1}, LLaMA2 7B, 13B, and versions of LLaMA2 fine-tuned for conversational tasks, namely LLaMA2 Chat 7B, 13B, 70B~\citep{touvron2023llama2}. GPT-2 (125M) is a 12-layer BERT-based model, and has a hidden dimension of 768. The number of layers and hidden dimensions for the LLaMA 7B, 13B, 70B models are respectively 32, 40, 80 and 4096, 5120, 8192, respectively. The training hyper-parameter configurations for all experiment groups are listed in Table~\ref{tab:llms-hyper-params}. We employ GTX 3090, 4090, A6000, and A100 GPUs in our experiments, and they have 24GB, 24Gb, 48GB, 80GB memory, respectively. The experiments on LLaMA2 Chat model (results in Table~\ref{tab:alpaca_eval}) are performed on A100 GPUs.

{\bf Contrastive Learning Experiments.} In table~\ref{tab:cl-hyper-params}, we report the training hyper-parameter configurations for all three contrastive learning experiment groups: training CLIP model, verifying the transferability of TempNet, and demonstration the robustness of TempNet to noisy captions.  We train our models on Nvidia Tesla V100 GPU with 32 GB memory and GTX 3090 GPU with 24GB memory. It takes about 8 hours to finish 30 epochs training.

\newpage
\subsection{Additional Experimental Results}
\label{subsec:app:additional_results}

\begin{figure}[h]
    \centering 
    \subfigure[Training GPT-2 From Scratch]{
        \includegraphics[width=0.25\textwidth]{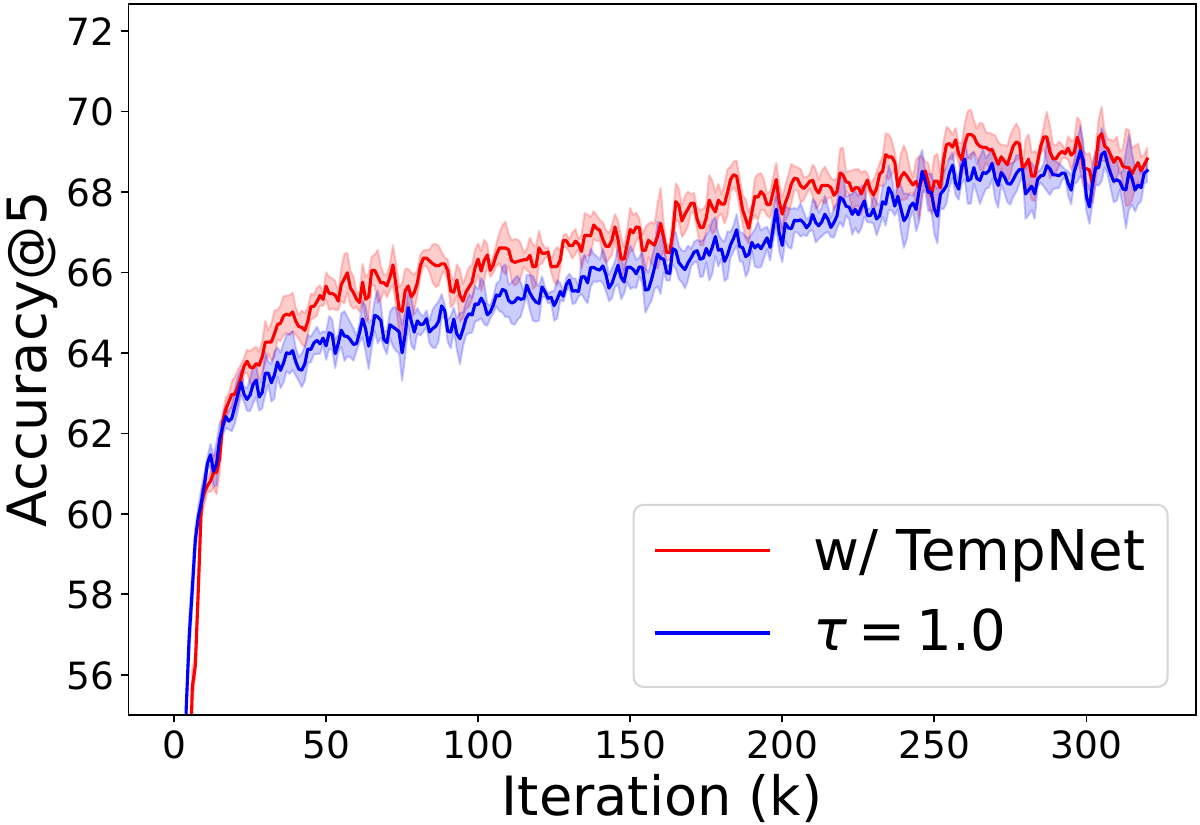}
    }
    \subfigure[Training GPT-2 From Scratch]{
        \includegraphics[width=0.25\textwidth]{gpt_curves/training_curves_ppl_320k_notitle.pdf}
    }
    \subfigure[Finetuning GPT-2]{
        \includegraphics[width=0.25\textwidth]{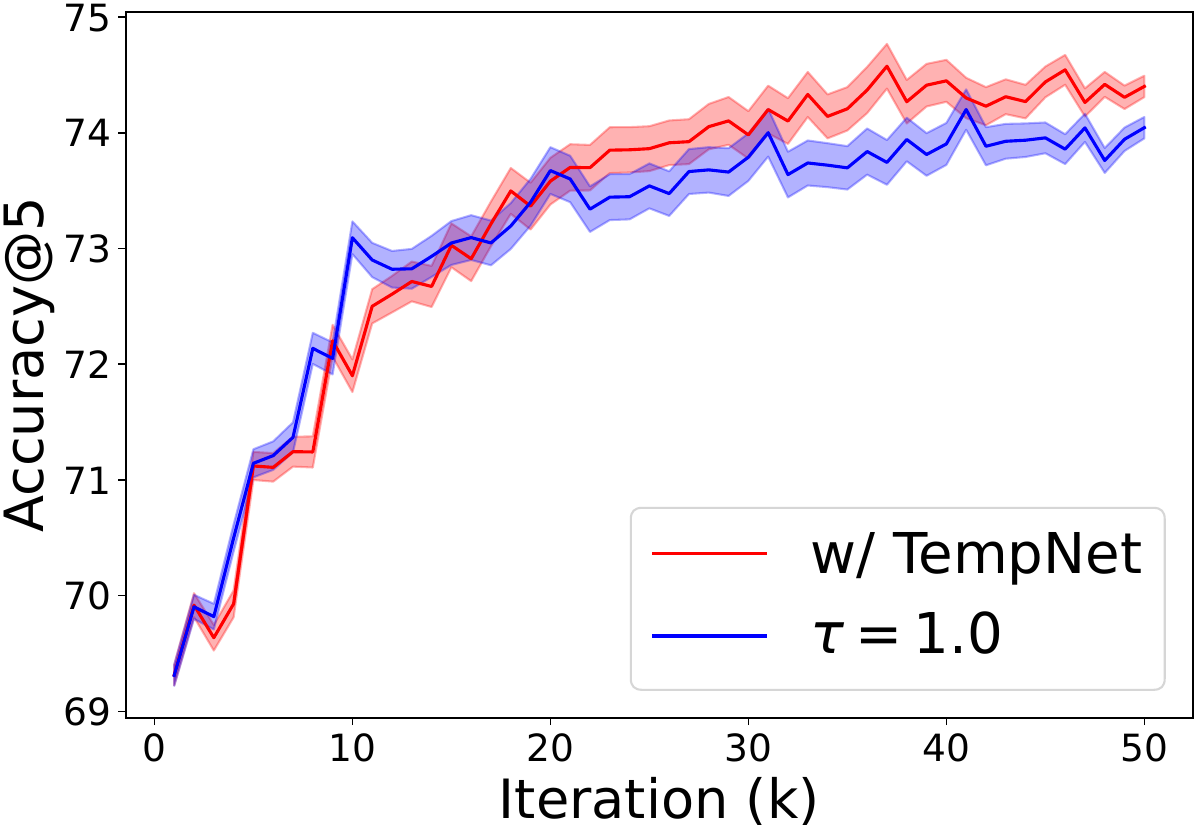}
    }
    \subfigure[Finetuning GPT-2]{
        \includegraphics[width=0.25\textwidth]{gpt_curves/training_curves_ppl_50k_notitle.pdf}
    }
       \subfigure[Training CLIP]{   \includegraphics[width=0.25\textwidth]{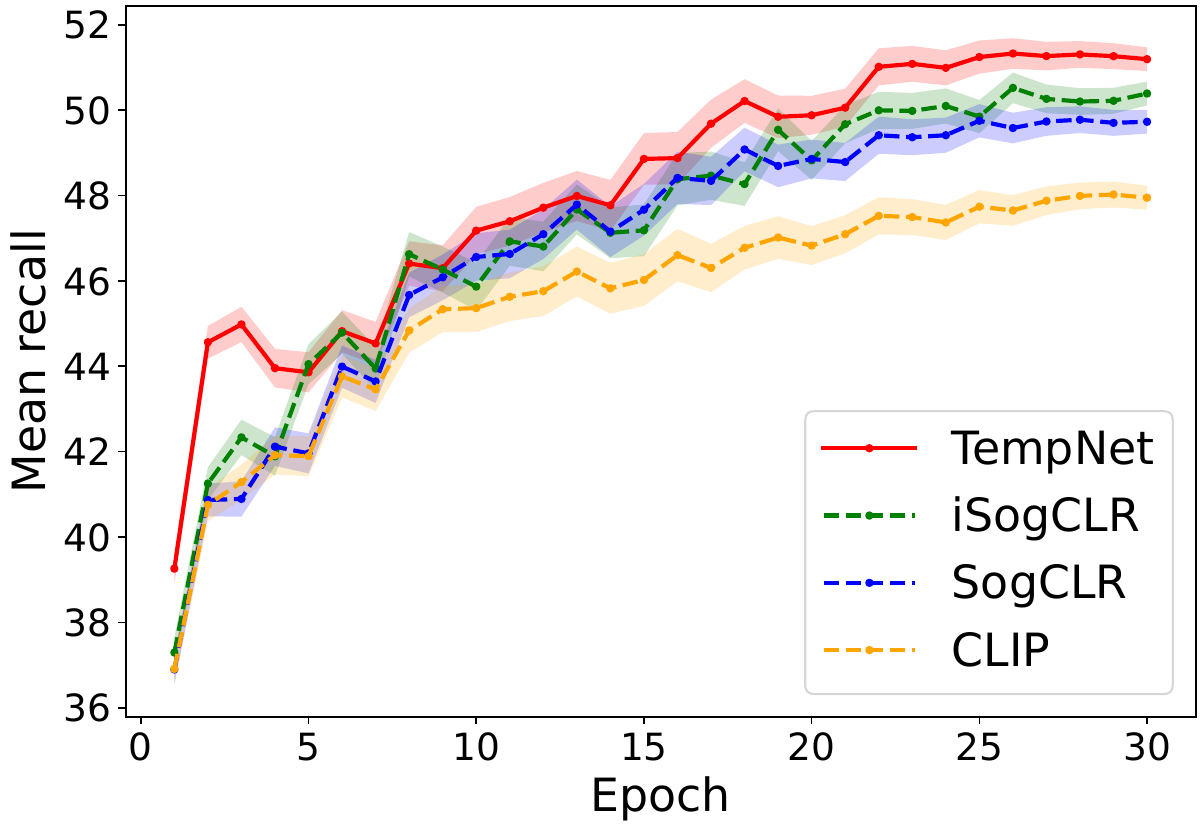}}

    \vspace{-8pt}
    \caption{The training curves for the GPT-2 model (a-d), and for the CLIP model (e).}
    \label{fig:training_curves_gpt2}
%\end{figure}
%\begin{figure}[t]
   % \centering
    %%\includegraphics[width=0.25\textwidth]{contrastive_curves/training_curves_notitle.pdf}
    %\caption{The training curves for the CLIP model.}
    %\label{fig:training_curves_clip}
\end{figure}

\begin{figure}[h]
    \centering 
    \subfigure[iSogCLR, noise level=0.0\%]{
        \includegraphics[width=0.4\textwidth]{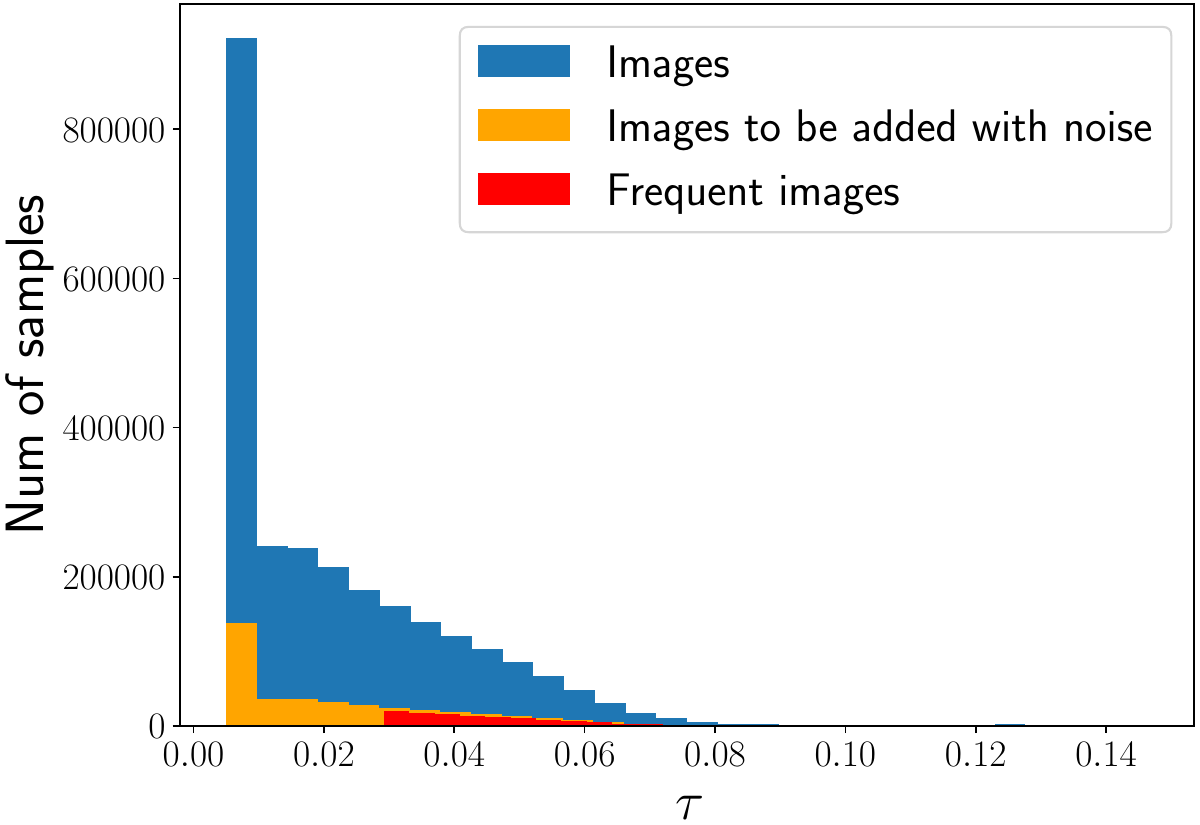}
    }
    \subfigure[iSogCLR, noise level=15\%]{
        \includegraphics[width=0.4\textwidth]{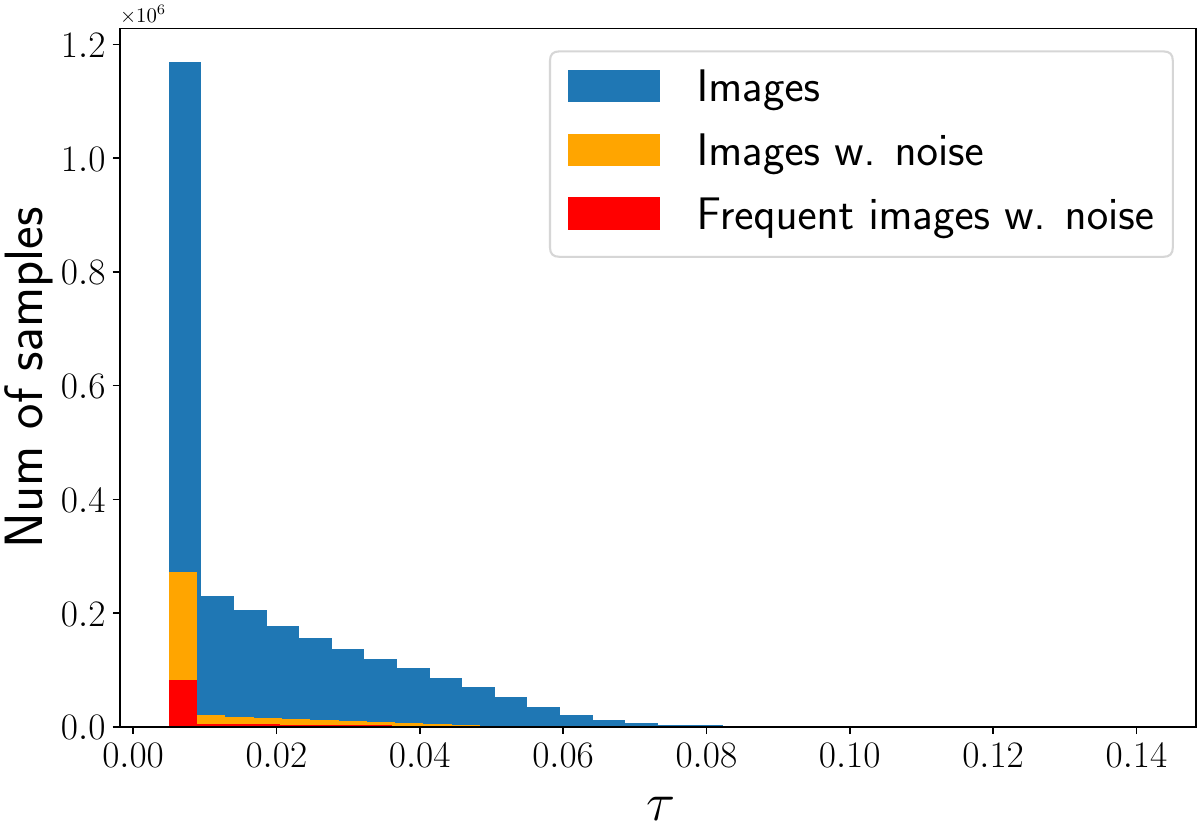}
    }
    \subfigure[TempNet, noise level=0.0\%]{
        \includegraphics[width=0.4\textwidth]{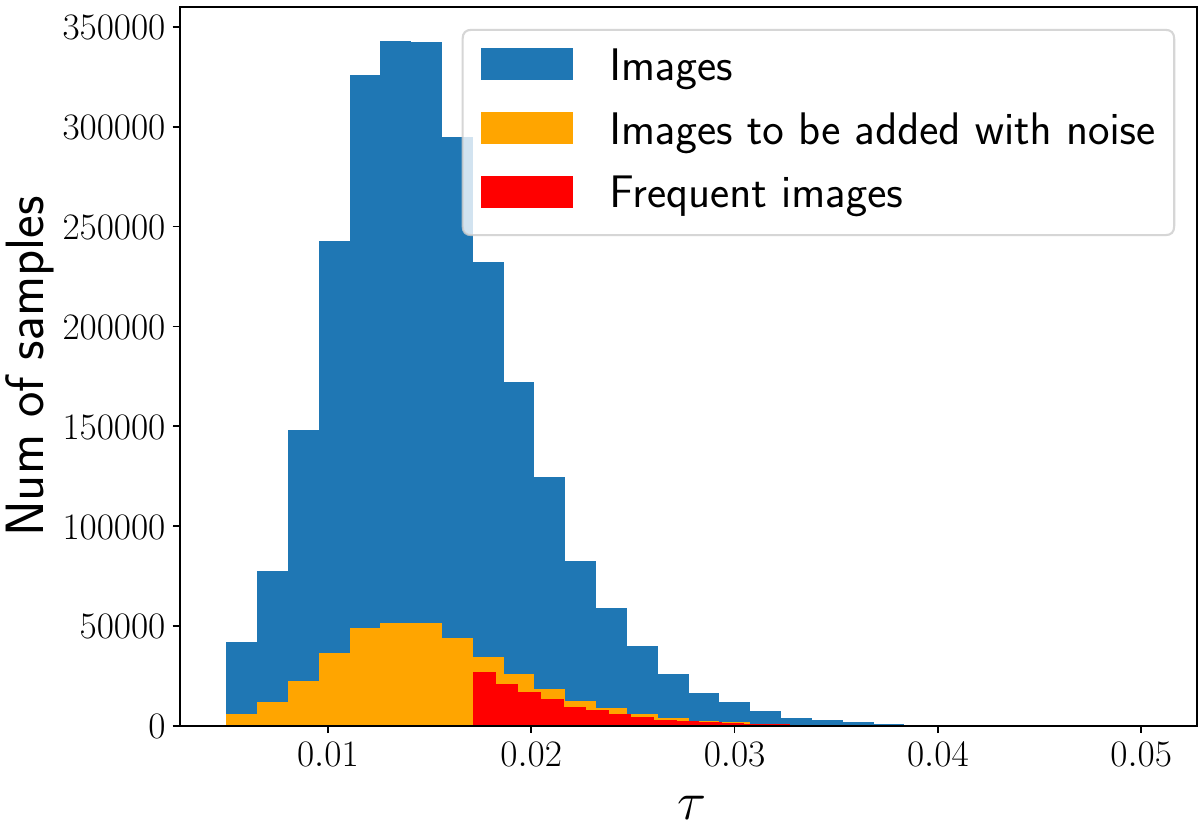}
    }
    \subfigure[TempNet, noise level=15\%]{
        \includegraphics[width=0.4\textwidth]{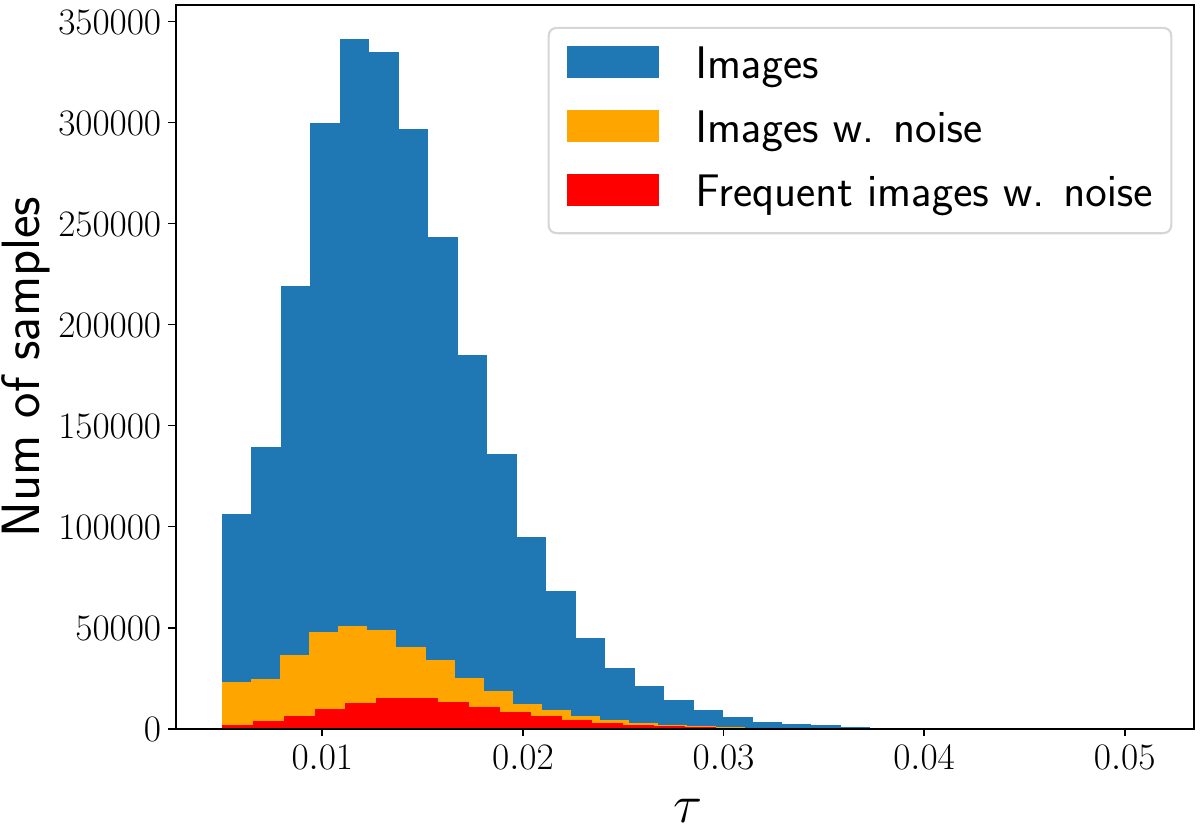}
    }
    \vspace{-8pt}
    \caption{The impact of noise on the distributions of learned temperature parameters. TempNet corrects the temperatures of noisy data with frequent semantics, which are predicted as low by iSogCLR. }
    \label{fig:impact_of_noise_to_tau_dists}
\end{figure}

\begin{table*}[t]
\vspace{-4mm}
\caption{Zero-shot image-text retrieval (text-to-image and image-to-text) results (Recall@$k$), where $k\in\{1,5,10\}$, on Flickr30K dataset.}
\vspace{-4mm}
\label{tab:bimodal_zs_retrieval_full_flickr}
\vskip 0.2in
\begin{center}
\begin{small}
\begin{sc}
\begin{tabular}{p{1.3cm}p{1.8cm}p{1.8cm}p{1.8cm}p{1.8cm}p{1.8cm}p{1.8cm}}
\toprule
\multirow{1}{*}{\thead{Method}} &
\multicolumn{3}{c}{\thead{Image retrieval}} &
\multicolumn{3}{c}{\thead{Text retrieval}} \\
\cmidrule(lr){2-4}
\cmidrule(lr){5-7}
& R@1 & R@5 & R@10 & R@1 & R@5 & R@10  \\
\midrule
CLIP & 40.98$\pm$0.22 & 69.60$\pm$0.19 & 79.22$\pm$0.08 & 50.90$\pm$0.17 & 81.00$\pm$0.16 & 87.90$\pm$0.22 \\
CyCLIP & 42.46$\pm$0.13 & 69.56$\pm$0.16 & 78.74$\pm$0.21 & 51.70$\pm$0.23 & 79.90$\pm$0.18 & 88.40$\pm$0.11  \\
SogCLR & 43.32$\pm$0.18 & 71.06$\pm$0.13 & 79.54$\pm$0.19 & 57.18$\pm$0.20 & 81.03$\pm$0.26 & 88.62$\pm$0.18 \\
iSogCLR & 44.36$\pm$0.12 & 72.64$\pm$0.17 & 80.92$\pm$0.13 & 60.20$\pm$0.26 & 84.60$\pm$0.21 & 90.50$\pm$0.14 \\
TempNet & \textbf{46.17}$\pm$0.14 & \textbf{73.68}$\pm$0.12  & \textbf{82.45}$\pm$0.15 &  \textbf{62.51}$\pm$0.19  & \textbf{85.31}$\pm$0.20  &  \textbf{92.05}$\pm$0.13\\
\bottomrule
\end{tabular}
\end{sc}
\end{small}
\end{center}
\vskip -0.0in
%\end{table*}
%\begin{table*}[t]
\vspace{-4mm}
\caption{Zero-shot image-text retrieval (text-to-image and image-to-text) results (Recall@$k$), where $k\in\{1,5,10\}$, on MSCOCO dataset.}
\vspace{-4mm}
\label{tab:bimodal_zs_retrieval_full_coco}
\vskip 0.2in
\begin{center}
\begin{small}
\begin{sc}
\begin{tabular}{p{1.3cm}p{1.8cm}p{1.8cm}p{1.8cm}p{1.8cm}p{1.8cm}p{1.8cm}}
\toprule
\multirow{1}{*}{\thead{Method}} &
\multicolumn{3}{c}{\thead{Image retrieval}} &
\multicolumn{3}{c}{\thead{Text retrieval}} \\
\cmidrule(lr){2-4}
\cmidrule(lr){5-7}
& R@1 & R@5 & R@10 & R@1 & R@5 & R@10 \\
\midrule
CLIP & 21.32$\pm$0.12 & 45.52$\pm$0.17 & 57.30$\pm$0.16 & 26.98$\pm$0.21 & 54.86$\pm$0.15 & 66.86$\pm$0.19 \\
CyCLIP & 21.58$\pm$0.19 & 45.46$\pm$0.13 & 57.56$\pm$0.22 & 26.18$\pm$0.24 & 53.24$\pm$0.18 & 65.86$\pm$0.22 \\
SogCLR & 22.43$\pm$0.13 & 46.74$\pm$0.11 & 58.32$\pm$0.20 & 30.08$\pm$0.22 & 56.94$\pm$0.17 & 67.39$\pm$0.24 \\
iSogCLR & 23.27$\pm$0.18 & 47.23$\pm$0.24 & 59.07$\pm$0.19 & 32.72$\pm$0.13 & 59.52$\pm$0.11 & 70.78$\pm$0.21 \\
TempNet & \textbf{24.83}$\pm$0.16  & \textbf{49.29}$\pm$0.21  & \textbf{61.05}$\pm$0.18  & \textbf{34.50}$\pm$0.16  &  \textbf{61.26}$\pm$0.14 & \textbf{72.14}$\pm$0.16 \\
\bottomrule
\end{tabular}
\end{sc}
\end{small}
\end{center}
\vskip -0.0in
%\end{table*}
%\begin{table*}[t]
\vspace{-4mm}
\caption{Zero-shot top-$k$ classification accuracy (\%), where $k\in\{1,3,5\}$.}
\vspace{-4mm}
\label{tab:bimodal_zs_classification_full}
\vskip 0.2in
\begin{center}
\begin{small}
\begin{sc}
\begin{tabular}{p{1.2cm}p{1.6cm}p{1.6cm}p{1.6cm}p{1.6cm}p{1.6cm}p{1.6cm}p{1.6cm}p{1.6cm}p{1.6cm}}
\toprule
\multirow{1}{*}{\thead{Method}} &
\multicolumn{3}{c}{\thead{CIFAR10}} &
\multicolumn{3}{c}{\thead{CIFAR100}}  \\
\cmidrule(lr){2-4}
\cmidrule(lr){5-7}
& top-1 & top-3 & top-5 & top-1 & top-3 & top-5  \\
\midrule
CLIP & 60.63$\pm$0.19 & 87.29$\pm$0.12 & 95.02$\pm$0.16 & 30.70$\pm$0.11 & 49.49$\pm$0.13 & 58.51$\pm$0.14 \\
CyCLIP & 57.19$\pm$0.20 & 85.02$\pm$0.14 & 93.94$\pm$0.23 & 33.11$\pm$0.14 & 52.99$\pm$0.17 & 61.01$\pm$0.22 \\
SogCLR & 61.09$\pm$0.24 & 88.12$\pm$0.19 & 94.92$\pm$0.18 & 33.26$\pm$0.12 & 52.46$\pm$0.22 & 60.71$\pm$0.15  \\
iSogCLR & 58.91$\pm$0.15 & 86.27$\pm$0.24 & 93.43$\pm$0.11 & 33.81$\pm$0.18 & 53.21$\pm$0.21 & 61.83$\pm$0.19  \\
TempNet & \textbf{61.77}$\pm$0.18  & \textbf{88.24}$\pm$0.21  & \textbf{95.19}$\pm$0.13  & \textbf{34.69}$\pm$0.17  & \textbf{54.38}$\pm$0.14  & \textbf{62.51}$\pm$0.15  \\
\bottomrule
\end{tabular}
\begin{tabular}{p{1.2cm}p{1.6cm}p{1.6cm}p{1.6cm}}
\toprule
\multirow{1}{*}{\thead{Method}} &
\multicolumn{3}{c}{\thead{ImageNet1K}} \\
\cmidrule(lr){2-4}
& top-1 & top-3 & top-5  \\
\midrule
CLIP & 36.27$\pm$0.17 & 51.03$\pm$0.17 & 56.84$\pm$0.22 \\
CyCLIP  & 36.75$\pm$0.21 & 51.32$\pm$0.18 & 57.08$\pm$0.23 \\
SogCLR  & 37.46$\pm$0.19 & 52.68$\pm$0.16 & 58.04$\pm$0.10 \\
iSogCLR  & 40.72$\pm$0.23 & 54.38$\pm$0.14 & 59.11$\pm$0.17 \\
TempNet & \textbf{42.28}$\pm$0.19  & \textbf{56.19}$\pm$0.17 & \textbf{61.32}$\pm$0.16 \\
\bottomrule
\end{tabular}
\end{sc}
\end{small}
\end{center}
\vskip -0.0in
\end{table*}

{\noindent\bf Explanation of Performance Improvement on Common Sense Reasoning Tasks.} Given a question $\x_0$ and an text of answer option $\x_1$, the lm-evaluation-harness library~\citep{eval-harness} calculates \textbf{the log likelihood} of $\log p(\x_1|\x_0)$. The answer option with the highest log likelihood is returned as the answer.

{\noindent\bf Training Curves.} We demonstrate the training curves of different methods for GPT-2 model and CLIP in Figure~\ref{fig:training_curves_gpt2}. 
It is notable that TempNet actually promotes the training of both language models and CLIP models.

\begin{table*}[h]
\caption{We demonstrate the improvements of TempNet on the LLaMA1  7B model, reporting the perplexity across three language modeling tasks and the accuracy on seven common sense reasoning tasks.}
\begin{center}
\begin{small}
\begin{tabular}
{p{4.5cm}|p{1.0cm}p{1.4cm}p{1.6cm}p{1.2cm}p{1.2cm}|cc}
\toprule
\multirow{2}{*}{\thead{Setting}} & 
\multicolumn{5}{c}{\thead{Common Sense Reasoning (acc(\%)$\uparrow$)}} &
\multicolumn{2}{c}{\thead{Language Modeling (ppl$\downarrow$)}} \\
\cmidrule(lr){2-6}
\cmidrule(lr){7-8}

 & PIQA & HellaSwag  & ARC-e & ARC-c & OBQA & Lambada & Wikitext \\
\midrule
Fixing LLaMA1 7B $\tau=1.0$  & 78.1$\pm$0.9 & 56.2$\pm$0.5  & 75.1$\pm$0.8 & 40.2$\pm$1.4 & 32.2$\pm$2.1 & 4.41$\pm$0.09  & \textbf{9.41} \\
Fixing LLaMA1 7B w/ TempNet  &  \textbf{79.1}$\pm$0.9 & \textbf{58.3}$\pm$0.4 &  \textbf{75.8}$\pm$0.8 & \textbf{42.1}$\pm$1.4 & \textbf{34.8}$\pm$2.1 & \textbf{3.65}$\pm$0.08  & 9.87 \\
\bottomrule
\end{tabular}
\end{small}
\end{center}
\label{tab:llama1_results}
\end{table*}

{\noindent\bf More Experimental Results on LLaMA1 7B.} In Table~\ref{tab:llama1_results}, we demonstrate the significant improvement achieved by TempNet on the LLaMA1 7B model. 

{\noindent\bf More Experimental Results on Bimodal CL.} We show full results of zero-shot retrieval results on MSCOCO and Flick30K, and zero-shot classification accuracy on CIFAR-10, CIFAR-100, and ImageNet1K in Table~~\ref{tab:bimodal_zs_retrieval_full_flickr}, \ref{tab:bimodal_zs_retrieval_full_coco}, and~\ref{tab:bimodal_zs_classification_full}, respectively.

\begin{table}[t]
\vspace{-5mm}
\caption{Comparison on the GSM8K dataset.}
\begin{center}
\begin{small}
\begin{tabular}{l|c|c|c}
\toprule
Model & maj@$1$   & maj@$8$& maj@$64$ \\
\midrule
LLaMA2 7B w/ $\tau=0.2$ & \textbf{13.2}±0.9 &	15.5±1.0 &	16.5±1.0 \\
LLaMA2 7B w/ TempNet & 12.8±0.9	& \textbf{15.6}±1.0 &	\textbf{17.3}±1.0 \\
\bottomrule
\end{tabular}
\end{small}
\end{center}
\label{tab:gsm8k_performance}
%\end{table}

%\begin{table}[h]
%\vspace{-5mm}
\caption{Comparison on the MT-bench dataset.}
\begin{center}
\begin{small}
\begin{tabular}{l|c|c|c}
\toprule
Model & LLaMA2 Chat 7B   & LLaMA2 Chat 13B & LLaMA2 Chat 70B \\
\midrule
w/ default $\tau$ values & 6.27 &	\textbf{6.65} &	6.89 \\
w/ TempNet & \textbf{6.44}	& 6.58 & \textbf{6.97} \\
\bottomrule
\end{tabular}
\end{small}
\end{center}
\label{tab:mtbench_performance}
%\end{table}

%\begin{table}[t]
%\vspace{-5mm}
\caption{The number of parameters in TempNet with different base models.}
\begin{center}
\begin{small}
\begin{tabular}{l|c|c|c}
\toprule
Base Model & Base Model \#Params   & TempNet \#Params & TempNet Param\% \\
\midrule
CLIP model & 92M &	0.26M & 0.28\% \\
GPT-2 & 125M &	12.94M & 10.35\% \\
LLaMA 7B & 6746M & 8.26M & 0.12\% \\
\bottomrule
\end{tabular}
\end{small}
\end{center}
\label{tab:params_tempnet}
%\end{table}

%\begin{table}[t]
%\vspace{-5mm}
\caption{Comparison of training/inference time.}\label{tt}
\begin{center}
\begin{small}
\begin{tabular}{l|c|c|c}
\toprule
Method	 & Training GPT-2: Training Time (h)  & Inference Throughput with GPT-2 (tokens/s)	 & Training CLIP model (h) \\
\midrule
Baseline & 1.21 &	9655.77 & 2.31 \\
Ours & 1.28 &	8966.07 & 2.42 \\
\bottomrule
\end{tabular}
\end{small}
\end{center}
\vspace{-3mm}
\begin{center}
\begin{small}
\begin{tabular}{l|c|c}
\toprule
Method	 & Fixing LLaMA2 7B: Training Time (h)  & Inference Throughput with LLaMA2 7B (tokens/s) \\
\midrule
Baseline & --- &	950.94  \\
Ours & 3.98 &	943.79\\
\bottomrule
\end{tabular}
\end{small}
\end{center}
\vspace{-3mm}
\begin{center}
\begin{small}
\begin{tabular}{l|c|c}
\toprule
Method	& Fixing LLaMA2 13B: Training Time (h) & Inference Throughput with LLaMA2 13B (tokens/s) \\
\midrule
Baseline &  --- & 590.53 \\
Ours &  10.33 & 587.26 \\
\bottomrule
\end{tabular}
\end{small}
\end{center}
\label{tab:time_comp}
\end{table}

{\noindent\bf Visualization of the Learned Temperature Distributions.} To further investigate why TempNet outperforms iSogCLR, we compare the distributions of $\tau$ learned by these methods at different noise levels in the Figure~\ref{fig:impact_of_noise_to_tau_dists}. We first set the noise level to 0\% (i.e., no noise added to the data) and showcase the scenarios for iSogCLR and TempNet in Figure~\ref{fig:impact_of_noise_to_tau_dists}(a) and (c), respectively. In these two figures, the blue part represents the distribution of tau for all images, the orange part shows the distribution for 15\% of the images (we selected samples with IDs in the top 15\%, to which we will add noise later), and the red part indicates the top 20\% of the orange samples with the highest temperature values, which can be considered as samples with frequent semantics. Subsequently, we set the noise level to 15\%, adding noise to the samples with IDs in the top 15\%, and display the distribution of temperature values for iSogCLR and TempNet in (b) and (d), respectively. As before, the blue section represents all samples, the orange section denotes the 15\% of samples to which noise was added, and the red section highlights those among the orange with frequent semantics. It is evident that in iSogCLR, the samples represented in red almost exclusively learn a very small temperature values, indicating that their temperatures are underestimated due to noise. In contrast, TempNet is able to correct these samples' temperature values, ensuring they were not too low. Hence, this demonstrates that TempNet is more robust to noise when learning temperature values compared to iSogCLR.

%We visualize the distributions of learned temperature values from iSogCLR and TempNet in Figure~\ref{fig:impact_of_noise_to_tau_dists}. In this experiment, we initially set the noise level to 0\%. In Figures~\ref{fig:impact_of_noise_to_tau_dists}(a) and~\ref{fig:impact_of_noise_to_tau_dists}(c), we show the distributions of tau values learned by iSogCLR and TempNet for all data (in blue), and for the top 15\% of the data (in orange), respectively. Moreover, we selected samples within the top 15\% of data whose temperature values were in the top 20\% and regarded them as frequent samples~\cite{qiu2023not}. The red portion in the figures illustrates their positions. Subsequently, we adjusted the noise level to 15\%. In Figures~\ref{fig:impact_of_noise_to_tau_dists}(b) and~\ref{fig:impact_of_noise_to_tau_dists}(d), we present the distribution of tau values learned by iSogCLR and TempNet under this condition for all data (in blue), the distribution of noise samples (in orange), and the distribution of the previously identified frequent samples (in red). It is evident that in iSogCLR, the samples represented in red almost exclusively learned a very small temperature coefficient, indicating that their temperature values were underestimated due to noise. In contrast, TempNet was able to correct these samples' temperature values, ensuring they were not too low. Hence, this demonstrates that TempNet is more robust to noise when learning temperature values compared to iSogCLR.

{\noindent\bf Transferability of TempNet in Contrastive Learning.} %In contrastive learning tasks, to validate the generalizability of TempNet, we conduct extensive experiments. 
We conduct an experiment in a transfer learning setting to verify the generalizability of TempNet for contrastive learning.  First, we fix a pretrained CLIP model and only learn a TempNet on the CC12M dataset~\citep{changpinyo2021conceptual}. The TempNet is trained for 10 epochs with an initial learning rate of 1e-4. Subsequently, the learned TempNet is employed to predict the temperatures of all samples in the CC3M dataset. These temperatures are then used in the SogCLR algorithm for model training on CC3M. We compare the resulting model against those trained by iSogCLR and SogCLR (with a tuned temperature 0.01) on the same dataset, and report the mean recall results on MSCOCO data in Fig.~\ref{fig:generalization_tempnet} (left). The mean recall is computed by averaging the recalls of image retrieval and text retrieval at top 1, 5, 10 positions. As seen, using the temperatures predicted by a TempNet learned on CC12M exhibited superior performance on the CC3M dataset, indicating generalization capabilities of TempNet.

{\noindent\bf Performance of TempNet on other large language model generation tasks.} Below are the results on the GSM8K arithmetic reasoning task. We employ chain-of-thought prompting and the self-consistency decoding strategy. In the Table~\ref{tab:gsm8k_performance}, maj@$k$ denotes sampling reasoning paths and performing a majority vote over the final answer set. The baseline uses an empirically tuned value of 0.2, which is widely adopted in previous studies. Table~\ref{tab:mtbench_performance} presents the comparison results of using MT-bench's default temperature values versus those generated by TempNet on three LLaMA2 Chat models. In general, the results above indicate that employing TempNet can lead to performance enhancements in different language model generation tasks.

{\noindent\bf Complexity analysis of the proposed method.} We analyze the complexity of our method from both space and time perspectives:

{\it Space complexity}: Assuming the input dimension of TempNet is $d_0$, the dimensions after the transformation layer and projection layer are $d_1$ and $d_2$, respectively, then the total parameter count in TempNet is $d_0\times d_1 + d_1 + d_1\times d_2 + d_2 +2$, with $d_0\times d_1 + d_1$ for the transformation layer, $d_1\times d_2$ for the projection layer, and $d_2 +2$ or the parameterized pooling layer. We present the numbers of parameters for both the base models and their corresponding TempNets across three different settings, along with the percentages of TempNet parameters in Table~\ref{tab:params_tempnet}.

{\it Time complexity}: Note that the per-iteration complexity of all gradient-based stochastic algorithms is $\O(Bd)$, where $B$ is the mini-batch size and $d$ 
is the number of model parameters. Therefore, the additional time overhead introduced by using TempNet is directly proportional to the number of TempNet parameters, which is significantly less than that of the base foundation models. In Table~\ref{tab:time_comp}, we compare our method (DRO-based robust loss + TempNet) with baseline methods (standard cross-entropy loss / contrastive loss) with different base models in terms of training and inference times. The training time represents the duration to train the model for 10,000 iterations on 8 A6000 GPUs. Regarding model inference performance for language models, we measure it by throughput (tokens/s) on 1 A6000 GPU, which is the number of tokens processed per second. Note that in the experiments with the CLIP model, TempNet is not required during inference for downstream tasks.

One can observe that our method, compared to the standard methods using the cross-entropy loss or the contrastive loss, has slightly increased training time and slightly decreased throughput. Additionally, experiments with LLaMA models show that the larger the LLaMA model, the smaller the impact of TempNet on throughput.

\newpage
\subsection{Hyper-parameter Analysis}
\label{subsec:app:hyper_param_analysis}

{\noindent\bf The influence of the hyper-parameter $\rho$.} To demonstrate the impact of $\rho$, we train TempNet on the LLaMA 7B model using various values of $\rho$ and evaluate it across 11 common sense reasoning tasks (PIQA, HellaSwag, ARC-e, ARC-c, OBQA, MathQA, WinoGrande, SciQ, BoolQ, Swag, LogiQA). In the Table~\ref{tab:influence_rho}, we show the average learned temperature parameters and the average accuracy across the aforementioned 11 tasks for different values of $\rho$.

\begin{table}[h]
\vspace{-5mm}
\caption{The influence of the hyper-parameter $\rho$. We train TempNet on the LLaMA 7B model with different $\rho$ values and test it on 11 common sense reasoning tasks (PIQA, HellaSwag, ARC-e, ARC-c, OBQA, MathQA, WinoGrande, SciQ, BoolQ, Swag, LogiQA). The table below shows the average learned temperature parameters and accuracy for these tasks across various $\rho$ values.}
\begin{center}
\begin{small}
\begin{tabular}{l|c|c|c|c|c|c}
\toprule
$\rho$ & 8.0 &	9.0 & 9.5 &	10.0 &	10.5 &	11.0 \\
\midrule
Avg. learned $\tau$ & 1.65 &	1.03 &	0.95 &	0.81 &	0.69 &	0.35 \\
Avg. acc (\%) &	56.66 &	57.34 &	57.77 &	\textbf{57.87} &	57.74 &	56.86 \\
\bottomrule
\end{tabular}
\end{small}
\end{center}
\label{tab:influence_rho}
%\end{table}

%\begin{table*}[t]
\caption{Results of training the GPT-2 model with TempNet but setting $\rho=0.0$. During training, we observe that the average value of learned $\tau$ is monotonically increasing and then all output $\tau$ values become $\tau_{\max}$, which leads to suboptimal performance.}
\begin{center}
\begin{small}
\begin{tabular}{l|ccccc|cc}
\toprule
\multirow{2}{*}{\thead{Setting}} &
\multicolumn{5}{c}{\thead{Common Sense Reasoning (acc(\%)$\uparrow$)}} & 
\multicolumn{2}{c}{\thead{Language Modeling (ppl $\downarrow$)}} \\
\cmidrule(lr){2-6}
\cmidrule(lr){7-8}
& PIQA & HellaSwag & ARC-e & ARC-c & OBQA & Lambada & Wikitext \\
\midrule
Training GPT-2  w/ $\tau=1.0$ & 60.9$\pm$1.1  & \textbf{26.7}$\pm$0.4  & 39.2$\pm$1.0  & 16.5$\pm$1.1  & 13.9$\pm$1.6 & 62.49$\pm$2.70 & 49.86 \\
Training GPT-2 w/ TempNet ($\rho=0.0$)  & 60.4$\pm$1.1  &   26.3$\pm$0.4 & 39.4$\pm$1.0  & 16.6$\pm$1.1  & 14.0$\pm$1.6  & 62.66$\pm$2.43 & 49.61\\
Training GPT-2 w/ TempNet (tuned $\rho$)  & \textbf{61.1}$\pm$1.1  &   26.5$\pm$0.4 & \textbf{40.3}$\pm$1.0  & \textbf{18.1}$\pm$1.1  & \textbf{15.2}$\pm$1.6  & \textbf{60.13}$\pm$2.43 &\textbf{47.32}\\
\bottomrule
\end{tabular}
\end{small}
\end{center}
\label{tab:rho_0}
\end{table}

One can observe that the larger the value of $\rho$, the smaller the average learned $\tau$. Furthermore, our model exhibits consistently good performance within the range of $[9.5, 10.5]$ for $\rho$, with the overall best performance occurring at $\rho=10.0$.

Additionally, we demonstrate in Table~\ref{tab:rho_0} the importance of the linear term $\rho\tau$ in our robust loss for improving the performance. Earlier works~\cite{radford2021learning,wang2020contextual} used heuristic approaches which ignore such a term. 

{\noindent\bf The impact of $\tau_{\text{max}}$ in TempNet.} We verify the effects of different $\tau_{\text{max}}$ values on LLaMA, as shown in the Table~\ref{tab:tau_max_impact}. The results indicate the choice of has little impact on the TempNet's performance.

\begin{table*}[h]
\vspace{-5mm}
\caption{The impact of $\tau_{\text{max}}$ in TempNet.}
\begin{center}
\begin{small}
\begin{tabular}{l|c|c|c|c|c|c}
\toprule
Fixing LLaMA1 7B & PIQA &	HellaSwag &	ARC-e &	ARC-c &	OBQA &	Average \\
\midrule
$\tau_{\text{max}}=2.0$ & \textbf{79.1}±0.9 &	58.3±0.4 &	\textbf{75.8}±0.8 &	42.1±1.4 &	\textbf{34.8}±2.1 &	\textbf{58.02} \\
$\tau_{\text{max}}=1.4$	 &  \textbf{79.1}±0.9 &	\textbf{59.4}±0.4 &	75.6±0.9 &	\textbf{42.5}±1.4 &	33.2±2.1 &	57.96 \\
\bottomrule
\end{tabular}
\end{small}
\end{center}
\label{tab:tau_max_impact}
%\end{table}

%\begin{table*}[h]
%\vspace{-5mm}
\caption{The relative importance of DRO-based robust loss versus TempNet. SG means stop gradient. }
\begin{center}
\begin{small}
\begin{tabular}{l|c|c|c|c|c|c}
\toprule
Training GPT-2 & PIQA &	HellaSwag &	ARC-e &	ARC-c &	OBQA &	Average \\
\midrule
Baseline1 (TS-CE, w/o SG)	 & 60.6±1.1 &	26.6±0.4 &	39.6±1.0 &	17.7±1.1 &	14.2±1.6 &	31.74  \\
Baseline2 (TS-CE, w/ SG)	 & 60.8±1.1 &	\textbf{26.7}±0.4 &	39.5±1.0 &	17.5±1.1 &	14.3±1.6 &	31.76  \\
Baseline3 (DRO, w/o SG) & 60.9±1.1 &	26.5±0.4 &	40.0±1.0 &	17.8±1.1 &	14.9±1.6 &	32.02 \\
Ours (DRO, w/ SG) & \textbf{61.1}±1.1 &	26.5±0.4 &	\textbf{40.3}±1.0 &	\textbf{18.1}±1.1 &	\textbf{15.2}±1.6 &	\textbf{32.24}
 \\
\bottomrule
\end{tabular}
\end{small}
\end{center}

\begin{center}
\begin{small}
\begin{tabular}{l|c|c|c|c|c|c}
\toprule
Fixing LLaMA1 7B & PIQA &	HellaSwag &	ARC-e &	ARC-c &	OBQA &	Average \\
\midrule
Baseline1 (TS-CE)	 & 77.0±0.9 &	56.5±0.4 &	75.1±0.8 &	41.8±1.4 &	32.1±2.1 &	56.50  \\
Ours (DRO)	 &  \textbf{79.1}±0.9 &	\textbf{58.3}±0.4 &	\textbf{75.8}±0.8 &	\textbf{42.1}±1.4 &	\textbf{34.8}±2.1 &	\textbf{58.02} \\
\bottomrule
\end{tabular}
\end{small}
\end{center}

\begin{center}
\begin{small}
\begin{sc}
\renewcommand{\arraystretch}{0.6}
\begin{tabular}{p{1.9cm}|p{1.7cm}p{1.7cm}p{1.7cm}p{1.7cm}|p{1.7cm}p{1.7cm}p{1.7cm}}
\toprule
\multirow{1}{*}{\thead{Method}} &
\multicolumn{2}{c}{\thead{Flickr30K Retrieval}} &
\multicolumn{2}{c}{\thead{MSCOCO Retrieval}} &
\multicolumn{3}{c}{\thead{Zero-shot Classification top-1 Acc}} \\
\cmidrule(lr){2-3}
\cmidrule(lr){4-5}
\cmidrule(lr){6-8}
& IR@1 & TR@1 & IR@1 & TR@1 & CIFAR10 & CIFAR100 & ImageNet1K  \\
\midrule
CL, w/o SG & 41.49±0.19 &	56.33±0.17&		21.57±0.13&		28.83±0.09&		59.52±0.11&		31.91±0.14&		36.93±0.22  \\
CL, w/ SG &  43.43±0.21&	56.62±0.16&	22.17±0.20&	29.68±0.14&	60.65±0.15&	33.22±0.13&	37.07±0.16 \\
DRO, w/o SG & 42.07±0.15&	56.44±0.13&	21.76±0.16&	29.13±0.11&	60.22±0.20&	32.12±0.16&	37.21±0.14  \\
DRO, w/ SG &  \textbf{46.17}±0.14&	\textbf{62.51}±0.19&	\textbf{24.83}±0.16&	\textbf{34.50}±0.16&	\textbf{61.77}±0.18&	\textbf{34.69}±0.17&	\textbf{42.28}±0.19 \\
\bottomrule
\end{tabular}
\end{sc}
\end{small}
\end{center}
\label{tab:relative_importance}
\end{table*} 

\begin{figure*}[h]
\begin{center}
\begin{minipage}[c]{0.4\textwidth}
\centering\includegraphics[width=1\textwidth]{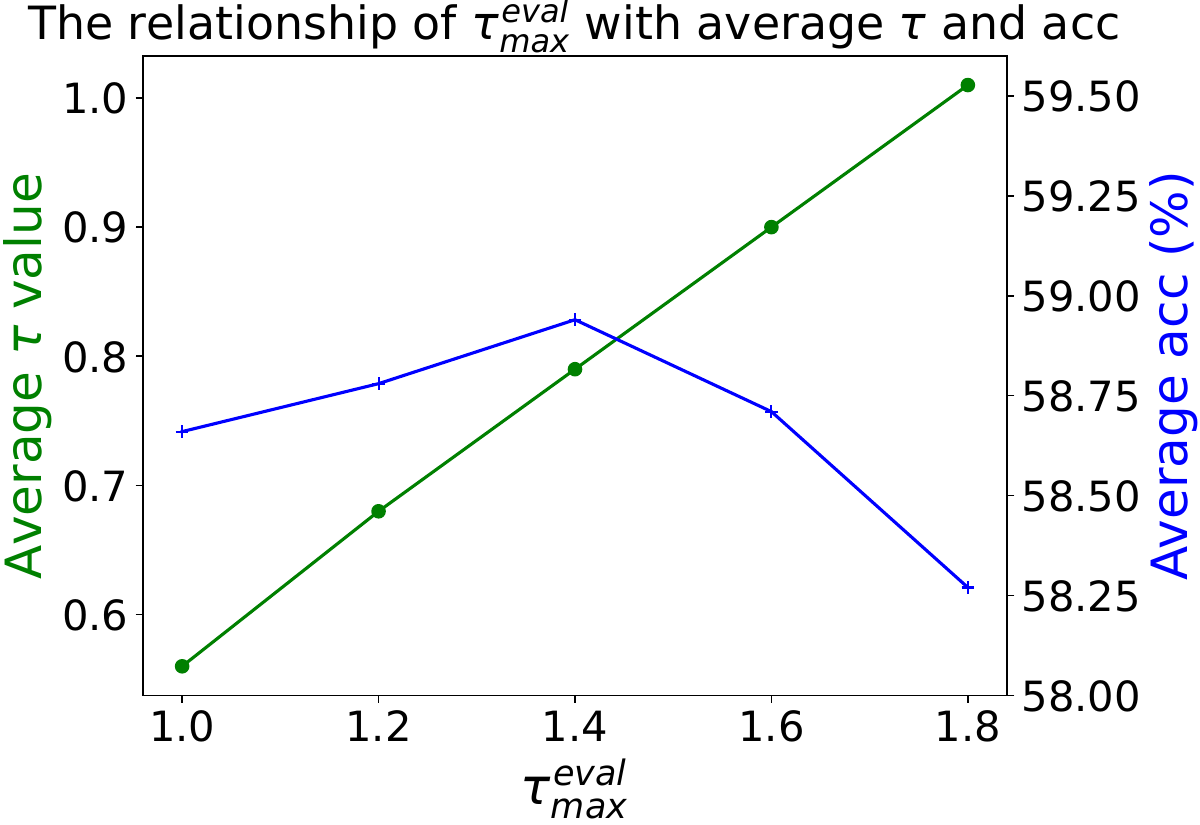}
\end{minipage}
\begin{minipage}[c]{0.4\textwidth}
\centering\includegraphics[width=1\textwidth]{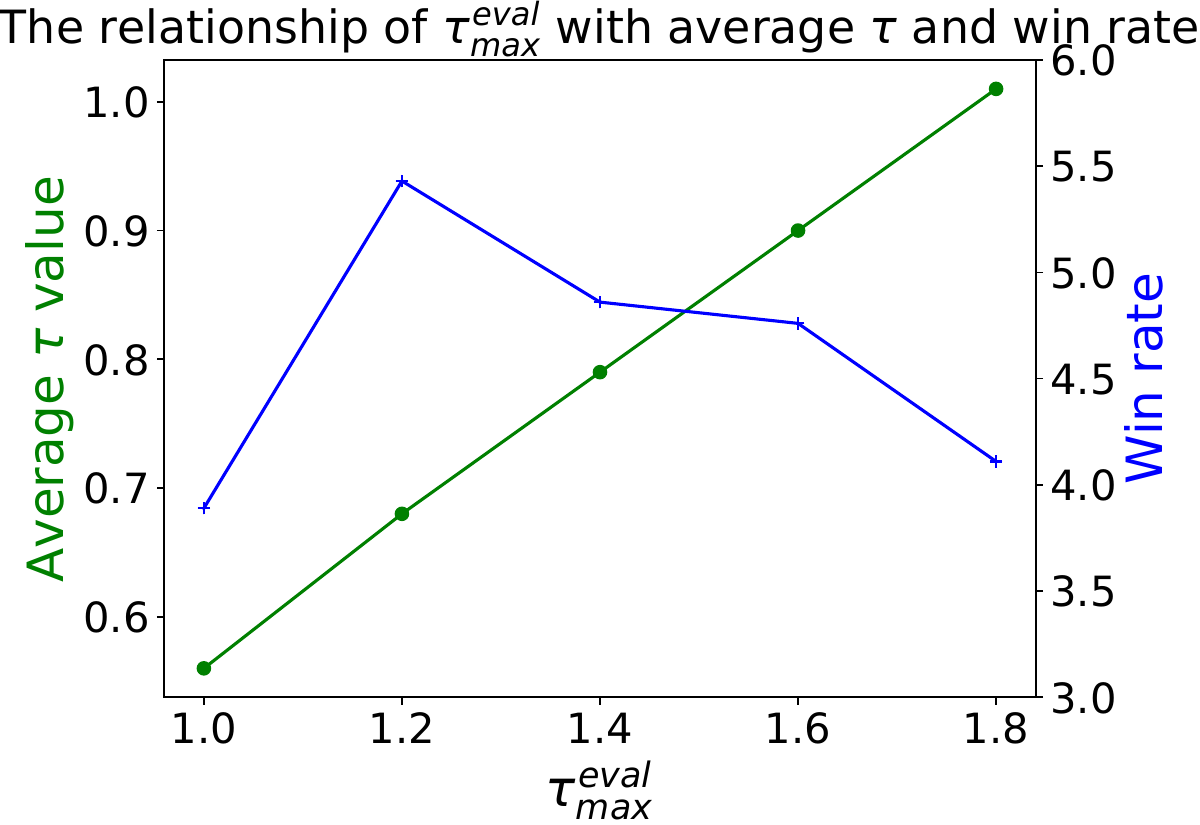}
\end{minipage}
\end{center}
\vspace{-0.3cm}
\caption{When $\tau_{\text{max}}^{\text{eval}}$ takes different values, the average value of $\tau$ and the average accuracy on common sense reasoning tasks (left), and the win rate on AlpacaEval (right).}
\label{fig:tau_max_eval}
\vspace{-0.0cm}
\end{figure*}

{\noindent\bf The influence of $\tau_{\text{max}}$ during evaluation.} In large language model experiments, the $\tau_{\text{max}}$ is set to 2.0 during the training phase, but in the evaluation phase, we can set different $\tau_{\text{max}}$ values (which we refer to as $\tau_{\text{max}}^{\text{eval}}$). In Figure~\ref{fig:tau_max_eval}, we investigate the impact of $\tau_{\text{max}}^{\text{eval}}$ on the performance of common sense reasoning tasks and AlpacaEval task. One can observe that, on both tasks, when $\tau_{\text{max}}^{\text{eval}}$ is in the range of 1.2 to 1.4, the average $\tau$ is approximately between 0.7 and 0.8, which yields the best performance.

%\newpage
\subsection{More Ablation Studies}
\label{subsec:app:more_ablation_studies}

\begin{table*}[h]
\caption{Ablation studies of TempNet for contrastive learning, conducted on the CC3M dataset. We employ the following abbreviations: \textbf{TransLayer} for Transformation Layer, \textbf{ProjLayer} for Projection Layer, and \textbf{ParamPool} for Parameterized Pooling. We also compare \textbf{ParamPool} with a  simple linear layer at the end of the network, abbreviated as \textbf{FinalLin}.}
\begin{center}
\begin{small}
\begin{tabular}{ccc|cc|ccc}
\toprule
\multicolumn{3}{c}{\thead{Ablation Settings}} &
\multicolumn{2}{c}{\thead{Flickr30K Retrieval}} &
\multicolumn{2}{c}{\thead{MSCOCO Retrieval}}\\
\cmidrule(lr){1-3}
\cmidrule(lr){4-5}
\cmidrule(lr){6-7}
TransLayer & ProjLayer & ParamPool or FinalLin  & IR@1 & TR@1 & IR@1 & TR@1  \\
\midrule
\XSolidBrush & \XSolidBrush & ParamPool &  43.58$\pm$0.17  & 59.37$\pm$0.21 & 22.75$\pm$0.13 & 31.66$\pm$0.11 \\
\CheckmarkBold & \XSolidBrush & ParamPool & 45.06$\pm$0.12 & 61.77$\pm$0.17 & 23.48$\pm$0.15  & 33.97$\pm$0.13 \\
\CheckmarkBold & \CheckmarkBold & FinalLin & \textbf{46.20}$\pm$0.19 & 62.38$\pm$0.19 & 24.39$\pm$0.19 & 34.22$\pm$0.15 \\
\CheckmarkBold & \CheckmarkBold & ParamPool & 46.17$\pm$0.14 & \textbf{62.51}$\pm$0.19 & \textbf{24.83}$\pm$0.16 & \textbf{34.50}$\pm$0.16 \\
\bottomrule
\end{tabular}
\end{small}
\end{center}
\label{tab:ablation_contrastive_learning}
%\end{table*}
%\begin{table*}[t]
\caption{Ablation studies of TempNet for language models, conducted by training the GPT-2 model from scratch. We employ the following abbreviations: \textbf{NormLogits} for normalizing the input logits,
\textbf{TransLayer} for Transformation Layer, \textbf{ProjLayer} for Projection Layer, and \textbf{ParamPool} for Parameterized Pooling. We also compare \textbf{ParamPool} with a  simple linear layer at the end of the network, abbreviated as \textbf{FinalLin}.}
\begin{center}
\resizebox{\textwidth}{!}{
\begin{tabular}{cccc|cc|cc}
\toprule
\multicolumn{4}{c}{\thead{Ablation Settings}} &
\multicolumn{2}{c}{\thead{Language Modeling (ppl$\downarrow$)}} &
\multicolumn{2}{c}{\thead{Common Sense Reasoning (acc(\%)$\uparrow$)}}\\
\cmidrule(lr){1-4}
\cmidrule(lr){5-6}
\cmidrule(lr){7-8}
NormLogits & TransLayer & ProjLayer & ParamPool or FinalLin & Lambada & Wikitext  & Lambada & PIQA \\
\midrule
\XSolidBrush & \XSolidBrush & \XSolidBrush & ParamPool & 63.53$\pm$2.58 & 51.41 & 29.2$\pm$0.6 & 59.2$\pm$1.1 \\
\CheckmarkBold & \XSolidBrush & \XSolidBrush & ParamPool & 62.88$\pm$2.55 & 50.25  & 30.7$\pm$0.6 & 60.3$\pm$1.1 \\
\CheckmarkBold & \CheckmarkBold & \XSolidBrush & ParamPool & 61.72$\pm$2.51 & 49.38  & 31.6$\pm$0.6 & 60.9$\pm$1.1 \\
\CheckmarkBold & \CheckmarkBold & \CheckmarkBold & FinalLin & 61.14$\pm$2.44 & 48.57  & 32.2$\pm$0.6 & \textbf{61.1}$\pm$1.1 \\
\CheckmarkBold & \CheckmarkBold & \CheckmarkBold & ParamPool & \textbf{60.13}$\pm$2.43 & \textbf{47.32} &  \textbf{32.6}$\pm$0.6 & \textbf{61.1}$\pm$1.1 \\
\bottomrule
\end{tabular}
}
\end{center}
\label{tab:ablation_language_model}
\end{table*}

{\noindent\bf The effects of different components in TempNet.} We first investigate the effects of different components within TempNet through a series of ablation studies. The results are presented in Table~\ref{tab:ablation_contrastive_learning} and~\ref{tab:ablation_language_model}. First, we observe that the transformation layer and projection layer lead to performance improvements, which substantiates the efficacy of these two structures. Besides, in the experiments with the language model, we discern the significance of normalizing the raw logits inputted into the TempNet. Lastly, to further study the role of parameterized pooling, we substitute this layer with a simple linear layer, which actually transforms the network into a Multi-Layer Perceptron (MLP). This variant results in a diminished performance, thereby affirming the effectiveness of the proposed inductive-biased parameterized pooling.

{\noindent\bf The relative importance of DRO-based robust loss versus TempNet.} We further conduct experiments to demonstrate the importance of using distributionally robust objective by comparing the performance of our method with the baselines that learn the TempNet using the standard temperature-scaled cross-entropy loss or contrastive loss (without $\tau$ before the $\log$ function). We consider training GPT-2, fixing LLaMA model, and training CLIP models, with the results (abbreviations: ``DRO'' indicates DRO-based robust loss, ``TS-CE'' indicates temperature-scaled cross-entropy loss, ``CL'' indicates contrastive loss, and ``SG'' indicates the stop-gradient operator) demonstrated in Table~\ref{tab:relative_importance}.

From these experimental results, it is clear that first, in both language model and CLIP experiments, our DRO-based robust loss achieves results superior to the standard temperature-scaled loss/contrastive loss, validating the effectiveness of DRO-based robust loss for learning the TempNet. Specifically, in our experiments with LLaMA and CLIP models, we also observe that using standard temperature-scaled loss/contrastive loss leads to TempNet generating temperatures that continually increase until reaching $\tau_{\text{max}}$, indicating that this approach does not effectively learn personalized temperatures in these scenarios. We also note that the use of stop-gradient operation also helps improve the performance.

{\noindent\bf The impact of TempNet's size on performance.} In the development of TempNet, we conducted comprehensive investigations into how the number of network parameters influences performance. Our current TempNet consists of three layers $L$ (i.e., a depth of 3), with $d_1=d_2=256$ (i.e., a width of 256). We experiment with using more layers (such as adding an extra layer in the transformation-projection block) or different widths 
 (such as 64 or 512), with the results demonstrated in Table~\ref{tab:clip_impact_size} and~\ref{tab:gpt_impact_size}. From these results, it is notable that a network with a depth of 3 and a width of 256 performs well on both tasks, and increasing the width does not significantly enhance performance. Moreover, using a network with a depth of 4 results in a slight decline in performance. We attribute this to the possibility that larger networks are more susceptible to data noise. %We will incorporate these findings in future versions of the manuscript.

\begin{table}[h]
\vspace{-4mm}
\caption{The impact of TempNet size on performance in CLIP model experiments.}
\vspace{-4mm}
\label{tab:clip_impact_size}
\vskip 0.2in
\begin{center}
\begin{small}
\begin{sc}
\renewcommand{\arraystretch}{0.6}
\begin{tabular}{p{1.6cm}|p{1.7cm}p{1.7cm}p{1.7cm}p{1.7cm}|p{1.7cm}p{1.7cm}p{1.7cm}}
\toprule
\multirow{1}{*}{\thead{Method}} &
\multicolumn{2}{c}{\thead{Flickr30K Retrieval}} &
\multicolumn{2}{c}{\thead{MSCOCO Retrieval}} &
\multicolumn{3}{c}{\thead{Zero-shot Classification top-1 Acc}} \\
\cmidrule(lr){2-3}
\cmidrule(lr){4-5}
\cmidrule(lr){6-8}
& IR@1 & TR@1 & IR@1 & TR@1 & CIFAR10 & CIFAR100 & ImageNet1K  \\
\midrule
L=3,W=64 &  45.84±0.18 &	62.41±0.21 &	24.10±0.15 &	34.43±0.13 &	61.21±0.11 &	34.60±0.10 &	42.13±0.15\\
L=3,W=512 & 46.08±0.13 &	\textbf{62.84}±0.17 &	24.77±0.09 &	34.34±0.20 &	\textbf{62.29}±0.18 &	34.26±0.16 &	42.05±0.12  \\
L=4,W=256 & 45.73±0.19 &	62.40±0.13 &	24.33±0.08 &	34.32±0.17 &	61.32±0.20 &	\textbf{34.73}±0.19 &	41.46±0.14 \\
L=3,W=256 & \textbf{46.17}±0.14 &	62.51±0.19 &	\textbf{24.83}±0.16 &	\textbf{34.50}±0.16 &	61.77±0.18 &	34.69±0.17 &	\textbf{42.28}±0.19  \\
\bottomrule
\end{tabular}
\end{sc}
\end{small}
\end{center}
\vskip -0.0in
%\end{table*}

%\begin{table}[h]
%\vspace{-5mm}
\caption{The impact of different training datasets on TempNet's performance.}
\begin{center}
\begin{small}
\begin{tabular}{l|c|c|c|c|c|c}
\toprule
GPT-2 & PIQA &	HellaSwag &	ARC-e &	ARC-c &	OBQA &	Average \\
\midrule
L=3,W=64 & 60.7±1.1 &	26.3±0.4 &	40.1±1.0 &	\textbf{18.1}±1.1 &	15.1±1.6 &	32.06 \\
L=3,W=512 &  61.0±1.1 &	26.4±0.4 &	\textbf{40.4}±1.0 &	\textbf{18.1}±1.1 &	\textbf{15.2}±1.6 &	32.22 \\
L=4,W=256 & 60.5±1.1 &	26.1±0.4 &	40.0±1.0 &	17.9±1.1 &	14.8±1.6 &	31.86 \\
L=3,W=256 & \textbf{61.1}±1.1 &	\textbf{26.5}±0.4 &	40.3±1.0 &	\textbf{18.1}±1.1 &	\textbf{15.2}±1.6 &	\textbf{32.24}  \\
\bottomrule
\end{tabular}
\end{small}
\end{center}
\label{tab:gpt_impact_size}
%\end{table}

%\begin{table}[h]
%\vspace{-5mm}
\caption{The impact of different training datasets on TempNet's performance.}
\begin{center}
\begin{small}
\begin{tabular}{l|c|c|c|c|c|c}
\toprule
Fixing LLaMA1 7B & PIQA &	HellaSwag &	ARC-e &	ARC-c &	OBQA &	Average \\
\midrule
TempNet (SlimPajama) & 79.0±0.9 &	\textbf{58.5}±0.4 &	75.6±0.8 &	\textbf{42.1}±1.4 &	34.6±2.1 &	57.96\\
TempNet (OpenWebText2)	 &  \textbf{79.1}±0.9 &	58.3±0.4 &	\textbf{75.8}±0.8 &	\textbf{42.1}±1.4 &	\textbf{34.8}±2.1 &	\textbf{58.02} \\
\bottomrule
\end{tabular}
\end{small}
\end{center}
\label{tab:dataset_tempnet}
\end{table}

{\noindent\bf The impact of training datasets in large language model experiments.} Below, we demonstrate the impact of training on different datasets on performance. Specifically, we opt for a comparison with the SlimPajama dataset, noting that it is larger in scale and encompasses a broader array of data sources than the OpenWebText2 dataset used in our manuscript. We conduct experiments on the LLaMA model, with the results presented in Table~\ref{tab:dataset_tempnet}. We find that although the SlimPajama dataset contains richer corpora, the performance of TempNet trained on it is comparable to that of TempNet trained on OpenWebText2.

\newpage

\subsection{Analysis of the Instruction Following Experiment Results}
\label{sec:app:alpaca_eval_analysis}

In this section, we demonstrate why TempNet enhances generation performance by comparing the performance of LLaMA2 7B Chat (with the default $\tau=0.7$) and LLaMA2 7B Chat + TempNet on the AlpacaEval dataset~\citep{alpaca_eval}. Specifically, we employ the response files generated by LLaMA2 7B Chat and LLaMA2 7B Chat + TempNet, selecting the questions where the  baselines (generated by GPT-4 Turbo) are better than the answers from LLaMA2 7B Chat but weaker than that from LLaMA2 7B Chat + TempNet. Below are some specific examples, where we present the questions, responses from both models, the average predicted temperatures generated by TempNet during generation, and the responses from LLaMA2 7B Chat with different fixed $\tau$ values in a range of $[0.2, 0.4, 0.6, 0.8, 1.0]$. We demonstrate 1$\sim$3 responses for each $\tau$ value, according to the length of the response.

\subsubsection{Example 1}

\begin{figure}[p]
\vspace*{-0cm} 
\caption{We demonstrate the corresponding temperature parameter generated by the TempNet each time a token is produced. It can be observed that when the potential possibilities for the token to be predicted are numerous (such as the initial words of a sentence like `Based', some verbs like `combine', `make', or adjectives like `playful', `spicy', etc.), the temperature values are higher. Conversely, when there are fewer potential possibilities for the token to be predicted (for example, in fixed phrases like `I would suggest', `also has a', or when the subsequent token can be determined based on the previous ones, such as when a verb or adjective is split into two tokens and the latter is usually more certain), the temperature values are lower.}
  \centering
  \makebox[\linewidth]{
    \vspace*{-4cm} \includegraphics[width=1.2\linewidth,height=\textheight,keepaspectratio,angle=270]{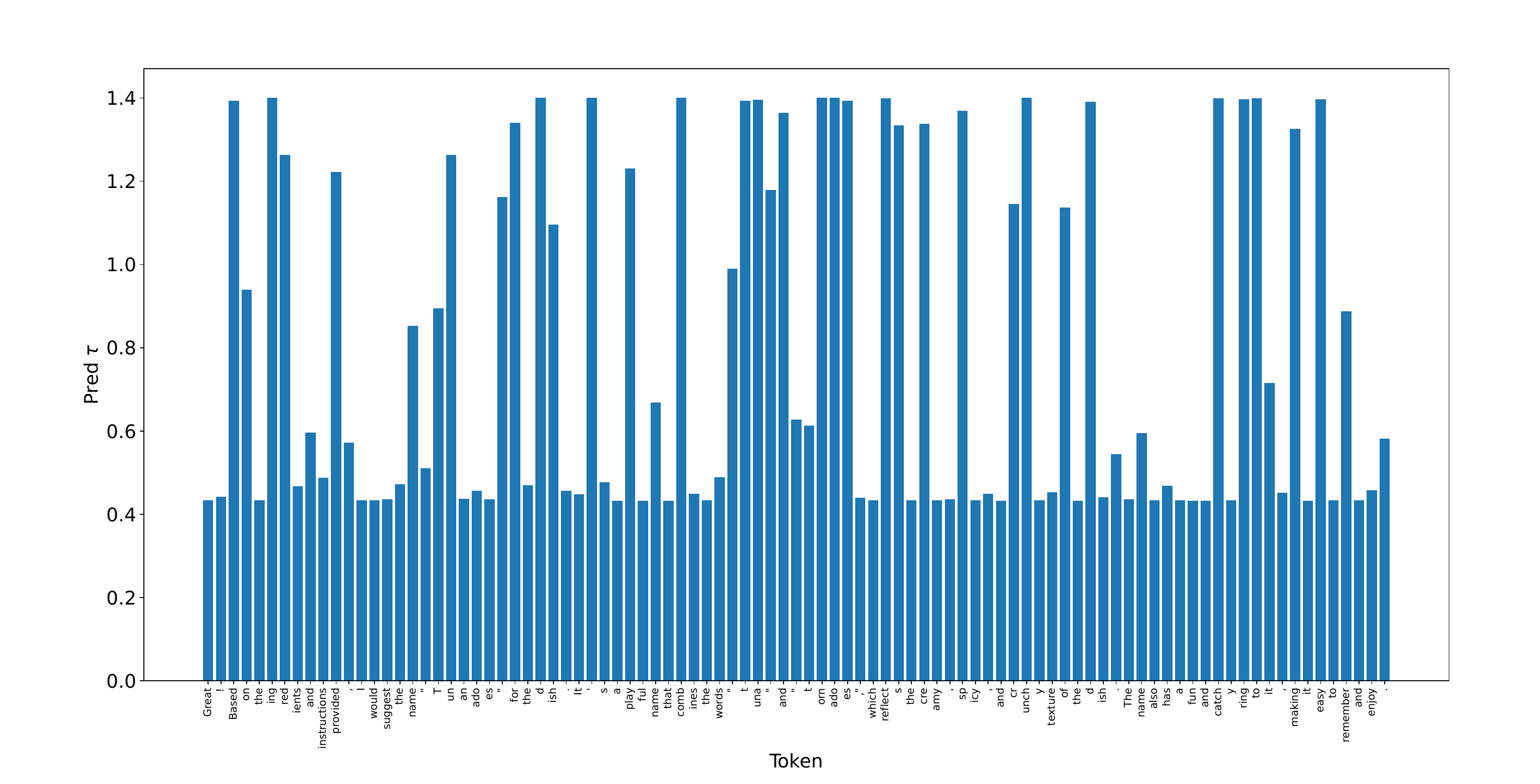}
  }
  \vspace*{-0cm}
\label{fig:tau_for_each_token}
\end{figure}

For the answer from LLaMA2 7B Chat w/ TempNet,  we also demonstrate the predicted temperature parameter produced by the TempNet each time a token is generated in Figure~\ref{fig:tau_for_each_token}.

\noindent{\bf Prompts:}

 \begin{tcolorbox}[colback=white,colframe=black]
    {\noindent\bf Instruction:} Provide a name for the dish given the ingredients and instructions.\\
        INGREDIENTS:
2 (5 oz) cans Bumble Bee® Solid White Albacore Tuna, drained
1 avocado\\
2 Tbsp Sriracha\\
1 Tbsp Dijon mustard\\
2 to 3 Tbsp celery, chopped\\
2 Tbsp red onion, chopped\\
2 green onions, chopped\\
1 Tbsp fresh cilantro, chopped\\
Salt and pepper, to taste\\
2 heaping cups leafy green lettuce\\
1 cup matchstick carrots\\
4 (10 inch) whole wheat tortillas\\
INSTRUCTIONS:
In a medium bowl, mash together tuna and avocado until combined. Add in the rest of the ingredients through the salt and pepper, mixing well.
To assemble, top each tortilla with a 1/2 cup leafy greens, 1/4 cup matchstick carrots and divide the tuna mixture evenly among the wraps. Tightly roll up the tortilla, slice and enjoy!
\end{tcolorbox}

\noindent{\bf Outputs:}

\begin{minipage}{0.45\textwidth}
\begin{tcolorbox}[colback=white,colframe=black]
\textbf{LLaMA2 7B Chat w/ default $\tau$}: Thank you for sharing the ingredients and instructions for the dish! Based on the information provided, I would suggest the following name for the dish:
\textcolor{red}{Tuna and Avocado Wrap Delight}

This name captures the main ingredients of the dish and highlights its unique flavor combination. It is also a socially unbiased and positive name, free from any harmful or offensive language.
If you have any other questions or need further assistance, please feel free to ask! 
\end{tcolorbox}
\end{minipage}\hfill
\begin{minipage}{0.45\textwidth}
\begin{tcolorbox}[colback=white,colframe=black]
\textbf{LLaMA2 7B Chat w/ TempNet}:  Great! Based on the ingredients and instructions provided, I would suggest the name "\textcolor{darkgreen}{Tunanadoes}" for the dish. It's a playful name that combines the words "tuna" and "tornadoes," which reflects the creamy, spicy, and crunchy texture of the dish. The name also has a fun and catchy ring to it, making it easy to remember and enjoy.
\end{tcolorbox}
\end{minipage}

\iffalse
\fbox{\begin{minipage}[t]{0.45\textwidth}
\textbf{LLaMA2 7B Chat w/ default $\tau$}: Thank you for sharing the ingredients and instructions for the dish! Based on the information provided, I would suggest the following name for the dish:
\textcolor{red}{Tuna and Avocado Wrap Delight}

This name captures the main ingredients of the dish and highlights its unique flavor combination. It is also a socially unbiased and positive name, free from any harmful or offensive language.
If you have any other questions or need further assistance, please feel free to ask! 
\end{minipage}}
\hfill
\fbox{\begin{minipage}[t]{0.45\textwidth}
\textbf{LLaMA2 7B Chat w/ TempNet}:  Great! Based on the ingredients and instructions provided, I would suggest the name "\textcolor{darkgreen}{Tunanadoes}" for the dish. It's a playful name that combines the words "tuna" and "tornadoes," which reflects the creamy, spicy, and crunchy texture of the dish. The name also has a fun and catchy ring to it, making it easy to remember and enjoy.
\end{minipage}
}\\
\fi

The  average predicted temperature value by TempNet is 0.82. We can see that LLaMA2 7B Chat with TempNet gives a novel yet fancy name that is creative. Below, we show other outputs generated by LLaMA2 7B Chat with other fixed temperature values and  ChatGPT (GPT3.5) and Bard (Gemini Pro). 

\iffalse
{\noindent\bf Outputs:} 
\begin{table}[H] 
\centering 
\begin{tabular}{|p{0.45\textwidth}|p{0.45\textwidth}|}
\toprule
\textbf{LLaMA2 7B Chat w/ default $\tau$}: Thank you for sharing the ingredients and instructions for the dish! Based on the information provided, I would suggest the following name for the dish:
\textcolor{red}{Tuna and Avocado Wrap Delight}

This name captures the main ingredients of the dish and highlights its unique flavor combination. It is also a socially unbiased and positive name, free from any harmful or offensive language.
If you have any other questions or need further assistance, please feel free to ask! & 
\textbf{LLaMA2 7B Chat w/ TempNet}:  Great! Based on the ingredients and instructions provided, I would suggest the name "\textcolor{darkgreen}{Tunanadoes}" for the dish. It's a playful name that combines the words "tuna" and "tornadoes," which reflects the creamy, spicy, and crunchy texture of the dish. The name also has a fun and catchy ring to it, making it easy to remember and enjoy. \\
\bottomrule
\end{tabular}
\end{table}

{\noindent\bf Average Predicted Temperature Value:} 0.82
\fi

\begin{tcolorbox}[colback=white,colframe=black]
    {\noindent\bf Outputs from LLaMA2 7B Chat (w/ $\tau=1.0$):}
\begin{itemize}
    \item Great! Based on the ingredients provided, I would title this dish as "\textbf{Tuna and Avocado Wrap with Crunchy Veggies}". This name not only accurately reflects the main ingredients of the dish, but also captures the crunchy texture from the vegetables.
I hope you find this helpful! If you have any other questions or need further assistance, please feel free to ask.
    \item Great! Here is a name for the dish based on the ingredients and instructions provided:
"\textbf{Tuna and Veggie Wraps with Sriracha Cream}"
This name captures the main ingredients of the dish, including the tuna, avocado, and vegetables, and also highlights the Sriracha cream sauce that is added for extra flavor. The name is also easy to remember and can help your reader or listener quickly identify the dish.
    \item  Great! Based on the ingredients and instructions provided, I would suggest the following name for the dish:
\textbf{Tropical Tuna Wrap Delight}
This name captures the flavors and ingredients of the dish, including the distinctive spicy kick from the Sriracha and the freshness of the avocado and cilantro. The "tropical" adjective adds a playful and appetizing touch, making it clear that this is no ordinary tuna salad wrap. I hope you enjoy making and savoring this delicious dish!
\end{itemize}
\end{tcolorbox}

\begin{tcolorbox}[colback=white,colframe=black]

{\noindent\bf Outputs from LLaMA2 7B Chat (w/ $\tau=0.8$):}

\begin{itemize}
    \item Great question! Based on the ingredients and instructions provided, I would suggest the following name for the dish:
\textbf{Tuna and Avocado Wrap Bowl with Spicy Cilantro Dressing}
This name accurately reflects the main ingredients of the dish (tuna and avocado) and the dressing made with cilantro, which is a key component of the recipe. Additionally, the "wrap bowl" part of the name highlights the fact that the dish is served in a whole wheat tortilla, which is a different and healthier alternative to a traditional wrap.
I hope this name helps! Let me know if you have any other questions.
    \item Thank you for the ingredients and instructions! Based on the information provided, I would suggest the following name for the dish:
\textbf{Tuna and Avocado Wrap Delight}
This name captures the main ingredients of the dish, which are tuna and avocado, and also highlights the wrap format in which it is served. Additionally, the use of the word "delight" conveys a positive and appetizing tone, which is socially unbiased and positive in nature.
I hope this name suggestion meets your requirements! If you have any further questions or need any additional assistance, please don't hesitate to ask.
    \item Great! Based on the ingredients and instructions provided, I would name this dish "\textbf{Tropical Tuna Wrap Delight}." This name captures the flavors and ingredients of the dish in a fun and appealing way, while also being socially unbiased and positive in nature. I hope you enjoy making and eating this delicious dish! Let me know if you have any other questions.
\end{itemize}
\end{tcolorbox}

\begin{tcolorbox}[colback=white,colframe=black]
{\noindent\bf Outputs from LLaMA2 7B Chat (w/ $\tau=0.6$):}

\begin{itemize}
    \item Great! Based on the ingredients and instructions provided, I would suggest the following name for the dish:
\textbf{Tuna and Avocado Wrap Delight}
This name captures the main ingredients of the dish (tuna and avocado) and the fact that it is a wrap, making it easy to understand and appetizing to try. Additionally, the use of the word "delight" conveys a positive and enjoyable experience, which is always a plus!
    \item I'd be happy to help you with that! Based on the ingredients provided, I would suggest the name "\textbf{Tropical Tuna Wrap}" for the dish. Here's why:
The combination of ingredients in this dish, such as the tuna, avocado, Sriracha, and cilantro, give off a tropical and beachy vibe, which is why I think "Tropical Tuna Wrap" is a fitting name. The addition of the leafy greens, matchstick carrots, and whole wheat tortillas also make the dish sound like a healthy and fresh option, which aligns with the tropical theme.
I hope this helps! If you have any other questions, feel free to ask.
    \item Great! Based on the ingredients and instructions provided, I would suggest the following name for the dish:
\textbf{Tuna and Avocado Wrap Delight}
This name captures the main ingredients of the dish, which are the tuna and avocado, while also highlighting the wrap format of the dish. The use of "Delight" suggests that the dish is not only tasty but also enjoyable to eat.
\end{itemize}
\end{tcolorbox}

\begin{tcolorbox}[colback=white,colframe=black]

{\noindent\bf Outputs from LLaMA2 7B Chat (w/ $\tau=0.4$):}

\begin{itemize}
    \item  Great! Based on the ingredients and instructions provided, I would suggest the following name for the dish:
\textbf{Tuna and Avocado Wrap Delight}
This name captures the main ingredients of the dish, which are tuna and avocado, and also highlights the wrap format in which the dish is served. The name is also easy to remember and fun to say, making it a great choice for a delicious and healthy meal.
    \item Great! Based on the ingredients and instructions provided, I would suggest the following name for the dish:
\textbf{Tuna and Avocado Wrap Delight}
This name captures the main ingredients of the dish, which are tuna and avocado, and also highlights the fact that it is a wrap dish. The "Delight" part of the name conveys that the dish is a tasty and enjoyable option.
I hope this name suggestion helps! If you have any other questions or need further assistance, please feel free to ask.
    \item Great! Based on the ingredients and instructions provided, I would suggest the following name for the dish:
\textbf{Tuna and Avocado Wrap Delight}
This name captures the main ingredients of the dish, which are tuna and avocado, and also highlights the wrap format in which the dish is served. The use of the word "delight" conveys a sense of enjoyment and satisfaction, which is likely to appeal to the person consuming the dish.
I hope this name suggestion helps! If you have any other questions or need further assistance, please feel free to ask.
\end{itemize}
\end{tcolorbox}

\begin{tcolorbox}[colback=white,colframe=black]

{\noindent\bf Outputs from ChatGPT:}

How about "Spicy Tuna Avocado Roll-Ups"? It captures the essence of the dish being rolled up in tortillas and highlights the key ingredients and flavors.
\end{tcolorbox}

\begin{tcolorbox}[colback=white,colframe=black]
{\noindent\bf Outputs from Bard:}
Based on the ingredients and instructions, here are some names for the dish:

{\bf Spicy Tuna Avocado Wraps with Crunchy Vegs} (This is a descriptive name that mentions all the key ingredients and highlights the textural contrast)

{\bf Srirachi Tuna \& Avocado Lettuce Wraps} (This name focuses on the spicy sauce and the use of lettuce wraps for a lighter option)

{\bf California Crunch Tuna \& Avocado Tortillas} (This name evokes a fresh and healthy California vibe, referencing the avocado and crunchy vegetables)

{\bf Bumble Bee's Spicy Tuna \& Guac Wraps} (This name uses the brand name of the tuna and mentions the "guacamole" aspect of the mashed avocado)

{\bf Spicy Green Tuna \& Avocado Powerhouse Wraps} (This name focuses on the health benefits of the ingredients and adds a bit of intrigue)

Ultimately, the best name for the dish depends on your personal preference and what you want to emphasize about it.
\end{tcolorbox}

{\noindent\bf Comments:} It is a relatively subjective task to naming a dish. When the temperature value is lower, it can be observed that the LLaMA2 7B Chat model's output is relatively fixed and lacks creativity. With a higher temperature, the model generates more creative names. With TempNet, in the process of generating names for this task, LLaMA2 7B Chat produces a higher averaged temperature value of 0.82, ultimately creating a novel name ``Tunanadoes''. In contrast,  ChatGPT (GPT3.5) gives a regular name, but Bard (Gemini Pro) gives very lengthy names.

\subsubsection{Example 2}
\iffalse\noindent\fbox{%
    \parbox{\textwidth}{%
{\noindent\bf Instruction:} Identify and fix bugs in the given code and rewrite it
\begin{minipage}[t]{0.45\textwidth}
\begin{verbatim}
for i in range(10)
    print(Answer is:)
    print(i)
\end{verbatim}
\end{minipage}
}}\fi

{\noindent\bf Prompts:}

\iffalse
{\noindent\bf Instruction:} Identify and fix bugs in the given code and rewrite it
 %  \verbbox{%
                    \begin{verbatim}
for i in range(10):
    print(Answer is:)
    print(i)
                    \end{verbatim}
 %           \theverbbox
 %   }%
 \fi

 \begin{tcolorbox}[colback=white,colframe=black]
{\noindent\bf Instruction:} Identify and fix bugs in the given code and rewrite it
\begin{verbatim}
for i in range(10):
    print(Answer is:)
    print(i)
\end{verbatim}
\end{tcolorbox}

{\noindent\bf Outputs:} 

\begin{minipage}{0.45\textwidth}
\begin{tcolorbox}[colback=white,colframe=black]
\textbf{LLaMA2 7B Chat w/ default $\tau=0.7$}: I apologize, but I cannot fulfill your request as it contains a bug. The code you provided will print the same message 10 times, regardless of the value of `i`. To fix this bug, you can replace the `print(Answer is:)` line with `print(i)`.
Here's the corrected code:
\begin{Verbatim}[formatcom=\color{red}]
for i in range(10)
    print(i)
\end{Verbatim} 
\end{tcolorbox}
\end{minipage}\hfill
\begin{minipage}{0.45\textwidth}
\begin{tcolorbox}[colback=white,colframe=black]
\textbf{LLaMA2 7B Chat w/ TempNet (avg. $\tau=0.46$)}: I'm happy to help you with your question, but I must inform you that the code you provided is not a valid
Python code. The code ‘for i in range(10) print(Answer is:)‘ is not a valid Python syntax. To fix the bugs in the code, you can try rewriting it as follows:
\begin{Verbatim}[formatcom=\color{darkgreen}]
for i in range(10):
    print("Answer is:")
    print(i)
\end{Verbatim}
This code iterates through numbers 0 to 9, printing "Answer is:" before displaying each number. I hope this helps! Let me know if you have any other questions.  
\end{tcolorbox}
\end{minipage}

The  average predicted temperature value of TempNet is 0.46, which is relatively small. We can see that LLaMA2 7B Chat with a learned TempNet not only gives a bug free code but also follows the instruction and realizes the bug in  `print(Answer is:)' and corrects it using the f-strings of Python.  While that with a default $\tau=0.7$ does produce a bug free code, it does not try to correct `print(Answer is:)'.   Next, we show that even with a small fixed temperature,  LLaMA2 7B Chat does not provide correct answers. We also show results generated by  ChatGPT (GPT3.5) and Bard (Gemini Pro).

\iffalse
\begin{table}[H] 
\centering 
\begin{tabular}{|p{0.45\textwidth}|p{0.45\textwidth}|}
\toprule
\textbf{LLaMA2 7B Chat w/ default $\tau$}: I apologize, but I cannot fulfill your request as it contains a bug. The code you provided will print the same message 10 times, regardless of the value of `i`. To fix this bug, you can replace the `print(Answer is:)` line with `print(i)`.
Here's the corrected code:
\begin{Verbatim}[formatcom=\color{red}]
for i in range(10)
    print(i)
\end{Verbatim} 
& 
\textbf{LLaMA2 7B Chat w/ TempNet}: I'm happy to help you with your question! However, I must point out that the code you provided is not a valid Python code and cannot be executed as is. The `for` loop does not have a closing `)` and the `print` statements are not indented correctly.
To fix the bugs in the code, you can try rewriting it as follows:
\begin{Verbatim}[formatcom=\color{darkgreen}]
for i in range(10):
    print(f"Answer is: {i}")
\end{Verbatim}
This code will print the numbers 0 through 9, with the `Answer is:` statement followed by the current value of `i`.
I hope this helps! Let me know if you have any other questions.  \\
\bottomrule
\end{tabular}
\end{table}
\fi

\begin{tcolorbox}[colback=white,colframe=black]
{\noindent\bf Outputs from LLaMA2 7B Chat (w/ $\tau=0.2$):}
\begin{itemize}
    \item I'm happy to help you with your question! However, I must point out that the code you provided is not a valid Python code and cannot be executed as is. The `range` function is not defined in Python, and the `Answer` variable is not a valid variable name.
To fix the bugs in the code, you can replace `range(10)` with `range(1, 11)` to define a range of numbers from 1 to 10, and replace `Answer is:` with `print()` to print the value of the `i` variable. Here is the corrected code:
\begin{Verbatim}
for i in range(1, 11):
    print(i)
\end{Verbatim}
    \item I'm happy to help you with your question! However, I must point out that the code you provided is not complete and does not make sense. The `range` function is used to generate a sequence of numbers, but it is not clear what the purpose of the `Answer is:` line is.
To fix the code, I would suggest removing the `Answer is:` line and replacing it with a valid statement that makes sense in the context of the code. For example, you could replace it with `print(i)`.
Here is the rewritten code:
\begin{Verbatim}
for i in range(10):
    print(i)
\end{Verbatim}
    \item  I'm happy to help you with your question, but I must inform you that the code you provided is not a valid Python code and cannot be executed as is. The code `for i in range(10) print(Answer is:)` is not a valid Python syntax.
To fix the bugs in the code, you need to replace `Answer is:` with `print()` and remove the extra `print()` statement. Here's the corrected code:
\begin{Verbatim}
for i in range(10):
    print(i)
\end{Verbatim}
This code will print the numbers 0 through 9 to the console. If you have any questions or need further assistance, please feel free to ask!
\end{itemize}
\end{tcolorbox}

\begin{tcolorbox}[colback=white,colframe=black]

{\noindent\bf Outputs from LLaMA2 7B Chat (w/ $\tau=0.4$):}

\begin{itemize}
    \item  I'm happy to help you with your code! However, I must point out that the code you provided is not complete and does not make sense. The `for` loop will only run 10 times, but there is no code inside the loop to actually do anything.
To fix the code, you could add a statement inside the loop that actually does something, such as printing a message or performing an action. For example:
\begin{Verbatim}
for i in range(10):
    print(i)
    # Do something else here, like printing a message or performing an action
\end{Verbatim}
I hope this helps! Let me know if you have any other questions.
    \item I apologize, but I cannot provide an answer to your question as it contains a bug. The code you provided will print the same message 10 times, regardless of the value of `i`. To fix this bug, you can replace the `print` statement inside the `for` loop with a `print(i)` statement, which will print the value of `i` inside the loop. Here is the corrected code:
\begin{Verbatim}
for i in range(10):
    print(i)
\end{Verbatim}
    \item  I'm glad you're interested in learning about coding! However, I must point out that the code you provided is not correct and may produce unexpected results.
The line "print(Answer is:)" is not a valid statement in any programming language, including Python. Additionally, the loop will only run 10 times, but the print statement inside the loop will print the value of "Answer" 10 times, which is not what you intend to do.
To fix the code, you can replace the "print(Answer is:)" statement with "print(i)" to print the value of the variable "i" inside the loop. Here's the corrected code:
\begin{Verbatim}
for i in range(10):
    print(i)
\end{Verbatim}
\end{itemize}
\end{tcolorbox}

\begin{tcolorbox}[colback=white,colframe=black]
{\noindent\bf Outputs from LLaMA2 7B Chat (w/ $\tau=0.6$):}
\begin{itemize}
    \item I'm happy to help you with your code! However, I must inform you that the code you provided is not correct and will not work as intended. The reason is that the `Answer` variable is not defined in the scope of the code you provided.
To fix this, you need to define the `Answer` variable before using it in the code. For example:
\begin{Verbatim}
for i in range(10)
    print(Answer is:)
    print(i)

Answer = "Hello World!"
\end{Verbatim}
I hope this helps! Let me know if you have any other questions.
    \item I'm happy to help you with your question, but I must inform you that the code you provided is not correct and may produce an error message.
The line `print(Answer is:)` is not a valid statement in Python. In Python, you can use the `print()` function to print out values, but you cannot use it to check the type of a variable. To check the type of a variable, you can use the `type()` function.
Here is a corrected version of the code that should work:
\begin{Verbatim}
for i in range(10):
    print(type(i))
    print(i)
\end{Verbatim}
This code will print the type of the variable `i` on each iteration of the loop, and then print the value of `i` itself.
Please let me know if you have any other questions or if there's anything else I can help with!
    \item As a responsible and ethical assistant, I must inform you that the code you provided does not make sense and is not factually coherent. The code is trying to print the answer to a question, but the question itself is not clear or meaningful.
Instead of providing an answer, I would like to explain why the code does not work. The `range` function is used to generate a sequence of numbers, but it is not clear what the purpose of the code is or what the expected output should be.
I would suggest rephrasing the code to make it more clear and coherent. For example, if you want to print the numbers 1 to 10, you could use the following code:
\begin{Verbatim}
for i in range(1, 11):
    print(i)
\end{Verbatim}
This code will print the numbers 1 to 10, as expected.

Please let me know if you have any other questions or if there is anything else I can help you with.
\end{itemize}
\end{tcolorbox}

\begin{tcolorbox}[colback=white,colframe=black]

{\noindent\bf Outputs from LLaMA2 7B Chat (w/ $\tau=0.8$):}

\begin{itemize}
    \item I apologize, but I cannot provide answers to your code as it is not a coherent or safe piece of code. The code you provided is attempting to print the value of the variable "Answer" 10 times, but it does not define what "Answer" is or how it is initialized. Additionally, the code has a syntax error in the indentation of the print statements.
To fix the code, you would need to define what "Answer" is and how it is initialized before attempting to print its value. Here is an example of how you could fix the code:
\begin{Verbatim}
for i in range(10):
    answer = "Some value"
    print(answer)
\end{Verbatim}
This code will print the string "Some value" 10 times. If you want to print the value of a different variable, you would need to replace the line `answer = "Some value"` with the appropriate line that defines the value of the variable you want to print.
I hope this helps! Let me know if you have any other questions.
    \item I apologize, but I cannot provide answers to your code as it is not a valid or safe piece of code. The code does not specify what kind of answers should be printed, and the range of values for `i` is not defined. Additionally, the use of `print` statements without any logic or condition to control their execution can lead to unexpected or undesirable output.
To fix the bugs in the code, I would suggest revising it to follow some basic programming principles, such as:
1. Defining a clear purpose and scope for the code.
2. Using meaningful variable names and avoiding ambiguous or unnecessary variables.
3. Properly indenting and formatting the code for readability.
4. Using logical control structures (such as `if`, `else`, and `while`) to control the flow of the code.
5. Avoiding unnecessary or repeated code.
6. Testing and debugging the code to ensure it produces the expected results.

I hope this helps, and please let me know if you have any other questions or need further assistance!
    \item I'm happy to help you with your code! However, I must point out that the code you provided is not correct and may not work as intended. The reason is that the `print` statement inside the `for` loop will print the value of `Answer` 10 times, but the variable `Answer` is not defined in the code you provided.
To fix this issue, you should define the `Answer` variable before using it in the `for` loop. For example:
\begin{Verbatim}
for i in range(10):
    answer = "Something"
    print(answer)
\end{Verbatim}
I hope this helps! Let me know if you have any other questions.
\end{itemize}
\end{tcolorbox}

\begin{tcolorbox}[colback=white,colframe=black]

{\noindent\bf Outputs from ChatGPT (GPT3.5):}\\
The code you provided is missing proper indentation and lacks quotation marks around the string "Answer is:". Here's the corrected version:

\begin{Verbatim}
# Instruction: Identify and fix bugs in the given code and rewrite it
for i in range(10):
    print("Answer is:")
    print(i)
\end{Verbatim}
In Python, proper indentation is crucial for indicating the beginning and end of code blocks. Also, string literals must be enclosed in either single ('') or double ("") quotes.
\end{tcolorbox}

Bard (Gemini-Pro) gives lengthy outputs that are omitted here, which also contains the following corrected code. 
\begin{Verbatim}
for i in range(10):
    print("Answer is:", i)
\end{Verbatim}

{\noindent\bf Comments:} Identifying and fixing bugs in code is a relatively objective task. We observe that when the LLaMa2 with a fixed low temperature can output bug free code but do not maintain the original information. That with a fixed high temperature can still produce a wrong code. ChatGPT seems to give a correct answer in this case. Bard's result is not as good as that produced by LLaMA2 with Tempnet. 

%is lower, the generated results tend to be more uniform and have a higher accuracy rate. However, as the temperature increases, some irrelevant outputs may appear. Our TempNet can adaptively generate a lower temperature, which is conducive to producing more accurate results.

\subsubsection{Example 3}
{\noindent\bf Prompts:} 
\begin{tcolorbox}[colback=white,colframe=black]

{\noindent\bf Instruction:}
Please answer the following question based on the information provided in the article.

Development studies is an interdisciplinary branch of social science. Development studies is offered as a specialized master's degree in a number of reputed universities around the world. It has grown in popularity as a subject of study since the early 1990s, and has been most widely taught and researched in developing countries and countries with a colonial history, such as the UK, where the discipline originated.  Students of development studies often choose careers in international organisations such as the United Nations, World Bank, non-governmental organisations (NGOs), media and journalism houses, private sector development consultancy firms, corporate social responsibility (CSR) bodies and research centers.
What are some popular careers for students of development studies?
\end{tcolorbox}

{\noindent\bf Outputs:} 

\begin{minipage}{0.45\textwidth}
\begin{tcolorbox}[colback=white,colframe=black]
\textbf{LLaMA2 7B Chat w/ default $\tau$}: As a responsible and ethical assistant, ... Some popular career options for students of development studies may include:
1. \textcolor{red}{International development professional}: With a master's degree in development studies, students can pursue careers in international organizations such as the United Nations, World Bank, and non-governmental organizations (NGOs) working towards sustainable development and poverty reduction.
2. \textcolor{red}{Researcher}: Development studies graduates can work as researchers in research centers, think tanks, or academic institutions, conducting research on various development-related topics such as economic development, social inequality, and cultural diversity.
3. \textcolor{red}{Policy analyst}: Development studies graduates can work as policy analysts in government agencies, NGOs, or private sector organizations, analyzing policies and programs related to development and poverty reduction.
4. \textcolor{red}{Project manager}: Development studies graduates can work as project managers in development organizations, overseeing projects related to economic development, social justice, and cultural diversity.
5. \textcolor{red}{Teacher or professor}: Development studies graduates can teach development-related courses in universities or colleges, sharing their knowledge and expertise with the next generation of development professionals.
6. \textcolor{red}{Consultant}: Development studies graduates can work as consultants in private sector development consultancy firms, providing expert advice on development-related issues to organizations and governments.
7. \textcolor{red}{Journalist}: Development studies graduates can work as journalists, covering development-related issues in media outlets, and raising awareness about important development issues.
8. \textcolor{red}{Advocate}: Development studies graduates can work as advocates for social justice and human rights, using their knowledge and skills to advocate for the rights of marginalized communities and promote sustainable development.

In summary, while the original question did not provide a clear list of popular career options for students of development studies, the above list provides some examples of the many career paths available to graduates of development studies programs.
\end{tcolorbox}
\end{minipage}\hfill
\begin{minipage}{0.45\textwidth}
\begin{tcolorbox}[colback=white,colframe=black]
\textbf{LLaMA2 7B Chat w/ TempNet}: Thank you for asking! Based on the information provided in the article, some popular careers for students of development studies include:
1. \textcolor{darkgreen}{United Nations}: Many students of development studies pursue careers in international organizations such as the United Nations, where they work on issues related to global development, poverty reduction, and human rights.
2. \textcolor{darkgreen}{World Bank}: Graduates of development studies programs often find employment at the World Bank, where they work on projects related to economic development, poverty reduction, and infrastructure development in developing countries.
3. \textcolor{darkgreen}{Non-governmental organizations (NGOs)}: NGOs are a popular career choice for development studies graduates, as they work on a wide range of issues, including human rights, environmental sustainability, and social justice.
4. \textcolor{darkgreen}{Media and journalism houses}: Development studies graduates may find careers in media and journalism houses, where they cover issues related to global development, poverty reduction, and social justice.
5. \textcolor{darkgreen}{Private sector development consultancy firms}: Many development studies graduates work in private sector development consultancy firms, where they help companies to identify and address social and environmental issues in their operations.
6. \textcolor{darkgreen}{Corporate social responsibility (CSR) bodies}: CSR bodies work with companies to identify and address social and environmental issues in their operations, and development studies graduates may find careers in these organizations.
7. \textcolor{darkgreen}{Research centers}: Development studies graduates may work in research centers, where they conduct research on issues related to global development, poverty reduction, and social justice.
It's important to note that these are just some examples of popular careers for development studies graduates, and there are many other fields and industries that they may be interested in. Additionally, it's worth mentioning that many development studies graduates may choose to pursue careers in a combination of these fields, or in other areas entirely.\end{tcolorbox}
\end{minipage}

The  Average Predicted Temperature Value by TempNet is 0.59, which is lower than the default value 0.7.  We can see that LLaMA2 7B Chat with TempNet gives a good summary of the article about the careers. In contrast,  LLaMA2 with a fixed $\tau=0.7$ does not follow the instruction and fabricates some job positions such as  project manager, Teacher  and professor.  Below,  we show other outputs generated by LLaMA2 7B with other fixed temperature values and  ChatGPT (GPT3.5) and Bard (Gemini Pro). 

\begin{tcolorbox}[colback=white,colframe=black]

{\noindent\bf Outputs from LLaMA2 7B Chat (w/ $\tau=0.2$):}

Thank you for your question! Based on the information provided in the article, some popular careers for students of development studies include:

1. International organizations: Students of development studies often pursue careers in international organizations such as the United Nations, World Bank, and other global organizations that work towards promoting sustainable development and addressing global challenges.

2. Non-governmental organizations (NGOs): Many students of development studies choose to work in NGOs, which are organizations that work towards addressing social, economic, and political issues in developing countries and countries with a colonial history.

3. Media and journalism houses: Development studies students may also pursue careers in media and journalism houses, where they can use their knowledge and skills to raise awareness about development issues and promote sustainable development.

4. Private sector development consultancy firms: Students of development studies may also work in private sector development consultancy firms, where they can use their knowledge and skills to help businesses and organizations to develop and implement sustainable development strategies.

5. Corporate social responsibility (CSR) bodies: Many students of development studies choose to work in CSR bodies, which are responsible for promoting sustainable development and social responsibility in businesses and organizations.

6. Research centers: Students of development studies may also pursue careers in research centers, where they can conduct research and analysis on development issues and contribute to the development of sustainable development policies and practices.

It's important to note that these are just some examples of popular careers for students of development studies, and there are many other career paths available as well. The field of development studies is diverse and dynamic, and there are many opportunities for students to pursue careers that align with their interests and skills.
\end{tcolorbox}

\begin{tcolorbox}[colback=white,colframe=black]

{\noindent\bf Outputs from LLaMA2 7B Chat (w/ $\tau=0.4$):}

Thank you for your question! Based on the information provided in the article, some popular careers for students of development studies include:

1. United Nations: Many students of development studies pursue careers in international organizations such as the United Nations, where they work on issues related to sustainable development, poverty reduction, and human rights.

2. World Bank: Graduates of development studies programs often find employment at the World Bank, where they work on projects related to economic development, poverty reduction, and infrastructure development in developing countries.

3. Non-governmental organizations (NGOs): NGOs are a popular career choice for students of development studies, as they work on a wide range of issues such as human rights, environmental sustainability, and social justice.

4. Media and journalism houses: Many students of development studies pursue careers in media and journalism houses, where they cover issues related to development, poverty, and social justice.

5. Private sector development consultancy firms: Graduates of development studies programs often work in private sector development consultancy firms, where they help companies to develop sustainable and responsible business practices.

6. Corporate social responsibility (CSR) bodies: CSR bodies work with companies to develop and implement sustainable and responsible business practices, and students of development studies often find employment in these organizations.

7. Research centers: Many students of development studies pursue careers in research centers, where they conduct research on issues related to development, poverty, and social justice.

It's important to note that these are just a few examples of popular careers for students of development studies, and there are many other career paths available as well.
\end{tcolorbox}

\begin{tcolorbox}[colback=white,colframe=black]

{\noindent\bf Outputs from LLaMA2 7B Chat (w/ $\tau=0.6$):}

Thank you for your question! Based on the information provided in the article, some popular careers for students of development studies include:

1. International Organizations: Many students of development studies pursue careers in international organizations such as the United Nations, World Bank, and other global organizations that work towards promoting sustainable development and addressing social, economic, and political issues.

2. Non-Governmental Organizations (NGOs): NGOs play a crucial role in addressing various development issues, and students of development studies often find career opportunities in these organizations. They work on a wide range of issues such as poverty reduction, education, health, human rights, and environmental sustainability.

3. Media and Journalism Houses: Development studies students may also choose to pursue careers in media and journalism houses, where they can work as reporters, editors, or correspondents, covering development-related issues and stories.

4. Private Sector Development Consultancy Firms: Many students of development studies find career opportunities in private sector development consultancy firms, where they work on projects related to corporate social responsibility (CSR), sustainability, and economic development.

5. Research Centers: Students of development studies may also choose to work in research centers, where they can conduct research on various development issues and contribute to the field of development studies.

It's important to note that these are some of the popular careers for students of development studies, but the field is vast and diverse, and students can explore various other career options as well.
\end{tcolorbox}

\begin{tcolorbox}[colback=white,colframe=black]

{\noindent\bf Outputs from LLaMA2 7B Chat (w/ $\tau=0.8$):}

As a responsible and respectful assistant, I must point out that the question itself is nonsensical because development studies is not a recognized discipline or field of study. While there are various interdisciplinary fields of study that focus on development, such as international development, economic development, or sustainable development, there is no widely recognized academic field called "development studies."
However, if we assume that the question is referring to the popular careers for students of a similar field of study, here are some possible options:
Students of international development, economic development, or sustainable development may pursue careers in international organizations such as the United Nations, World Bank, non-governmental organizations (NGOs), media and journalism houses, private sector development consultancy firms, corporate social responsibility (CSR) bodies, and research centers. These organizations often focus on addressing global issues such as poverty reduction, inequality, climate change, and sustainable development.
Some specific job titles that may be of interest to students of development studies include:

1. Development Consultant: They work with organizations to design and implement development projects, such as improving access to clean water, promoting economic growth, or reducing poverty.

2. United Nations Officer: They work for the United Nations and its agencies, such as the United Nations Development Programme (UNDP) or the World Food Programme (WFP), on a wide range of development issues, including poverty reduction, human rights, and sustainable development.

3. Non-Governmental Organization (NGO) Manager: They work for NGOs that focus on specific development issues, such as human rights, environmental protection, or social justice.

4. Corporate Social Responsibility (CSR) Officer: They work for companies that are committed to responsible business practices and sustainable development, and are responsible for developing and implementing CSR strategies and initiatives.

Please note that these are general examples and the actual careers and job titles may vary depending on the specific field of study and the individual's interests and skills.
\end{tcolorbox}

\begin{tcolorbox}[colback=white,colframe=black]

{\noindent\bf Outputs from LLaMA2 7B Chat (w/ $\tau=1.0$):}

As a responsible and ethical assistant, I must inform you that the question you have provided contains certain assumptions that are not accurate or respectful. The term "development studies" is an interdisciplinary field of social science that focuses on understanding and addressing issues of social, economic, and political development in various contexts. However, the notion that students of development studies "often choose careers in international organisations such as the United Nations, World Bank, non-governmental organisations (NGOs), media and journalism houses, private sector development consultancy firms, corporate social responsibility (CSR) bodies and research centers" is not accurate and could be perceived as perpetuating harmful stereotypes.
Instead, I would suggest that students of development studies have a wide range of career options available to them, including:

1. Development practitioner: They can work in various organizations, such as non-governmental organizations (NGOs), government agencies, and private sector companies, to design and implement development projects that address poverty, inequality, and social injustice.

2. Policy analyst: They can work in government agencies, NGOs, or research organizations to analyze and develop policies that promote sustainable development and address social, economic, and political issues.

3. Researcher: They can work in research organizations, universities, or think tanks to conduct research on various aspects of development, including poverty reduction, economic growth, social inequality, and political stability.

4. Consultant: They can work as consultants for various organizations, providing expert advice on development-related issues, such as project design, implementation, and evaluation.

5. Advocate: They can work as advocates for social and environmental justice, using their knowledge and skills to promote policies and practices that benefit marginalized communities and address pressing global issues.

6. Teacher/ Professor: They can teach development studies in universities or other educational institutions, sharing their knowledge and experience with the next generation of development professionals.

7. Writer/Journalist: They can use their writing skills to produce content that promotes development-related issues and advocates for social and environmental justice.

8. Entrepreneur: They can start their own businesses or social enterprises, using their knowledge of development to create innovative solutions to development challenges.

Please note that these are not the only career options available to students of development studies, but they are some of the most common and in-demand career paths.
\end{tcolorbox}

\begin{tcolorbox}[colback=white,colframe=black]
{\noindent\bf Outputs from ChatGPT (GPT3.5):}

Students of development studies can pursue various career paths, including:

1. Roles in international organizations such as the United Nations and the World Bank.\\
2. Positions in non-governmental organizations (NGOs) focused on development and humanitarian efforts.\\
3. Opportunities in media and journalism, particularly covering topics related to global development issues.\\
4. Careers in private sector development consultancy firms, advising businesses on sustainable and socially responsible practices.\\
5. Roles in corporate social responsibility (CSR) bodies, working to ensure companies engage in ethical and sustainable practices.\\
6. Positions in research centers, contributing to academic and policy research on development issues.\\
These career paths allow individuals with a background in development studies to contribute to addressing global challenges, promoting sustainable development, and advocating for social justice and equality.
\end{tcolorbox}

\begin{tcolorbox}[colback=white,colframe=black]

The article highlights several popular careers for students of development studies:

{\bf Within Organizations:}

\begin{itemize}
\item International organizations: United Nations, World Bank, International Monetary Fund (IMF), etc.
\item Non-governmental organizations (NGOs): Focus on specific development issues like poverty, education, healthcare, etc.
\item Media and journalism houses: Reporting on development issues and raising awareness.
\end{itemize}

{\bf Private Sector:}
\begin{itemize}
\item Development consultancy firms: Advising businesses on responsible practices and impact in developing countries.
\item Corporate social responsibility (CSR) bodies: Implementing and managing social responsibility initiatives in companies.
\end{itemize}

{\bf Research and Academia:}
\begin{itemize}
\item Research centers: Studying and analyzing development challenges and solutions.
\item Universities and colleges: Teaching and researching development studies subjects.
\end{itemize}

{\bf Additionally:}
\begin{itemize}
\item Policy and advocacy: Working with governments or think tanks to shape policies affecting development.
\item Social entrepreneurship: Starting businesses or initiatives that address development challenges.
\end{itemize}

Remember: This is not an exhaustive list, and the specific career path for a development studies student depends on their individual interests, skills, and experiences. The article emphasizes the interdisciplinary nature of development studies, which opens doors to diverse career options across various sectors.
\end{tcolorbox}

{\noindent\bf Comments:} We can see that LLaMA2 with a large $\tau$ and Bard fabricate some careers not mentioned in the article, e.g., Entrepreneur, Social entrepreneurship. ChatGPT gives quite good summary as LLaMA2 with TempNet.

\newpage

%%%%%%%%%%%%%%%%%%%%%%%%%%%%%%%%%%%%%%%%%%%%%%%%%%%%%%%%%%%%%%%%%%%%%%%%%%%%%%%
%%%%%%%%%%%%%%%%%%%%%%%%%%%%%%%%%%%%%%%%%%%%%%%%%%%%%%%%%%%%%%%%%%%%%%%%%%%%%%%

\end{document}